%% file: main.tex
\pdfoutput=1
\documentclass{article}

\usepackage{arxiv}

\usepackage[utf8]{inputenc} 
\usepackage[T1]{fontenc}  
\usepackage{hyperref} 
\usepackage{url} 
\usepackage{booktabs} 
\usepackage{amsfonts} 
\usepackage{nicefrac} 
\usepackage{microtype} 
\usepackage{lipsum} 
\usepackage{graphicx}
\usepackage[numbers]{natbib}
\usepackage{doi}
\usepackage{caption}
\usepackage{multirow}
\usepackage{mathtools}
\usepackage{amssymb}
\usepackage{bbm}
\usepackage{dsfont}
\usepackage{subfig}
\usepackage{xcolor}
\usepackage{wrapfig}
\usepackage[font=small,labelfont=bf,tableposition=top]{caption}
\usepackage{floatrow}
\usepackage{wrapfig}

\newcommand{\tableScale}{0.8}
\newcommand{\figScaleThree}{0.39}
\newcommand{\figScale}{0.35}

\usepackage{amsmath}

\newfloatcommand{capbtabbox}{table}[][\FBwidth]

\title{Adaptive Fine-Tuning of Transformer-Based Language Models for Named Entity Recognition}

\date{}

\author{Felix Stollenwerk\\
Arbetsförmedlingen\\ 
(Swedish Public Employment Service)\\ 
\texttt{felix.stollenwerk@arbetsformedlingen.se}\\	
}

\renewcommand{\headeright}{F.~Stollenwerk}
\renewcommand{\undertitle}{}
\renewcommand{\shorttitle}{Adaptive Fine-Tuning of Transformer-Based Language Models for NER}

\hypersetup{
pdftitle={Adaptive Fine-Tuning of Transformer-Based Language Models for Named Entity Recognition},
pdfsubject={cs.CL, cs.LG},
pdfauthor={Felix Stollenwerk},
pdfkeywords={language models, fine-tuning, named entity recognition},
}

\begin{document}
\maketitle

\input{tex/0_abstract.tex}

\section{Introduction \label{sec:introduction}}
\input{tex/1_introduction.tex}

\section{Background \& Related Work \label{sec:background}}
\input{tex/2_background_stability.tex}
\input{tex/2_background_fine_tuning.tex}

\section{Adaptive Fine-tuning \label{sec:adaptive}}
\input{tex/3_finetuning_approach.tex} 
\input{tex/3_finetuning_computational_effort.tex}

\section{Experiments \label{sec:experiments}}
\input{tex/4_experiments_introduction.tex}
\input{tex/4_experiments_analysis.tex}

\section{Results \label{sec:results_fine_tuning}}

\input{tex/5_results_finetuning.tex}

\section{Summary \label{sec:summary}}
\input{tex/6_summary.tex}

\clearpage
\bibliographystyle{unsrt}
\bibliography{references}

\clearpage
\appendix
\section{Hyperparameters \label{app:hyperparameters}}
\input{tex/appendix_hyperparameters.tex}

\section{Notation \label{app:notation_uncertainties}}
\input{tex/appendix_uncertainties.tex}

\section{Model Evaluation \label{sec:model_evaluation}}
\input{tex/appendix_model_evaluation.tex}

\section{Reproduction of Literature Results \label{app:experiments_confirmation}}
\input{tex/appendix_results_confirmation.tex}

\section{Computational Effort \label{app:finetuning_computational_effort}}
\input{tex/appendix_finetuning_computational_effort.tex}

\clearpage
\section{Results: Generalization \label{app:finetuning_generalization}}
\input{tex/appendix_finetuning_generalization.tex}

\clearpage
\section{Variants of the Adaptive Fine-tuning Approach \label{app:escd_variations}}
\input{tex/appendix_variations.tex}

\clearpage 
\section{Ablation Studies \label{app:finetuning_ablation}}
\input{tex/appendix_finetuning_ablation.tex}

\clearpage 
\section{Learning Curves \label{app:finetuning_additional}}
\input{tex/appendix_finetuning_additional.tex}

\clearpage 
\section{Dependency of the Model Performance on the Training Dataset Size and Training Epochs \label{app:finetuning_dependency}}
\input{tex/appendix_finetuning_dependency.tex}

\clearpage 
\section{Dependency of the Model Performance on the Validation Dataset Size \label{app:finetuning_dependency_validation}}  
\input{tex/appendix_adaptive_finetuning_in_practice_xval_dependency.tex}

\clearpage 
\section{Adaptive Fine-tuning in Practice: Using the Validation Dataset for Training \label{sec:adaptive_finetuning_in_practice}} 
\input{tex/appendix_adaptive_finetuning_in_practice_train_on_val.tex} 

\end{document}

%% file: tex/0_abstract.tex
\begin{abstract}
The current standard approach for fine-tuning transformer-based language models includes a fixed number of training epochs and a linear learning rate schedule.
In order to obtain a near-optimal model for the given downstream task,  a search in optimization hyperparameter space is usually required. 
In particular, the number of training epochs needs to be adjusted to the dataset size.  
In this paper, we introduce \textit{adaptive fine-tuning}, which is an alternative approach that uses early stopping and a custom learning rate schedule to dynamically adjust the number of training epochs to the dataset size. 
For the example use case of named entity recognition, we show that our approach not only makes hyperparameter search with respect to the number of training epochs redundant, but also leads to improved results in terms of performance, stability and efficiency. 
This holds true especially for small datasets, where we outperform the state-of-the-art fine-tuning method by a large margin. 
\end{abstract}

%% file: tex/1_introduction.tex
The use of transformer-based models has had a game-changing impact on natural language processing in the last few years.  A multitude of pretrained models are readily available in many languages \citep{wolf-etal-2020-transformers} and can be fine-tuned on a labeled dataset for a specific downstream task.  Due to the inherent transfer learning capabilities of the models, excellent performance can be achieved with relatively small datasets. 
As the model architecture is determined by the pretrained model, only optimization hyperparameters need to be specified for the fine-tuning.  This includes e.g.~the choice of an optimizer, the learning rate schedule, the batch size as well as the number $N_{\rm epochs}$ of times the training dataset is fed to the model. 
Usually, the dataset is split into a training, validation and test set.  A simple grid in hyperparameter space (e.g.~\citep{devlin2019bert}) is then explored, and the model with the hyperparameter combination that performs best on the validation set is used for inference, while the test set serves to verify the ability of the model predictions to generalize.

For most of the hyperparameters, strong and generally applicable baseline settings have been found, such that the need for hyperparameter search is somewhat reduced (more on this in Sec.~\ref{subsec:background_fine_tuning}). 
The number of training epochs, however, is special among the hyperparameters as its optimal value is intrinsically tied to the size\footnote{Other features of the dataset like complexity also have an influence.} of the dataset, see e.g.~\citep{zhang2021revisiting}.  In practice, state-of-the-art performances have been achieved using a wide range of training epochs, from $N_{\rm epochs} = 3$ \citep{devlin2019bert} to $N_{\rm epochs} = 20$ \citep{mosbach2021stability}. While a small number of training epochs is suitable for larger datasets, more epochs are needed for smaller datasets. In the end,  one needs to tune the number of training epochs to achieve optimal results for a given dataset \citep{zhang2021revisiting}.

In this work, we introduce \textit{adaptive fine-tuning}.  In contrast to the previously described \textit{fixed epoch} approach, it uses early stopping and automatically adjusts the number of training epochs to account for the dataset size.  This way,  a close-to-optimal model is found for any dataset without the need for multiple hyperparameter runs. 
While adaptive fine-tuning (like the fixed epoch approach) can be applied to any downstream task, we study it here for named entity recognition (NER). 

The paper is structured as follows:
In Sec.~\ref{sec:background}, the background of our contribution and related work are outlined.  Next, we explain the adaptive fine-tuning approach in Sec.~\ref{sec:adaptive}.
Afterwards, we discuss the experiments we conduct (Sec.~\ref{sec:experiments}) and their results (Sec.~\ref{sec:results_fine_tuning}).
We conclude with a summary in Sec.~\ref{sec:summary}.

%% file: tex/2_background_stability.tex
\subsection{
Performance, stability and efficiency
\label{subsec:background_stability}
}

The quality of a fine-tuned model is evaluated by measuring its \textit{performance} on the test dataset.  The very definition of performance depends on the downstream task. For a multiclass classification problem like NER,  the main metric to consider is usually the $f_1$ score (see App.~\ref{sec:model_evaluation} for details).

However, regarding the fine-tuning process itself, performance is not the only relevant quantity. 
Training of neural networks includes stochastic processes, like weight initialization or shuffling of the training dataset. 
Hence---even if the same data, architecture and hyperparameters are used---the results are not deterministic but depend on the employed random seed, and the dependency can be particularly strong for small datasets \citep{dodge2020finetuning}.
Even if training with one specific random seed leads to great performance, the use of another one may lead to poorer results or even divergence.  It is of great interest to keep the dependence of the results on the random seed weak,  not least in order to reduce the need for retraining.

The \textit{stability} of fine-tuning for a given setting can be assessed using multiple identical runs with different random initializations. 
In a first step, the fraction of converged runs is determined. 
In a second step,  on the subset of converged runs, one may compute the standard deviation of the performance (and other quantities of interest) to gain control over the statistical uncertainties.
Note that a certain dependence of the performance on the initialization is unavoidable.  Trying out several random seeds has been shown to improve the expected performance \citep{dodge2020finetuning, dodge2019work}, i.e.  the performance of the best of the trained models.

Finally,  the computational effort that is needed to fine-tune a model is of relevance. If the same performance and stability is achieved with two different approaches, we use the term \textit{efficiency} synonymously with computational effort. 

%% file: tex/2_background_fine_tuning.tex
\subsection{Fixed epoch fine-tuning \label{subsec:background_fine_tuning}}

For the first transformer-based encoder model, BERT,  a hyperparameter grid was specified in the original paper \citep{devlin2019bert}. In particular, a model was fine-tuned for $N_{\rm epochs} = 3-5$ training epochs with a linearly decaying learning rate.
Similar heuristic global hyperparameter grids were employed for other transformer-based architectures, see e.g.~\citep{clark2020electra} for the ELECTRA model. 
The original fine-tuning setup turned out to be prone to the instabilities described in Sec. \ref{subsec:background_stability}, i.e.  large performance variance or even divergence depending on the employed random initialization.  This problem as well as potential remedies have been discussed in many papers, see e.g.~\citep{dodge2020finetuning, lee2020mixout}.
It is nowadays standard practice to use the Adam optimizer with weight decay and a linear warm-up period of the learning rate followed by linear decay, for instance.
The most recent publication on the topic,  \citep{mosbach2021stability}, attributes the instabilities to vanishing gradients at the beginning of training and generalization incapabilities at the end of training, and proposes small learning rates with Adam bias correction and an increased number of $N_{\rm epochs} = 20$ training epochs to overcome the problems. 
Their approach achieves improvements in terms of performance and stability. 
We refer to it as \textit{stable} fine-tuning, use it as a baseline and adopt all hyperparameters from it throughout this work (except for the number of training epochs and the learning rate schedule---more on this in Sec.~\ref{sec:adaptive}).  An overview of the hyperparameters can be found in App.~\ref{app:hyperparameters}.  
In addition, we consider a variant of the stable approach with identical hyperparameters except for that it uses $N_{\rm epochs} = 5$. We will simply call it the \textit{original} fine-tuning approach although it contains all the aforementioned improvements with respect to the original BERT paper \citep{devlin2019bert}.  
The original and stable fine-tuning approaches are two special cases of the \textit{fixed epoch} approach that is currently state-of-the-art.

\begin{figure}
    \centering
    \begin{floatrow}
             \ffigbox{
                \includegraphics[scale=\figScaleThree]{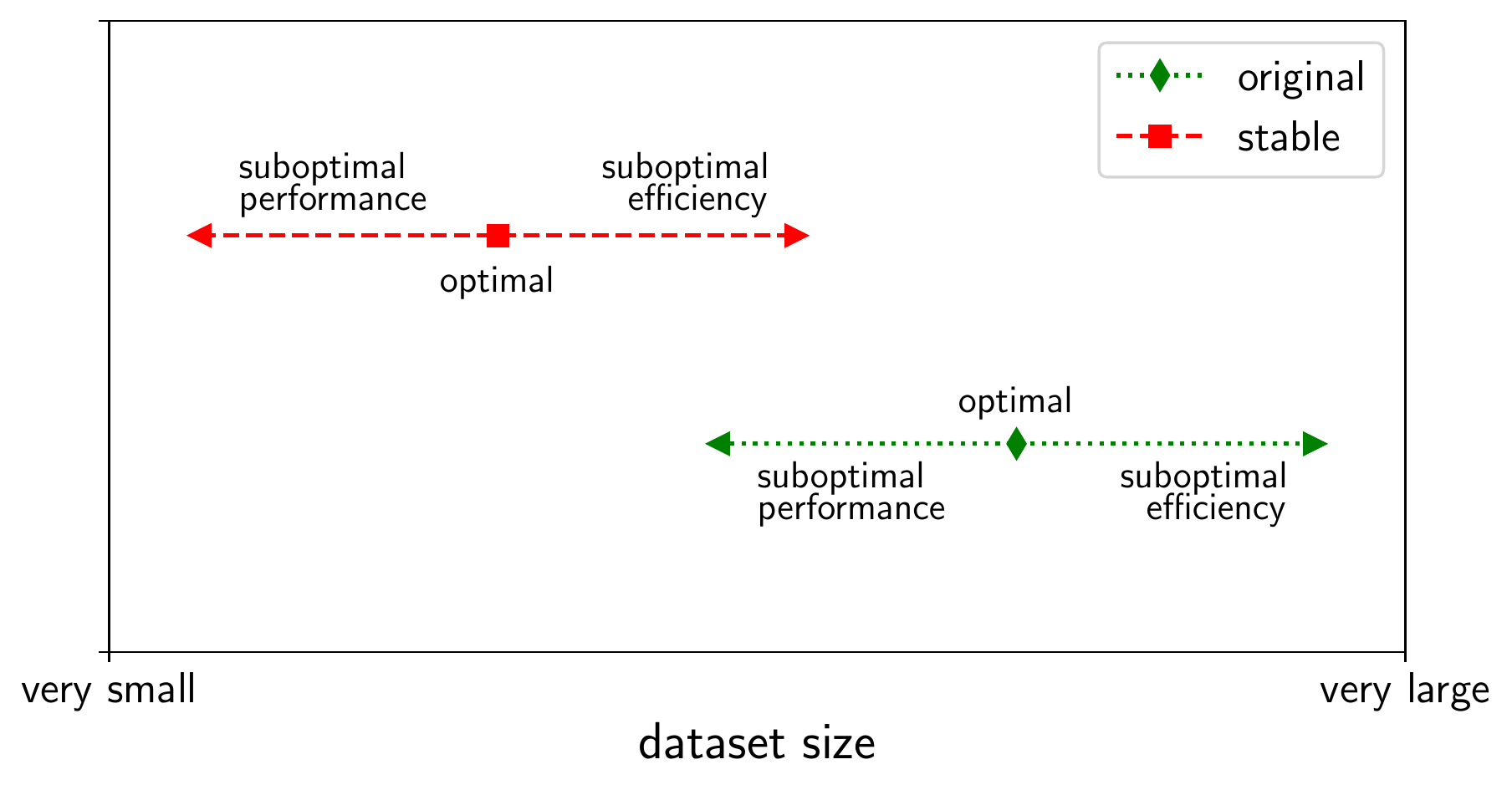}
             }{
			\caption{Qualitative dependence of performance and efficiency on the dataset size for the fixed epoch fine-tuning approach (original: $N_{\rm epochs} = 5$, stable: $N_{\rm epochs} = 20$).}
 			\label{fig:approaches_principle}        
            }
            \ffigbox{
				\includegraphics[scale=\figScale]{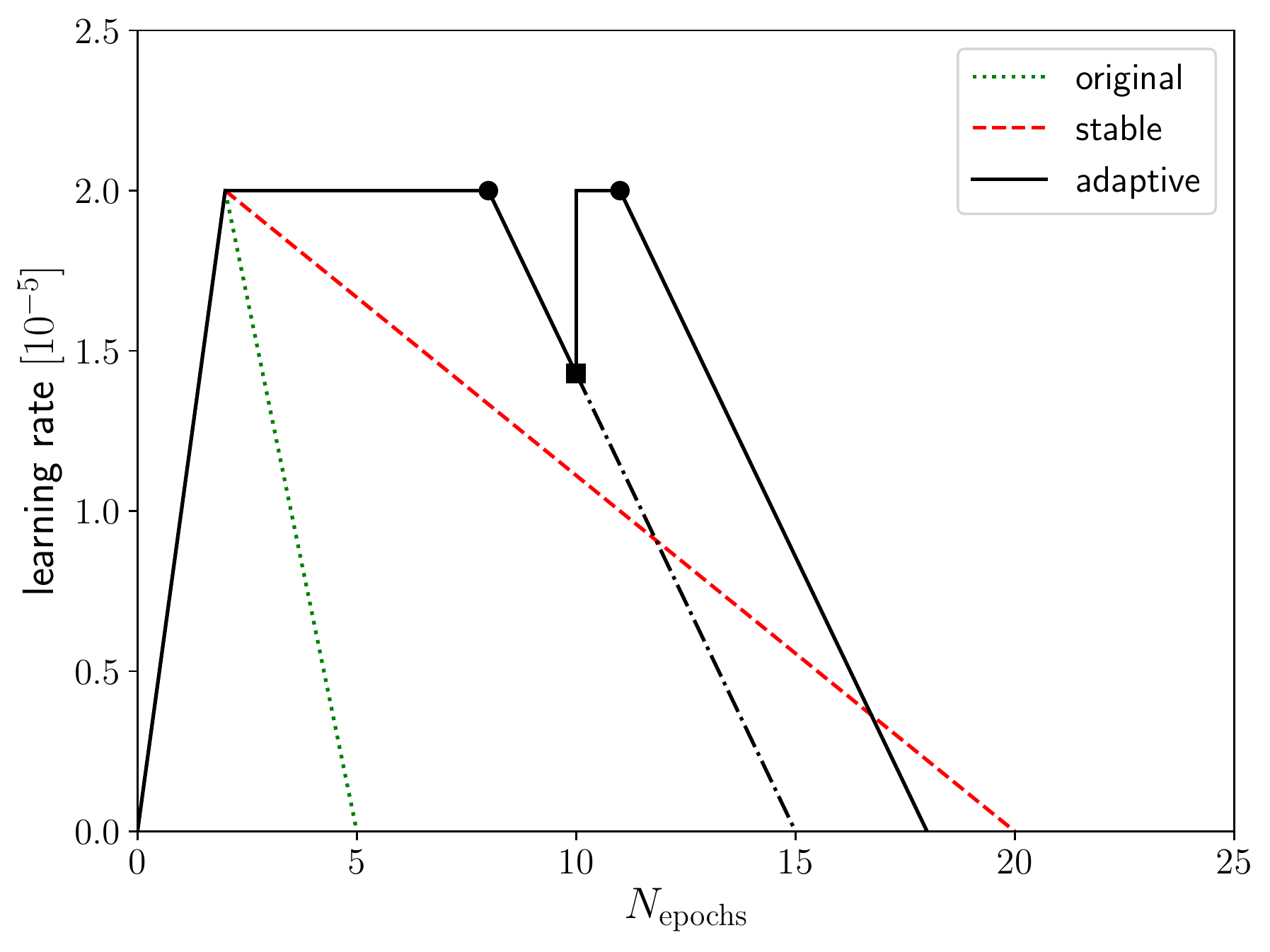}
             }{  
                  \caption{Illustration of the learning rate in different fine-tuning approaches.  The circles ($N_{\rm epochs} = 8, 11$) indicate that early stopping is triggered in the adaptive approach.  If the validation loss improves during the cool-down phase, training with a constant learning rate is resumed (square at $N_{\rm epochs} = 10$). }
 		\label{fig:training_hyperparameters_overview}                         
            }
   \end{floatrow}
\end{figure}

The suitability of a specific number of training epochs depends heavily on the size of the training dataset. 
While the original approach may work well for large datasets, the number of associated training iterations for a smaller dataset often does not suffice for the model to master the downstream task well \citep{zhang2021revisiting, mosbach2021stability}. 
In contrast, the stable approach significantly improves the performance on smaller datasets while the performance on large datasets is not impaired \citep{mosbach2021stability}.  A downside, as the authors point out themselves, is however that the training for large datasets may run for more epochs and thus use more computational resources than necessary. 
We infer from this that the choice of $N_{\rm epochs}$ is a trade-off between performance and stability on the one side and efficiency on the other side, see Fig.~\ref{fig:approaches_principle}. 

The optimal $N_{\rm epochs}$ thus depends on the dataset and its size and can only be found using hyperparameter search \citep{zhang2021revisiting}.  Simply setting $N_{\rm epochs}$ to a high number like in the stable approach may be a useful alternative if one has a sufficiently large dataset and computational resources are abundant \citep{mosbach2021stability}. 
However, while most publicly available benchmark datasets are rather large,  one often faces significantly smaller datasets in industrial applications, where the data may have to be expensively generated using manual annotation.  Below a certain dataset size threshold,  even the stable variant of the fixed epoch approach is suboptimal, as we will see below.

%% file: tex/3_finetuning_approach.tex
\subsection{Approach
\label{subsec:adaptive_approach}} 

In this work, we propose an \textit{adaptive} fine-tuning approach that automatically adjusts the number of training epochs to the dataset size using early stopping.
We start with a linear warm-up for two epochs like the fixed epoch approach. Subsequently,  training is continued using a constant learning rate.  After each epoch, we evaluate the loss on the validation dataset. Any decrease in validation loss is considered an improvement.  Any increase indicates that we are somewhat close to the aspired optimization minimum, and triggers a custom early stopping process.
We continue training for another 
\begin{equation}
N_{\rm patience} = 7
\label{eq:Npatience}
\end{equation}
epochs, while letting the learning rate decrease linearly to zero (similar to the fixed epoch approach) in order to fine-adjust the model parameters. 
The validation loss is continuously monitored during this \textit{cool-down phase}. If we observe a new best value, the early stopping process is stopped and training with a constant learning rate is resumed.
This procedure is crucial as there is always the possibility to encounter a statistical fluctuation that leads to a temporary increase of the validation loss even before the training is actually near convergence, especially for small datasets.
If there is no reoccurrence of validation loss improvement, the training is stopped at the end of the cool-down phase.

The \textit{hybrid} learning rate schedule of our \textit{adaptive} approach is compared to the other methods in Fig.~\ref{fig:training_hyperparameters_overview}. 
We emphasize that it---just like the fixed epoch fine-tuning approach---requires only a single parameter (see Eq.~(\ref{eq:Npatience})) to be set. 
However, monitoring the validation loss allows us to automatically find a reasonable number of training epochs without having to specify it beforehand, irrespective of the training dataset size.  
Our choice of $N_{\rm patience}$ is motivated in App.~\ref{app:escd_variations}, where different variants of our adaptive approach are studied.
Note that apart from the learning rate schedule and the number of training epochs, we employ the same hyperparameters as the fixed epoch approach.   

%% file: tex/3_finetuning_computational_effort.tex
\subsection{Computational Effort
\label{subsec:adaptive_computational_effort}} 

As mentioned in Sec.~\ref{subsec:background_fine_tuning},  the \textit{fixed epoch} approach is often employed multiple times in connection with a simple grid search in hyperparameter space (which in our case is 1-dimensional).  In contrast, the \textit{adaptive} approach requires only a single run to determine the optimal number of training epochs. However, in order to compare the computational effort of both approaches, we need to take into account that adaptive fine-tuning relies on the computation of the validation loss after each epoch. 

In App.~\ref{app:finetuning_computational_effort}, we show that the \textit{relative computational effort} ($R^c$) of the adaptive approach ($C^{\rm adap;c}_{\rm total}$) with respect to a single fixed epoch hyperparameter run ($C^{\rm fixed}_{\rm total}$) can be estimated as follows:
\begin{eqnarray}
 R^c
 &:=& \frac{C^{{\rm adap;}c}_{\rm total}}{C^{\rm fixed}_{\rm total}}
 \approx \frac{N_{\rm epochs}^{{\rm adap;}c}}{N_{\rm epochs}^{\rm fixed}} \cdot \alpha^c
 \qquad \text{with} \qquad 
 \alpha^c = 1 + \frac{N_{\rm val}^c}{2~N_{\rm train}^c}  \ ,
 \label{eq:computational_effort_adaptive_vs_fixed}
\end{eqnarray}
where $c$ is an index denoting the dataset that is considered, and $N_{\rm train}^c$ ($N_{\rm val}^c$) is the size of its training (validation) dataset. 
Note that in Eq.~(\ref{eq:computational_effort_adaptive_vs_fixed}), the ratio of the number of epochs amount to the pure training effort, while $\alpha^c$ is the correction due to the additional validation effort in the adaptive approach.

In order to take into account a multitude of $N_{\rm hyper}$ fixed epoch hyperparameter runs, we simply need to add up the computational effort of the single runs, i.e. replace
\begin{subequations}
\begin{align}
 C^{\rm fixed}_{\rm total} &\to \sum_{i=1}^{N_{\rm hyper}} C^{{\rm fixed;}i}_{\rm total} \\
 N_{\rm epochs}^{\rm fixed} &\to \sum_{i=1}^{N_{\rm hyper}} N^{{\rm fixed;}i}_{\rm epochs}
\end{align}
 \label{eq:computational_effort_adaptive_vs_fixed_multiple}
\end{subequations}
in Eq.~(\ref{eq:computational_effort_adaptive_vs_fixed}).

%% file: tex/4_experiments_introduction.tex
\subsection{Setups \label{subsec:experiments_setups}}

Our goal is to compare the adaptive approach (Sec.~\ref{sec:adaptive}) to the fixed epoch approach (Sec.~\ref{subsec:background_fine_tuning}).  
For this purpose, we conduct experiments where we use these different methods to fine-tune a pretrained transformer-based language model on a dataset annotated for named entity recognition. 
In order to ensure generalizability, we consider various dataset-model combinations, listed in Tab.~\ref{tab:experiments_overview}. 
\begin{table}[t]
\begin{minipage}{\linewidth}
 \centering
  \floatbox[{\capbeside\thisfloatsetup{capbesideposition={left, center},capbesidewidth=4.5cm}}]{table}[\FBwidth]
{
\scalebox{\tableScale}{
\hspace{1mm}
\begin{tabular}{c|c|c|c|c|c|c}
          & \multicolumn{4}{c|}{dataset} & \multicolumn{2}{c}{model}  \\ 
  $c$ & language & name & $\alpha^c$ & ref.  & name & ref.  \\ \hline
 \multirow{2}{*}{I} & \multirow{2}{*}{English} & \multirow{2}{*}{CoNLL-2003} & \multirow{2}{*}{1.12} & \multirow{2}{*}{\citep{tjong-kim-sang-de-meulder-2003-introduction}} & \multirow{2}{*}{\texttt{bert-large-cased}} & \multirow{2}{*}{\citep{devlin2019bert}} \\ 
& & & & & & \\
  \multirow{2}{*}{II} & \multirow{2}{*}{English} & \multirow{2}{*}{CoNLL-2003} & \multirow{2}{*}{1.12}  & \multirow{2}{*}{\citep{tjong-kim-sang-de-meulder-2003-introduction}} & \texttt{distilbert-base-} & \multirow{2}{*}{\citep{Sanh2019DistilBERTAD}} \\ 
  & & & & & \texttt{multilingual-cased} & \\ \hline
 \multirow{2}{*}{III} & \multirow{2}{*}{Swedish} & Swedish & \multirow{2}{*}{1.21} & \multirow{2}{*}{\citep{swedish-ner-corpus}} & \multirow{2}{*}{\texttt{KB/bert-base-swedish-cased}} & \multirow{2}{*}{\citep{swedish-bert}} \\
 & & NER Corpus & & & & \\
 \multirow{2}{*}{IV} & \multirow{2}{*}{Swedish} & \multirow{2}{*}{Swe-NERC} & \multirow{2}{*}{1.06} & \multirow{2}{*}{\citep{swe_nerc}} & \multirow{2}{*}{\texttt{KB/bert-base-swedish-cased}} & \multirow{2}{*}{\citep{swedish-bert}} \\
  & & & & & & \\ \hline
 \multirow{2}{*}{V} & \multirow{2}{*}{Spanish} & \multirow{2}{*}{eHealth-KD 2020} & \multirow{2}{*}{1.12} & \multirow{2}{*}{\citep{overview_ehealthkd2020}} & \texttt{mrm8488/electricidad-} & \multirow{2}{*}{\citep{transformers-electricidad}} \\
  & & & & & \texttt{base-discriminator} & \\
 \end{tabular}
}
}
{
 \caption{Overview of models and datasets we use in our experiments.  We will refer to the different combinations by their dataset-model combination number $c$.  The size of the datasets in terms of the number of sentences are given in Fig.~\ref{fig:dataset_sizes}.  The column $\alpha^c$ denotes the ratio of the validation dataset size and the training dataset size, see Eq.~(\ref{eq:computational_effort_adaptive_vs_fixed}).}
 \label{tab:experiments_overview}
}
\end{minipage} \\ \vspace{3mm}
\begin{minipage}{\linewidth}
 \centering
 \floatbox[{\capbeside\thisfloatsetup{capbesideposition={left, center},capbesidewidth=4.5cm}}]{figure}[\FBwidth]
 {\hspace{0mm} \includegraphics[scale=0.41]{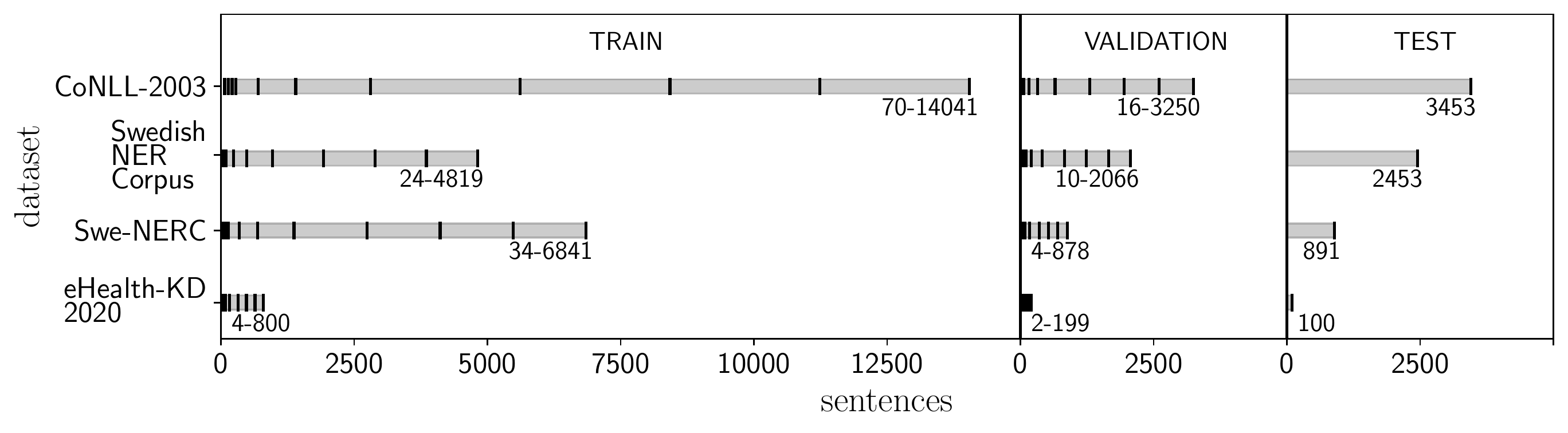} \hspace{0mm}}
 {
 \caption{Sizes of the different training, validation and test datasets. The respective number of sentences are given explicitly next to the gray bars. The black vertical lines correspond to $x_{\rm train}$,  $x_{\rm val}$ and $x_{\rm test}$, see Eq.~(\ref{eq:xtrain_xval}), (\ref{eq:xtest}), (\ref{eq:parameter_x}).}
 \label{fig:dataset_sizes}
}
\end{minipage}
\end{table}
The datasets all employ the \texttt{BIO} annotation scheme \citep{ramshaw1995text}, and each of them consists of a training, validation and test dataset\footnote{The Swedish NER Corpus dataset originally comes in a different scheme and only contains a training and test dataset.  We convert it to the \texttt{BIO} scheme and split off a validation dataset from the training dataset such that it is roughly of the same size as the test dataset. The Swe-NERC dataset originally comes as a single dataset with instructions on how to sample a validation dataset \citep{swe_nerc}. We apply this technique twice to sample both a validation and a test dataset. The remainder is used for training.}. 
As the adaptive approach is meant to work well for \textit{any} dataset size, we also simulate various dataset sizes by scaling down\footnote{A dataset is scaled down by randomly sampling a subset of the wanted size to ensure that the subset represents the original dataset as well as possible.} the original training and validation datasets simultaneously by a scaling factor $x \in [0, 1]$:
\begin{equation}
 x_{\rm train} = x_{\rm val} = x
 \label{eq:xtrain_xval}
\end{equation}
In a realistic scenario, the test datasets would be scaled down accordingly. 
However, we opt to leave them unchanged,
\begin{equation}
 x_{\rm test} = 1
 \label{eq:xtest}
\end{equation}
as this allows us to compare the generalization capacity of a model using the same, best possible estimator irrespective of the scaling factor $x$. 
The sizes of the different datasets using different scaling factors are visualized in Fig.~\ref{fig:dataset_sizes}.
We define an experiment by the following three parameters: 
\begin{subequations}
\begin{align}
  \notag \\
 &\textit{fine-tuning approach:} &&a \in A = \{ \text{original, stable, adaptive} \} 
 \label{eq:parameter_a} \\
 &\textit{dataset-model combination:} &&c \in C = \{ \text{I, II, III, IV, V} \} 
 \label{eq:parameter_c} \\
 &\textit{dataset scaling factor:} &&x \in X = \{ 0.005, 0.01, 0.015, 0.02, 0.05, 0.1, 0.2, 0.4, 0.6, 0.8, 1.0 \} 
 \label{eq:parameter_x} 
 \\ \notag
\end{align}
\end{subequations}
Each experiment consists of 
\begin{align}
 N_{\rm runs} = 5
 \label{eq:Nruns}
\end{align}
runs with different random seeds\footnote{For each experiment, the seeds 43-47 are used.}, and the hyperparameters specified in App.~\ref{app:hyperparameters} are used. 
We conduct experiments for all combinations $(a, c, x)$ on a machine with 1 TITAN RTX GPU (24GB RAM). 
The software that is used is our \texttt{nerblackbox} package \citep{nerblackbox}. In App.~\ref{app:experiments_confirmation}, we check its validity by reproducing literature results.

%% file: tex/4_experiments_analysis.tex
\subsection{Analysis \label{subsec:experiments_analysis}}

We discuss how to evaluate and compare the different experiments in terms of performance, stability and efficiency.
The main quantity of interest with regard to \textit{performance} is the micro-averaged $f_1$ score on the entity level, which we evaluate on the test dataset (see App.~\ref{sec:model_evaluation} for details).  As a proxy for \textit{efficiency} and computational effort,  we consider the number $N_{\rm epochs}$ of training epochs as it is directly measurable as well as comparable across experiments. 
For each experiment $(a, c, x)$ and both metrics $m \in \{ f_1, N_{\rm epochs} \}$, we compute the mean $\overline{m}$ and standard deviation $\sigma_{m}$ of the single run results $m = [m_1, \ldots, m_{N_{\rm runs}}]$ to get an estimate $\widehat{m}$ for the (unknown) real mean:
\begin{equation}
 \widehat{m} = \overline{m} \pm \Delta m \qquad \text{with} \qquad \Delta m = \frac{\sigma_{m}}{\sqrt{N_{\rm runs}}}
 \label{eq:mean_uncertainty}
\end{equation}
Following common practice, we will use the notation $m$ instead of $\widehat{m}$ in the following.
Note that $\Delta m$ is proportional to $\sigma_{m}$, which is widely used in the literature (e.g.~\citep{mosbach2021stability, dodge2020finetuning}). 
Numerical results for $\Delta m$ will be specified using the notation explained in App.~\ref{app:notation_uncertainties}.
In order to assess the relative performance and efficiency of the adaptive approach with respect to the fixed epoch approach variants, $\rm a^\prime \in \{ \text{original}, \text{stable} \}$, we compute
\begin{subequations}
\begin{eqnarray}
 f_1^{\rm adap/a^\prime} &=& f_1^{\rm adap} / f_1^{\rm a^\prime} \label{eq:f1ratio} \\
 N_{\rm epochs}^{\rm adap/a^\prime} &=& N_{\rm epochs}^{\rm adap} / N_{\rm epochs}^{\rm a^\prime} 
 \label{eq:epochsratio}
\end{eqnarray}%
\label{eq:f1epochsratio}%
\end{subequations}
The associated uncertainties of these ratios are derived using Gaussian error propagation (see App.~\ref{app:notation_uncertainties}).
In addition to the ratio of $f_1$, Eq.~(\ref{eq:f1ratio}),  we follow \citep{mosbach2021stability} and also consider the maximum value of $f_1$ to evaluate the performance, as the corresponding model is usually used for inference.
From the ratio of $N_{\rm epochs}$, Eq.~(\ref{eq:epochsratio}),  we can derive the relative computational effort $R^c$, Eq.~(\ref{eq:computational_effort_adaptive_vs_fixed}), which is larger by up to $21\%$ depending on the dataset $c \in C$ (see Tab.~\ref{tab:experiments_overview}).  
As outlined in Sec.~\ref{subsec:background_stability}, an approach is said to be more efficient if it achieves the same results with less computational effort.  More specifically,  we use the term \textit{efficiency} (of the adaptive approach with respect to the fixed epoch approach) synonymously with $R^c$ wherever we find $f_1^{\rm adap/a^\prime} \approx 1$.

To assess \textit{stability}, we first consider the single experiments $(a, c, x)$ and count the number of converged runs for each of them. 
Strictly speaking, convergence is defined in terms of the loss function.  However, a convergent run usually results in an $f_1$ score on the test dataset that is greater than zero, while a divergent run leads to zero.  For the sake of simplicity, we use this property to distinguish convergence from divergence, and define the number of converged runs as
\begin{equation}
 \text{cv}(a, c,x) := \sum_{n=1}^{N_{\rm runs}} \mathds{1}_{f_1^{(n)} (a, c,x) > 0}
 \label{eq:stability_estimator_convergence}
\end{equation} 
where $f_1^{(n)}$ is the $f_1$ score of the $n$-th run.
On the subset of converged runs, the computed relative uncertainties $\Delta f_1 / f_1$ provide information about the stability of the model training. 
As they can depend strongly on the employed dataset size (cf.~Sec.~\ref{subsec:background_stability}), 
we compute the \textit{average relative uncertainty}
\begin{equation}
 u^{a}(c) := \frac{1}{| \widetilde X |} \sum_{x \in \widetilde X} \frac{\Delta f_1(a,c,x)}{f_1(a,c,x)}
 \label{eq:stability_estimator}
\end{equation} 
for each set of experiments with $(a,c)$, where we average over all $x \in \widetilde X \equiv \widetilde X (c) := \{x \in X ~|~ x \geq \widetilde x(c)\}$ that are greater than or equal some thresholds $\widetilde x(c)$.  These thresholds serve two purposes. Firstly,  they are to ensure that all of the runs of the experiments that are taken into account in the sum above converge. This is typically the case once a certain training dataset size is exceeded, and could be ensured by thresholds $\widetilde x \equiv \widetilde x(a, c)$ that are specific for every combination $(a, c)$. However, secondly, our goal is to compare different fine-tuning approaches. In order to have a fair comparison, one needs to employ the same thresholds $\widetilde x \equiv \widetilde x(c)$ for every $a \in A$, due to the aforementioned dependence of $\Delta f_1 / f_1$ on $x$. 
Hence, we define the thresholds such that \textit{all} experiments $(a, c, x)$ converge for $x \geq \widetilde x(c)$:
\begin{equation}
 \widetilde x(c) := \min \left\{ x \in X ~|~ \text{cv}(a, c,x) = N_{\rm runs} ~~\forall~ a \in A \right\}
\label{eq:stability_estimator_threshold}
\end{equation}
The average relative uncertainty $u^{a}(c)$ may be averaged again over all $c$ to get the \textit{global average relative uncertainty}
\begin{equation}
 \overline{u}^{a} := \frac{1}{\left| C \right|} \sum_{c \in C} u^{a}(c)
 \label{eq:global_stability_estimator}
\end{equation} 
for a specific approach $a \in A$.

%% file: tex/5_results_finetuning.tex
For the sake of brevity, we restrict ourselves to the dataset-model combination $c = \text{I}$ here.  The other dataset-model combinations $c \in C$ are discussed in App.~\ref{app:finetuning_generalization}, showing that the findings hold generally. 

\begin{figure}[t]
\centering
\scalebox{\tableScale}{
\begin{tabular}{c|c|c|c}
  $c$ & $x$ & $f_1^{\rm adap./stab.}$ & $N_{\rm epochs}^{\rm adap./stab.}$  \\ \hline
   \input{analysis/tex/5_app_F_mean_std_ratio_stable_I.tex}
 \end{tabular}
}
    \qquad
    \begin{minipage}{0.45\textwidth}
     \includegraphics[scale=\figScale]{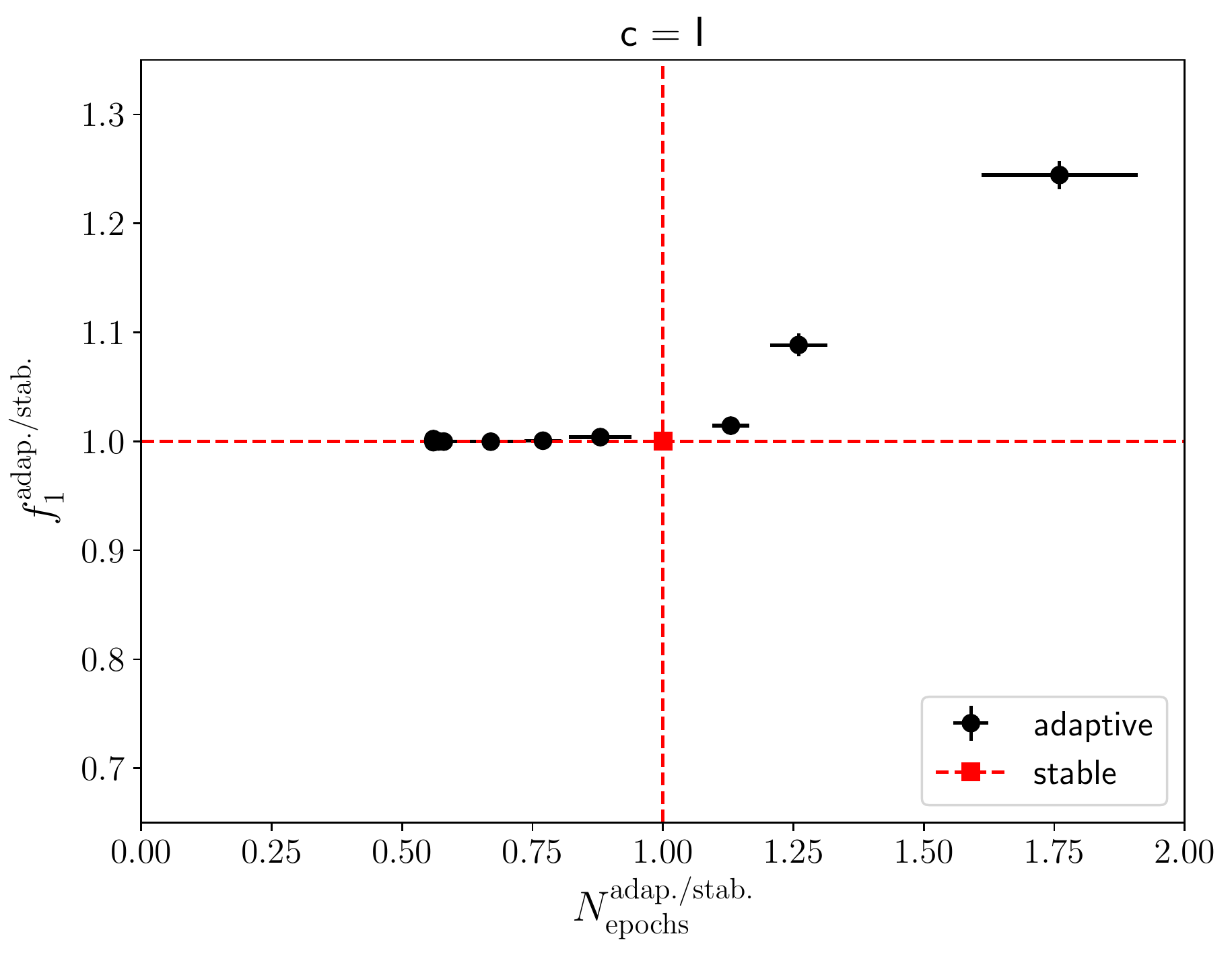}
    \end{minipage}
 \\ \vspace{2mm}
\scalebox{\tableScale}{
\begin{tabular}{c|c|c|c}
  $c$ & $x$ & $f_1^{\rm adap./orig.}$ & $N_{\rm epochs}^{\rm adap./orig.}$  \\ \hline
   \input{analysis/tex/5_app_F_mean_std_ratio_original_I.tex}
 \end{tabular}
}
 \qquad
  \begin{minipage}{0.45\textwidth}
     \includegraphics[scale=\figScale]{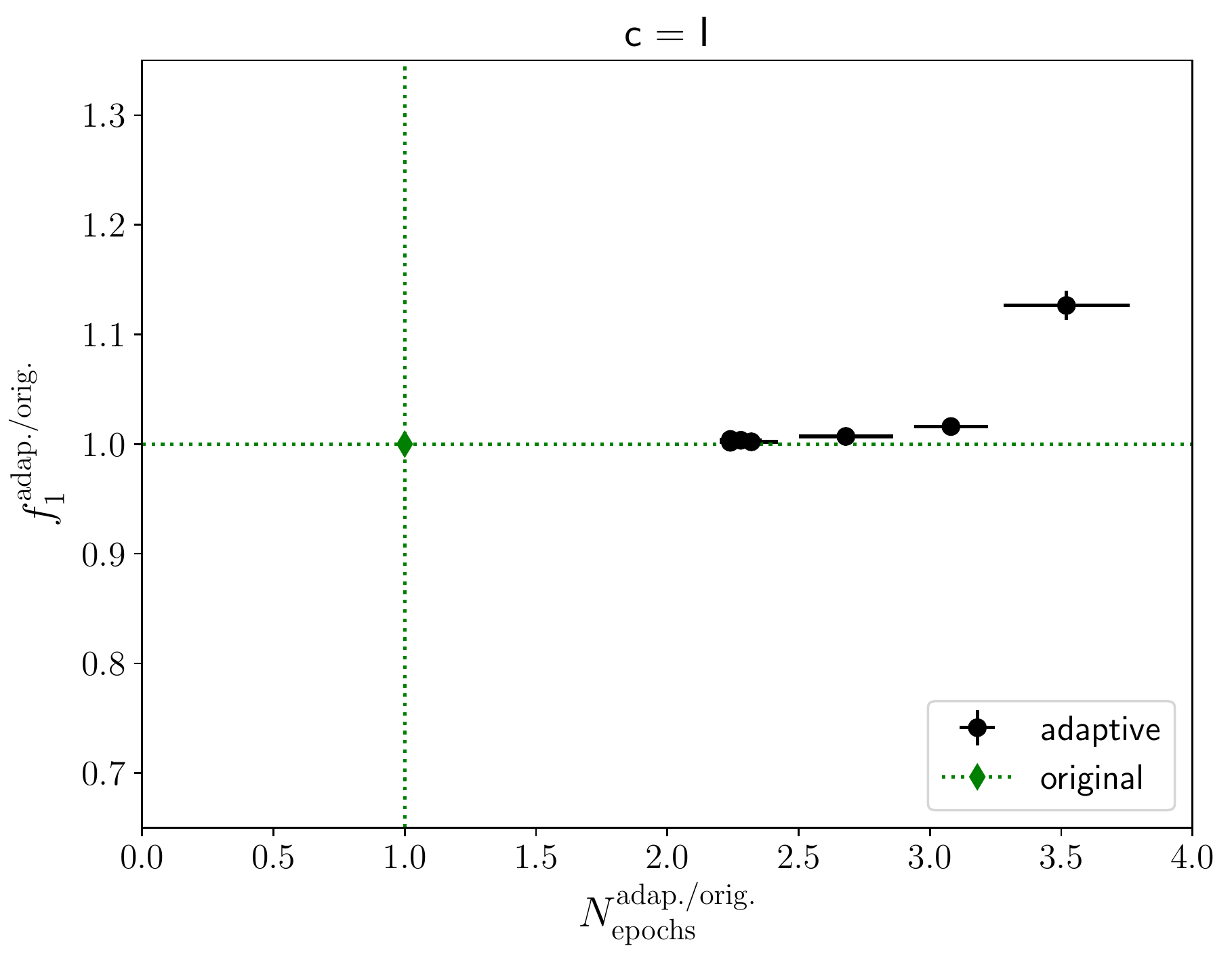}
  \end{minipage}
  \caption{Relative peformance $f_1^{\rm adap./a^\prime}$ (cf.~Eq.~(\ref{eq:f1ratio})) against the epoch ratio $N_{\rm epochs}^{\rm adap./a^\prime}$ (cf.~Eq.~(\ref{eq:epochsratio})) of the adaptive fine-tuning approach with respect to the other approaches $a^\prime \in \{ \text{original, stable} \}$, for different dataset scaling factors $x$ and the dataset-model combination $c = \text{I}$.  In the plots, the center point (1,1) corresponds to the performance and number of epochs in the \textit{stable} (top) and \textit{original} (bottom) strategy, and the horizontal (vertical) lines represents the same performance (epochs).  The other points show the relative performance gain and epochs ratio of our \textit{adaptive} approach, for different dataset scaling factors $x$.  The relative computational cost $R^c$ approximately equals the epoch ratio times a factor $\alpha^\text{I} = 1.12$ (cf.~Eq.~(\ref{eq:computational_effort_adaptive_vs_fixed}) and Tab.~\ref{tab:experiments_overview}). Note that the points where the performance gain is highest (for small $x$) are outside the plot range. For $x \leq 0.01$, the relative performance $f_1^{\rm adap./orig.}$ is undefined as the \textit{original} fine-tuning runs do not converge (cf.~Tab.~\ref{tab:experiments_optimization}).}
 \label{tab:result_training_epochs_ratio}
\end{figure}

The results for the different fine-tuning approaches $a \in A$ and dataset scaling factors $x \in X$ can be found in Tab.~\ref{tab:experiments_optimization}. 
The number of epochs after which the training stopped and the $f_1$ score are also visualized in Fig.~\ref{fig:result_training_epochs}. 
\begin{table}[t]
\begin{minipage}{\linewidth}
 \centering
  \floatbox[{\capbeside\thisfloatsetup{capbesideposition={left, center},capbesidewidth=6.2cm}}]{table}[\FBwidth]
{
\scalebox{\tableScale}{
  \begin{tabular}{c|c|c|c|c|c|c|c|c}
  & & \multicolumn{2}{c|}{original} & \multicolumn{2}{c|}{stable} & \multicolumn{3}{c}{adaptive} \\ \cline{3-9}
  \multirow{2}{*}{$c$} & \multirow{2}{*}{$x$} & \multirow{2}{*}{cv} & \multirow{2}{*}{$f_1$} & \multirow{2}{*}{cv} & \multirow{2}{*}{$f_1$} & \multirow{2}{*}{cv} & \multirow{2}{*}{$f_1$} & \multirow{2}{*}{$N_{\rm epochs}$} \\ 
  & & & & & & & \\ \hline
   \input{analysis/tex/5_app_F_mean_std_I.tex}
 \end{tabular}
}
}
{
 \caption{Comparison of fine-tuning results on the dataset-model combination $c = \text{I}$ for different fine-tuning approach $a \in A$ and dataset scaling factors $x \in X$. The $f_1$ scores are evaluated on the entity level and the test dataset. The number of training epochs for the original and stable strategy are fixed to 5 and 20, respectively.  The numbers in the cv columns denote the number of runs that converged, where a run is considered converged if the micro-averaged $f_1$ score is greater than 0.  If no number is given, all 5 runs converged and the corresponding mean $f_1$ score is specified.  Highlighted values are significantly better than the other $f_1$ scores in the same row (cf.~App.~\ref{app:notation_uncertainties}). 
}
 \label{tab:experiments_optimization}
}
\end{minipage} \\ \vspace{3mm}
\begin{minipage}{\linewidth}
 \centering
 \floatbox[{\capbeside\thisfloatsetup{capbesideposition={left, center},capbesidewidth=3.4cm}}]{figure}[\FBwidth]
 {\includegraphics[scale=\figScale]{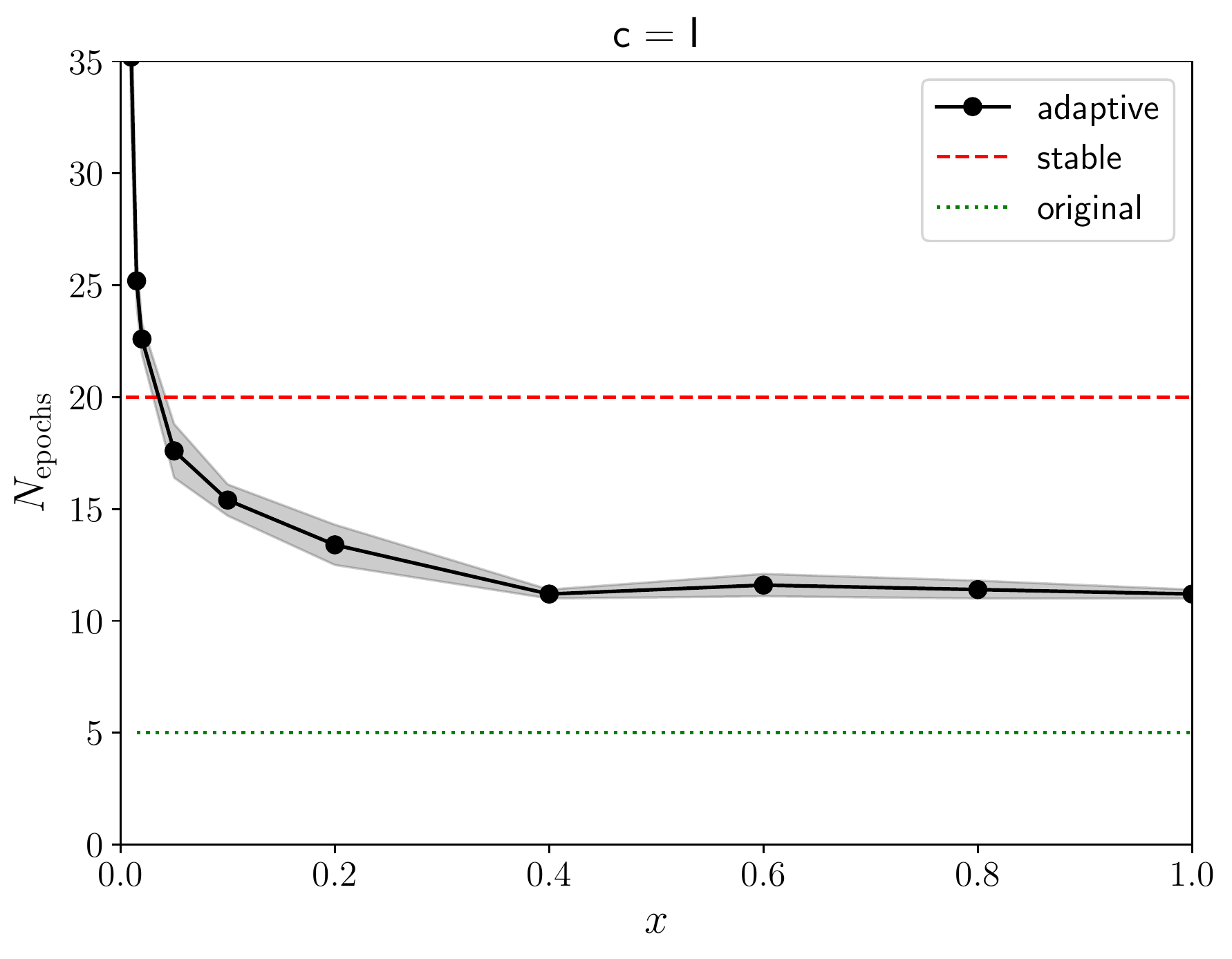}
 \includegraphics[scale=\figScale]{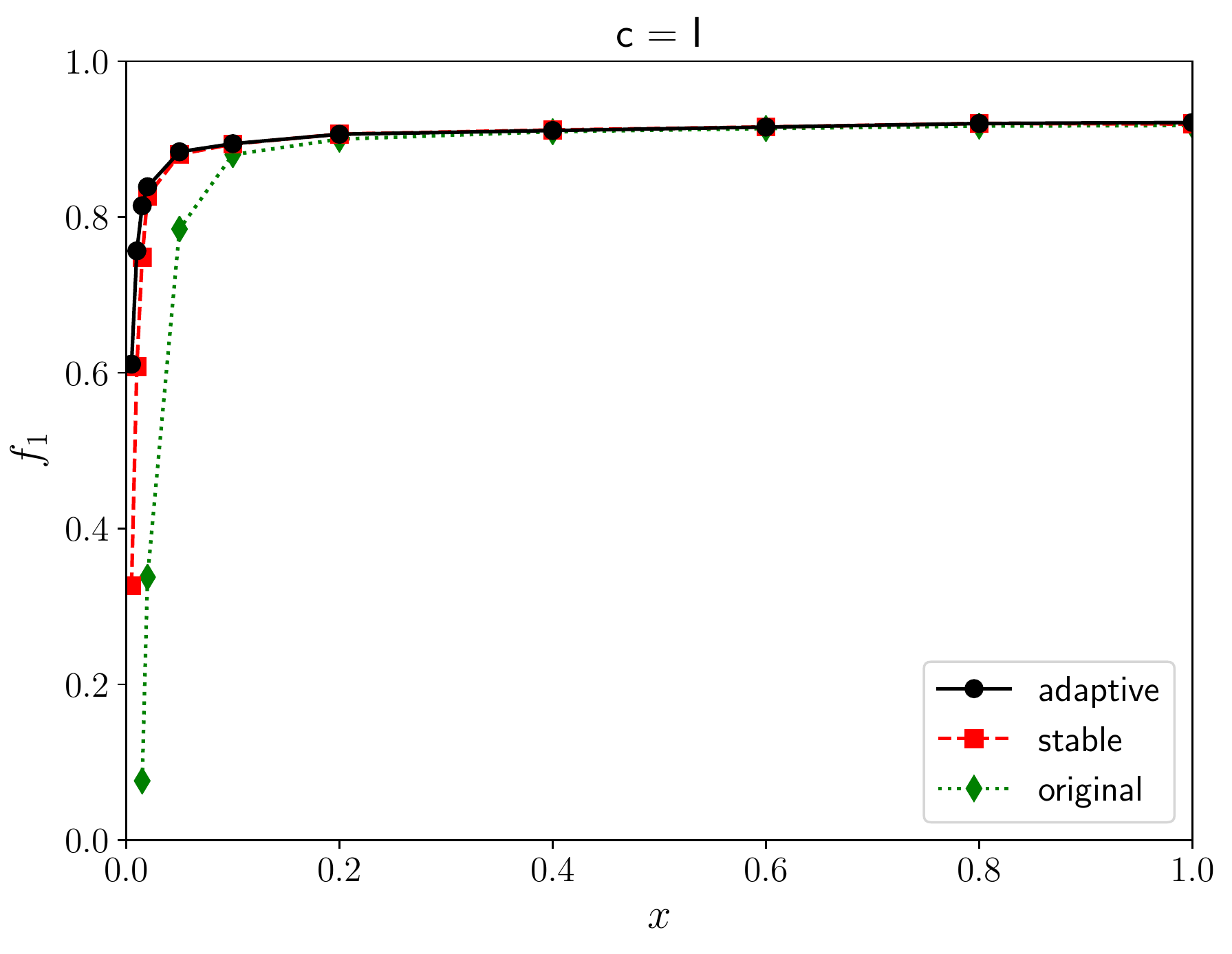}}
 {
 \caption{Results for different fine-tuning approaches on the dataset-model combination $c=\text{I}$.  \textit{Left panel:} Number $N_{\rm epochs}$ of training epochs as a function of the dataset scaling factor $x$.  \textit{Right panel:} $f_1$ score on the test dataset as a function of the dataset scaling factor  $x$. }
 \label{fig:result_training_epochs}
 }
\end{minipage}
\end{table}
We observe that training with the adaptive approach requires more iterations than the other approaches for very small datasets. However, the number of epochs decreases rapidly as the dataset size increases and lies quite far below the 20 epochs used in the stable approach for most dataset sizes.  More importantly, the performance of the adaptive approach is much better than the fixed epoch approach for small datasets ($x \leq 0.05$), and very similar to the stable variant for larger datasets ($x > 0.05$). 
Fig.~\ref{tab:result_training_epochs_ratio} lists the ratios $f_1^{\rm adap./a^\prime}$ and $N_{\rm epochs}^{\rm adap./a^\prime}$ (see Eq.~(\ref{eq:f1epochsratio})) and plots them in combination, for each $a^\prime \in \{ \text{original, stable} \}$.
Throughout all experiments, we find
\begin{subequations}
\begin{align}
 f_1^{\rm adap./orig.} > 1 \qquad
\end{align}
and
\begin{align}
 f_1^{\rm adap./stab.} > 1 \qquad 
\text{or} \qquad 
\bigg( f_1^{\rm adap./stab.} \approx 1 \quad \text{and} \quad N_{\rm epochs}^{\rm adap./stab.} < 1 \bigg)
\label{eq:result_performance_adaptive}
\end{align}%
\label{eq:result_performance}%
\end{subequations}
where the two cases in Eq.~(\ref{eq:result_performance_adaptive}) correspond to small and large datasets, respectively.
The maximum values of the $f_1$ score, listed in 
App.~\ref{app:finetuning_generalization} (Tab.~\ref{tab:experiments_optimization_maximum_ALL}),
display a very similar behavior.
We conclude that our approach outmatches the other approaches in terms of \textit{performance} and \textit{efficiency}: it either leads to higher performance (in particular on smaller datasets) or to competitive performance but higher efficiency (on larger datasets compared to the stable approach). 
The latter holds true even if one takes into account the additional computational effort corresponding to the validation steps in the adaptive approach (cf.~Sec.~\ref{subsec:adaptive_computational_effort}), which amounts to $12\%$ (see Tab.~\ref{tab:experiments_overview}).
Moreover, as pointed out in Sec.~\ref{subsec:adaptive_computational_effort}, there is no need for multiple hyperparameter runs with respect to the number of training epochs if the adaptive approach is used, which further contributes to an improved efficiency.

To assess \textit{stability}, we first note that the original approach does not converge for very small datasets (cf.~Tab.~\ref{tab:experiments_optimization}).  The threshold above which the experiments for all approaches converge (see Eq.~(\ref{eq:stability_estimator_threshold})) is $\widetilde x(\text{I}) = 0.015$. For the average relative uncertainties, we find 
\begin{equation}
 u^{a}(\text{I}) = 
 \begin{cases}
  0.0494 & \text{for } a = \text{original} \\
  0.0025 & \text{for } a = \text{stable} \\
  0.0012 & \text{for } a = \text{adaptive} \\
 \end{cases}
 \label{eq:stability_estimator_results}
\end{equation}
The results show that for small dataset scaling factors $x$, the stable and adaptive approach---while there is no large difference between the two of them---both outperform the original approach in terms of stability.

The appendices contain important additional studies. 
Most notably, in App.~\ref{app:finetuning_generalization},
we display the results for all dataset-model combinations $c \in C$, suggesting that the findings hold generally across datasets of various sizes and irrespective of the pretrained model architecture. 
In App.~\ref{app:escd_variations}, some variants of the adaptive approach are studied, motivating the choices we made in Sec.~\ref{sec:adaptive}.  In particular, we find that the effect of varying $N_{\rm patience}$ (cf.~Eq.~(\ref{eq:Npatience})) is small and somewhat inexplicit. This is an important result as it not only confirms that adaptive fine-tuning automatically uses a close-to-optimal $N_{\rm epochs}$, but also implies that the need to adjust $N_{\rm patience}$ is very low.  Consequently,  in practice, there is one hyperparameter less to tune manually.
Moreover, in App.~\ref{app:finetuning_ablation}, we study the different features of the adaptive approach that make it distinct from the fixed epoch fine-tuning approach, in order to gain an understanding of their individual contributions.
In App.~\ref{app:finetuning_additional}, we discuss the learning curves of the validation loss for the different approaches. 
App.~\ref{app:finetuning_dependency} investigates how the model performance depends on $x_{\rm train}$ and $N_{\rm epochs}$ and provides further insights on why adaptive fine-tuning works well.  
Finally, we cover two topics that have important implications in practice.  App.~\ref{app:finetuning_dependency_validation} shows that adaptive fine-tuning is very robust in the sense that the results hardly depend on the size of the validation dataset.  App.~\ref{sec:adaptive_finetuning_in_practice} demonstrates that the approach is also useful in the case where one chooses to add the validation dataset to the training dataset in a final step to make the most of the available data and push the performance.

%% file: analysis/tex/5_app_F_mean_std_I.tex
\multirow{11}{*}{I}
 & 0.005 & 0 & --- &  & 0.3267(249) &  &{ \bf 0.6112(97) }& 49.4(1.9)  \\
 & 0.01 & 2 & --- &  & 0.6080(87) &  &{ \bf 0.7566(44) }& 35.2(3.0)  \\
 & 0.015 &  & 0.0762(256) &  & 0.7483(95) &  &{ \bf 0.8145(12) }& 25.2(1.1)  \\
 & 0.02 &  & 0.3377(295) &  & 0.8270(28) &  &{ \bf 0.8389(28) }& 22.6(7)  \\
 & 0.05 &  & 0.7846(116) &  & 0.8804(17) &  &{ \bf 0.8838(14) }& 17.6(1.2)  \\
 & 0.1 &  & 0.8801(22) &  & 0.8937(12) &  & 0.8942(1) & 15.4(7)  \\
 & 0.2 &  & 0.9000(8) &  & 0.9066(2) &  & 0.9063(4) & 13.4(9)  \\
 & 0.4 &  & 0.9096(7) &  & 0.9118(11) &  & 0.9111(6) & 11.2(2)  \\
 & 0.6 &  & 0.9136(9) &  & 0.9157(7) &  & 0.9155(9) & 11.6(5)  \\
 & 0.8 &  & 0.9169(5) &  & 0.9202(4) &  & 0.9201(4) & 11.4(4)  \\
 & 1.0 &  & 0.9175(5) &  & 0.9195(7) &  &{ \bf 0.9214(8) }& 11.2(2)

%% file: analysis/tex/5_app_F_mean_std_ratio_stable_I.tex
\multirow{11}{*}{I}
 & 0.005 & 1.871(55) & 2.47(1)  \\
 & 0.01 & 1.244(13) & 1.76(15)  \\
 & 0.015 & 1.088(1) & 1.26(6)  \\
 & 0.02 & 1.014(4) & 1.13(3)  \\
 & 0.05 & 1.004(2) & 0.88(6)  \\
 & 0.1 & 1.001(2) & 0.77(3)  \\
 & 0.2 & 1.000(0) & 0.67(4)  \\
 & 0.4 & 0.999(1) & 0.56(1)  \\
 & 0.6 & 1.000(1) & 0.58(3)  \\
 & 0.8 & 1.000(1) & 0.57(2)  \\
 & 1.0 & 1.002(1) & 0.56(1)

%% file: analysis/tex/5_app_F_mean_std_ratio_original_I.tex
\multirow{11}{*}{I}
 & 0.005 & --- & ---  \\
 & 0.01 & --- & ---  \\
 & 0.015 & 10.689(274) & 5.04(22)  \\
 & 0.02 & 2.484(74) & 4.52(14)  \\
 & 0.05 & 1.126(13) & 3.52(24)  \\
 & 0.1 & 1.016(3) & 3.08(14)  \\
 & 0.2 & 1.007(1) & 2.68(18)  \\
 & 0.4 & 1.002(1) & 2.24(4)  \\
 & 0.6 & 1.002(1) & 2.32(1)  \\
 & 0.8 & 1.003(1) & 2.28(8)  \\
 & 1.0 & 1.004(1) & 2.24(4)

%% file: tex/6_summary.tex
In this paper, we have introduced \textit{adaptive} fine-tuning for transformer-based language models. It is based on a hybrid learning rate schedule in conjunction with monitoring of the validation loss and the use of early stopping, with the goal of training for an optimal amount of epochs.  For the example use case of named entity recognition, we have shown that our approach has two big advantages. Firstly, it makes the hyperparameter search regarding $N_{\rm epochs}$ redundant. Secondly, it is generally applicable for any dataset size and achieves optimal results. In particular, it surpasses the current state-of-the-art \textit{fixed epoch} approach in terms of model performance and training stability on small datasets, while reaching equivalent results more cost-efficiently on large datasets.  

An example use case where these advantages appear at their best is iterative manual annotation, where after each iteration a model is trained on the gathered data to assist the annotators with its predictions.  The very different dataset sizes after each iteration require very different numbers of training epochs.  Adaptive fine-tuning automatically takes care of this, and can be employed throughout the whole process to guarantee that the model training makes the most of the available data.

The adaptive fine-tuning approach is implemented in our \texttt{nerblackbox} package \citep{nerblackbox} for named entity recognition.  The source code can be found under \url{https://github.com/flxst/nerblackbox} and contains a script to reproduce the numerical results of this paper. 

Note that adaptive fine-tuning can be employed for other downstream tasks as well.  We leave it for future work to follow up on this.

\subsubsection*{Acknowledgements}

The author would like to thank Magnus Sahlgren for his helpful comments on an early draft of this paper. 

%% file: tex/appendix_hyperparameters.tex
The training hyperparameters of the different fine-tuning approaches we use are listed in Tab.~\ref{tab:training_hyperparamters_overview}.
\begin{table}[ht!]
 \centering
\scalebox{\tableScale}{
 \begin{tabular}{c|c|c|c|c|c}
 \multicolumn{2}{c|}{hyperparameter} & ~~~~~~~~original~~~~~~~~ & ~~~~~~~~~~stable~~~~~~~~~~ & adaptive & source \\ \hline
 \multirow{2}{*}{data} & batch size & \multicolumn{3}{c|}{16} & \citep{devlin2019bert}, \citep{mosbach2021stability} \\ 
                                           & sequence length & \multicolumn{3}{c|}{128} & \\ \hline
  \multirow{5}{*}{optimizer} & type & \multicolumn{3}{c|}{Adam} & \citep{devlin2019bert}, \citep{mosbach2021stability} \\ 
  & bias correction & \multicolumn{3}{c|}{yes} & \citep{mosbach2021stability} \\ 
  & $\beta_1$ & \multicolumn{3}{c|}{0.9} & \citep{devlin2019bert}, \citep{mosbach2021stability} \\
  & $\beta_2$ & \multicolumn{3}{c|}{0.999} & \citep{devlin2019bert}, \citep{mosbach2021stability} \\
  & weight decay & \multicolumn{3}{c|}{0.01} & \citep{devlin2019bert}, \citep{mosbach2021stability} \\ \hline
  \multirow{3}{*}{learning rate} & maximum & \multicolumn{3}{c|}{2e-5} & \citep{devlin2019bert}, \citep{mosbach2021stability} \\ 
  & warm-up & \multicolumn{3}{c|}{linear} & \citep{devlin2019bert}, \citep{mosbach2021stability} \\
  & \textbf{decay} & \textbf{linear} & \textbf{linear} & \textbf{constant+linear}& \citep{devlin2019bert}, \citep{mosbach2021stability} \\ \hline
  \multirow{3}{*}{training epochs} & warm-up & 2 & 2 & 2 & \citep{mosbach2021stability} \\ 
  & \textbf{decay} & \textbf{3} & \textbf{18} & \textbf{varying} & \citep{devlin2019bert}, \citep{mosbach2021stability} \\ 
  & \textbf{total} & \textbf{5} & \textbf{20} & \textbf{varying} & 
 \end{tabular}
}
 \caption{
Training hyperparameters of the different fine-tuning approaches.  Only the highlighted hyperparameters are varied, the other ones (which except for the sequence length are exactly the BERT hyperparameters from \citep{mosbach2021stability}) are held fixed throughout all experiments. 
 }
 \label{tab:training_hyperparamters_overview}
\end{table}
The \textit{stable} hyperparameters are taken from \citep{mosbach2021stability} and target the BERT architecture.  The hyperparameters used for the other architectures employed in this paper are identical, with a single exception being the weight decay for ELECTRA, which is zero in the original paper \citep{clark2020electra}. For the sake of simplicity, we ignore this difference assuming that it has the same effect on the fine-tuning approaches that we aim to compare.  The \textit{original} hyperparameters are the same as the stable hyperparameters, except for the number of training epochs. They can also be considered a variation of the default hyperparameter combination specified in \citep{devlin2019bert},  where bias correction and a specific warm-up period are employed in order to improve stability at the beginning of training (see \citep{mosbach2021stability}). The adaptive hyperparameters employ a hybrid of a constant and linearly decaying learning rate and a varying number of training epochs, as described in Sec.~\ref{sec:adaptive}.  Note that the models we use (cf. ~Tab.~\ref{tab:experiments_overview}) all have a maximum sequence length of 512.  However, at least $99.7\%$ of the samples of each dataset consist of 128 tokens or less. Hence, we opt to use a sequence length of 128, which significantly reduces the computational effort.

%% file: tex/appendix_uncertainties.tex
We use a shorthand notation where the uncertainty (see Eq.~(\ref{eq:mean_uncertainty})) is specified in parentheses. The figures refer to the last decimals, e.g. 
\begin{equation}
 0.1234(56) \equiv 0.1234 \pm 0.0056
\end{equation}
In case of derived quantities $f(a, b)$, we use Gaussian error propagation to compute the uncertainty $\Delta f(a, b)$, e.g.
\begin{equation}
 f(a, b) = \frac{a}{b} \qquad \Rightarrow \qquad 
\Delta f(a, b) = \sqrt{ 
\bigg( \frac{\Delta a}{b}     \bigg)^2 + 
\bigg( \frac{a \cdot \Delta b}{b^2} \bigg)^2
}
\end{equation}
The difference $a-b$ between two quantities is considered significant (with respect to one standard deviation) if 
\begin{equation}
 \left| a - b \right| > \Delta \left( a - b \right)
\end{equation}
$f_1$ scores (cf.~App.~\ref{sec:model_evaluation}) that are significantly better than others in terms of this definition are often highlighted in bold, see e.g. Tab.~\ref{tab:experiments_optimization}.

%% file: tex/appendix_model_evaluation.tex
Model performance (and its capability to generalize) is assessed on the test dataset.  The standard quantities to use for the evaluation of a model on the NER task are \textit{precision}, \textit{recall} and \textit{$f_1$ score}.  They can be computed both on the \textit{token} and \textit{entity} level.  After those metrics are determined individually for each class, the respective \textit{micro-average} over all classes is often taken to get numbers that quantify the model performance for the whole class system. 

Throughout this work, we consider the \textit{micro-averaged $f_1$ score on the entity level} as a single metric to assess model performance.  In particular, we employ the strict evaluation scheme defined in the \texttt{CoNLL-2003} task \citep{tjong-kim-sang-de-meulder-2003-introduction}, where a multi-token entity is considered correct if and only if each of its tokens is predicted correctly.

%% file: tex/appendix_results_confirmation.tex
This section serves to cross-check that the use of our \texttt{nerblackbox} package \citep{nerblackbox} leads to performances on a par with literature results.  We employ the \textit{original} fine-tuning approach with the hyperparameters listed in App.~\ref{app:hyperparameters}. Our results are shown in Tab.~\ref{tab:experiments_confirmation}.
\begin{table}[ht!]
 \centering
\scalebox{\tableScale}{
 \begin{tabular}{c|c|c|c|c|c}
   & \multicolumn{2}{c|}{val} & \multicolumn{2}{c|}{test} & \\ 
  $c$ & our result & literature & our result & literature & ref.  \\ \hline
\input{analysis/tex/1_confirmation_app_D_mean_std.tex}
 \end{tabular}
}
 \caption{Comparison of our fine-tuning results with literature results. The numbers denote the micro-averaged $f_1$ score on the entity level. We average over 5 identical runs, leading to the uncertainty on the last decimals displayed in parentheses (see App.~\ref{app:notation_uncertainties}). Note that $c = \text{I}^*$ is the same as $c = \text{I}$ but with the base (\texttt{bert-base-cased}) instead of the large (\texttt{bert-large-cased}) English BERT model \citep{devlin2019bert}. The literature value for $c=\text{IV}$ is the result of 6 single runs with different train/test data splits \citep{swe_nerc}, and thus---strictly speaking---not directly comparable to our result that employs a single train/test data split only.}
 \label{tab:experiments_confirmation}
\end{table}
We find that they are in rough agreement with the literature results. The small differences are to be expected as the literature results were created using very similar, but slightly different hyperparameters (without the improvements described in Sec.~\ref{subsec:background_fine_tuning}).

%% file: analysis/tex/1_confirmation_app_D_mean_std.tex
~I* & 0.9507(7) & 0.951 & 0.9149(9)~~ & 0.913 & \citep{transformers-conll2003-bert} \\
II & 0.9454(4) & 0.941 & 0.9028(8)~~ & --- & \citep{transformers-conll2003-distilbert} \\
IV & --- & --- & 0.8127(23) & 0.8180(34) & \citep{swe_nerc}

%% file: tex/appendix_finetuning_computational_effort.tex
In this section, we derive Eq.~(\ref{eq:computational_effort_adaptive_vs_fixed}) for the relative computational effort of the adaptive approach with respect to the fixed epoch approach.
We consider a dataset with $N_{\rm train}$ training samples and $N_{\rm val}$ validation samples, and a model that we train for $N_{\rm epochs}$ epochs.
Let $C_{\rm forward}$ be the computational effort of a forward pass in terms of FLOPs.  We follow the assumption that the computational effort $C_{\rm backward}$ of a backward pass is approximately the same\footnote{Note that other sources use $C_{\rm backward} = 2 \cdot C_{\rm forward}$, see e.g. \citep{kaplan2020scaling, openai}, which leads to a $3$ in the denominator of Eq.~(\ref{eq:computational_effort_proof_3}) and therefore a slightly more optimistic estimate of $R$.} \citep{clark2020electra}:
\begin{eqnarray}
 C_{\rm backward} &\approx& C_{\rm forward}
\label{eq:computational_effort_proof_0}
\end{eqnarray}
Training basically amounts to a forward and backward pass per batch, while validation only requires a forward pass. For simplicity, we assume that the batch size is $N_{\rm batch}$ for both training and validation:
\begin{subequations}
\begin{eqnarray}
 C_{\rm train/epoch} &\approx& \left( C_{\rm forward} + C_{\rm backward} \right) \cdot N_{\rm train} / N_{\rm batch} \\
 C_{\rm val/epoch} &\approx& C_{\rm forward} \cdot N_{\rm val} / N_{\rm batch}
\end{eqnarray}%
\label{eq:computational_effort_proof_1}%
\end{subequations}%
The adaptive approach applies validation after each epoch, while the fixed epoch approach only applies it once at the end of the training process. Hence, the respective total computational efforts are
\begin{subequations}
\begin{eqnarray}
 C^{\rm adap}_{\rm total} &=& C_{\rm train/epoch} \cdot N_{\rm epochs}^{\rm adap} + C_{\rm val/epoch} \cdot N_{\rm epochs}^{\rm adap} \\
 C^{\rm fixed}_{\rm total} &=& C_{\rm train/epoch} \cdot N_{\rm epochs}^{\rm fixed} + C_{\rm val/epoch}
\end{eqnarray}%
\label{eq:computational_effort_proof_2}%
\end{subequations}
Putting it all together, for the ratio of these two quantities we find
\begin{eqnarray}
 R
 &=&  \frac{C^{\rm adap}_{\rm total}}{C^{\rm fixed}_{\rm total}} \notag \\
 &\stackrel{(\ref{eq:computational_effort_proof_0}, \ref{eq:computational_effort_proof_1}, \ref{eq:computational_effort_proof_2})}{\approx}& \frac{2 N_{\rm train} \cdot N_{\rm epochs}^{\rm adap} + N_{\rm val} \cdot N_{\rm epochs}^{\rm adap}}{2 N_{\rm train} \cdot N_{\rm epochs}^{\rm fixed} + N_{\rm val}} \notag \\
  &\lesssim& \frac{2 N_{\rm train} \cdot N_{\rm epochs}^{\rm adap} + N_{\rm val} \cdot N_{\rm epochs}^{\rm adap}}{2 N_{\rm train} \cdot N_{\rm epochs}^{\rm fixed}} \notag \\
 &=& \frac{N_{\rm epochs}^{{\rm adap}}}{N_{\rm epochs}^{\rm fixed}} \cdot \left( 1 + \frac{N_{\rm val}}{2 N_{\rm train}} \right)
\label{eq:computational_effort_proof_3} 
\end{eqnarray}
Note that the quantities $N_{\rm train}$, $N_{\rm val}$, $N_{\rm epochs}^{\rm adap}$, and therefore also $C^{\rm adap}_{\rm total}$ and $R$, are dataset-dependent. Taking this into account by adding a dataset index $c$ leads to Eq.~(\ref{eq:computational_effort_adaptive_vs_fixed}).

%% file: tex/appendix_finetuning_generalization.tex
\begin{figure}[ht!]
\centering
\scalebox{\tableScale}{
\begin{tabular}{c|c|c|c}
  $c$ & $x$ & $f_1^{\rm adap./stab.}$ & $N_{\rm epochs}^{\rm adap./stab.}$  \\ \hline
  \input{analysis/tex/5_app_F_mean_std_ratio_stable_II.tex}
 \end{tabular}
}
    \qquad
    \begin{minipage}{0.45\textwidth}
     \includegraphics[scale=\figScale]{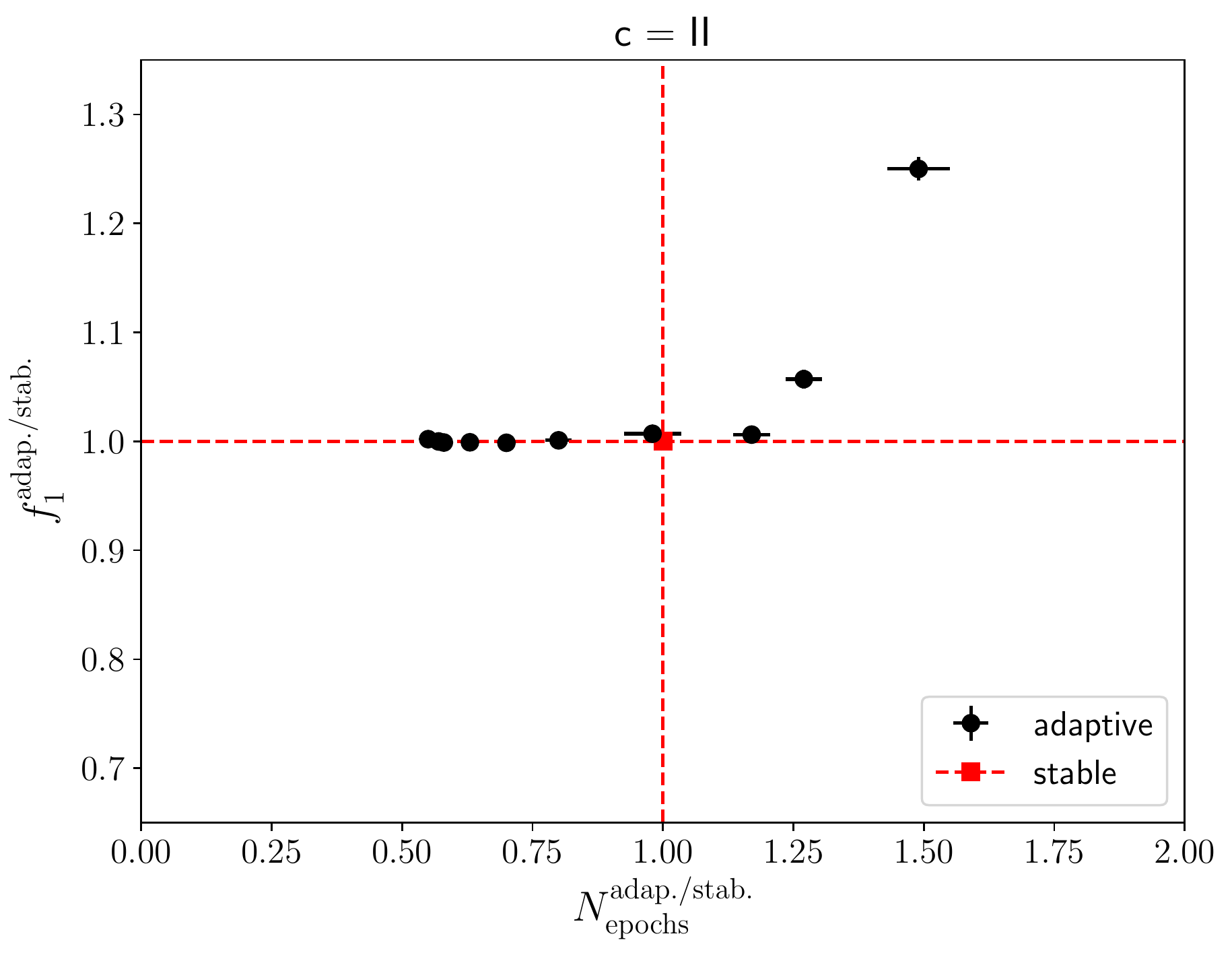}
    \end{minipage}
 \\ \vspace{2mm}
\scalebox{\tableScale}{
\begin{tabular}{c|c|c|c}
  $c$ & $x$ & $f_1^{\rm adap./orig.}$ & $N_{\rm epochs}^{\rm adap./orig.}$  \\ \hline
  \input{analysis/tex/5_app_F_mean_std_ratio_original_II.tex}
 \end{tabular}
}
    \qquad
    \begin{minipage}{0.45\textwidth}
     \includegraphics[scale=\figScale]{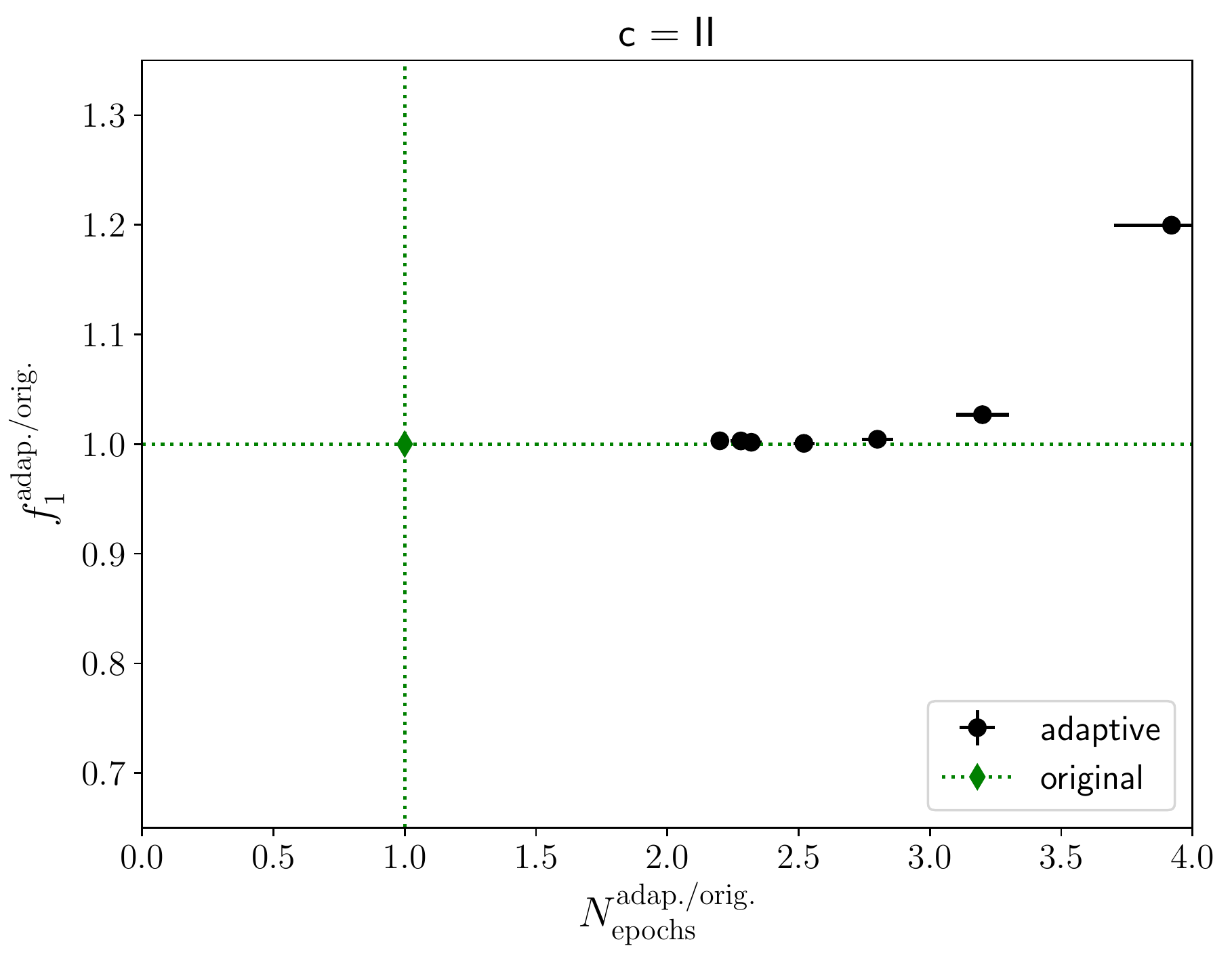}
    \end{minipage}
\\
\vspace{2mm}
\noindent\rule{\textwidth}{0.4pt}
\vspace{2mm}
\\
\scalebox{\tableScale}{
\begin{tabular}{c|c|c|c}
  $c$ & $x$ & $f_1^{\rm adap./stab.}$ & $N_{\rm epochs}^{\rm adap./stab.}$  \\ \hline
  \input{analysis/tex/5_app_F_mean_std_ratio_stable_III.tex}
 \end{tabular}
}
    \qquad
    \begin{minipage}{0.45\textwidth}
     \includegraphics[scale=\figScale]{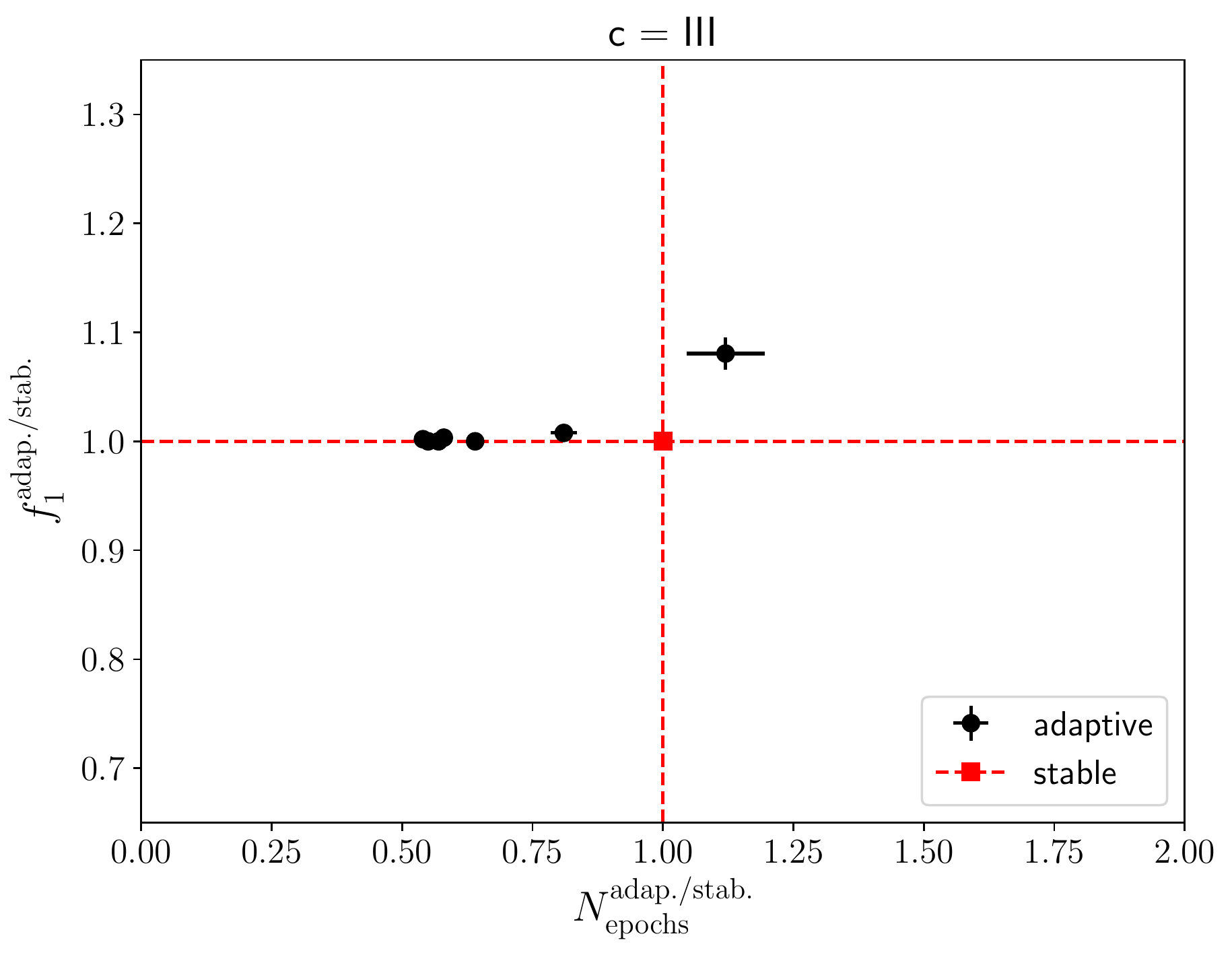}
    \end{minipage}
\\ \vspace{2mm}
\scalebox{\tableScale}{
\begin{tabular}{c|c|c|c}
  $c$ & $x$ & $f_1^{\rm adap./orig.}$ & $N_{\rm epochs}^{\rm adap./orig.}$  \\ \hline
  \input{analysis/tex/5_app_F_mean_std_ratio_original_III.tex}
 \end{tabular}
}
    \qquad
    \begin{minipage}{0.45\textwidth}
     \includegraphics[scale=\figScale]{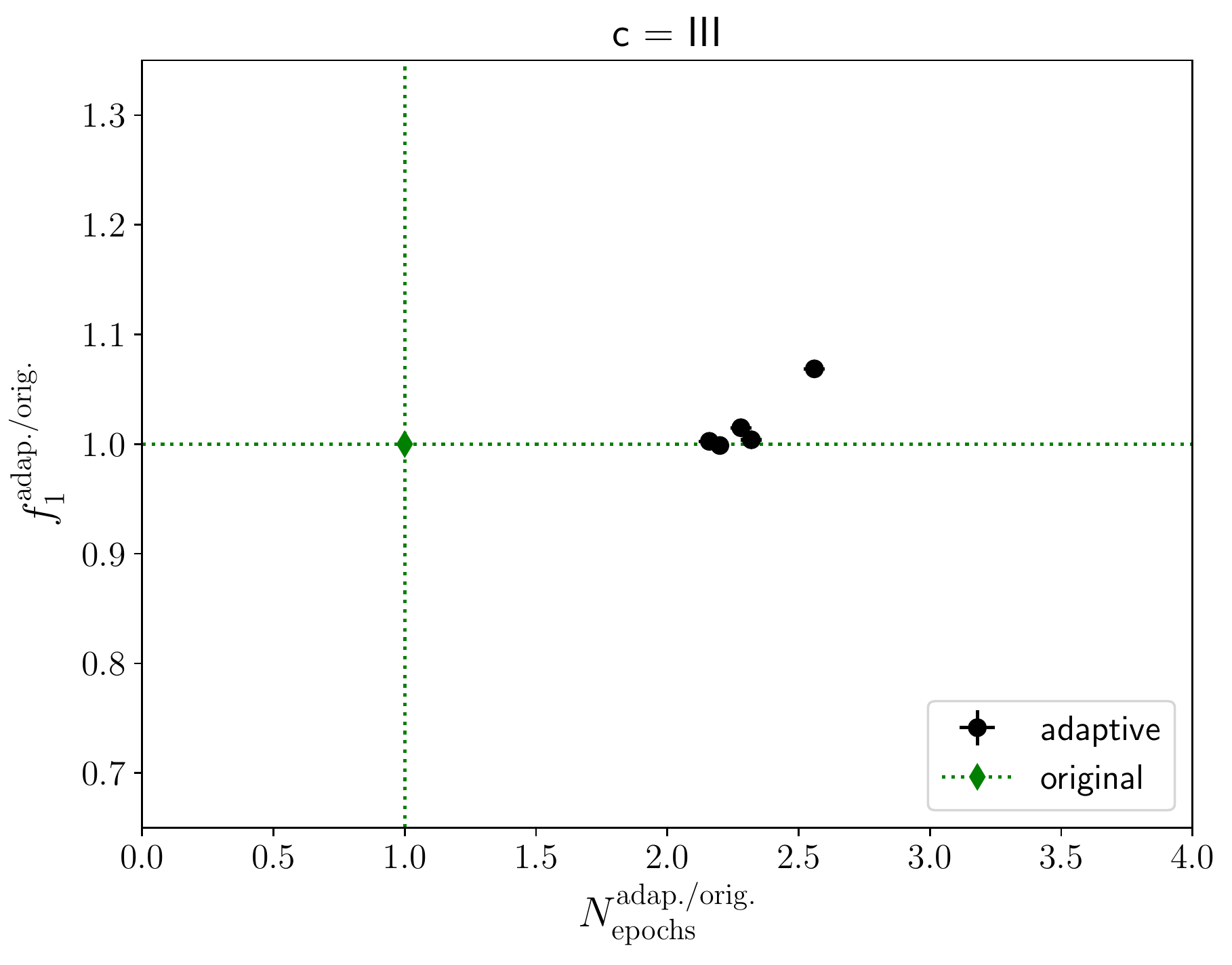}
    \end{minipage}
    \caption{Relative peformance $f_1^{\rm adap./a^\prime}$ (cf.~Eq.~(\ref{eq:f1ratio})) against the epoch ratio $N_{\rm epochs}^{\rm adap./a^\prime}$ (cf.~Eq.~(\ref{eq:epochsratio})) of the adaptive fine-tuning approach with respect to the fixed epoch approach variants $a \in \{ \text{original, stable} \}$, for different dataset scaling factors $x \in X$ and the dataset-model combination $c = \text{II}$ (top) and $c = \text{III}$ (bottom).  The relative computational cost $R^c$ approximately equals the epoch ratio times a factor $\alpha^\text{II} = 1.12$ and $\alpha^\text{III} = 1.21$ (cf.~Eq.~(\ref{eq:computational_effort_adaptive_vs_fixed}) and Tab.~\ref{tab:experiments_overview}),  respectively.  See Fig.~\ref{tab:result_training_epochs_ratio} for further details.}
 \label{fig:result_training_epochs_ratio_ALL_1}
\end{figure}
\begin{figure}[ht!]
\centering
\scalebox{\tableScale}{
\begin{tabular}{c|c|c|c}
  $c$ & $x$ & $f_1^{\rm adap./stab.}$ & $N_{\rm epochs}^{\rm adap./stab.}$  \\ \hline
  \input{analysis/tex/5_app_F_mean_std_ratio_stable_IV.tex}
 \end{tabular}
}
    \qquad
    \begin{minipage}{0.45\textwidth}
     \includegraphics[scale=\figScale]{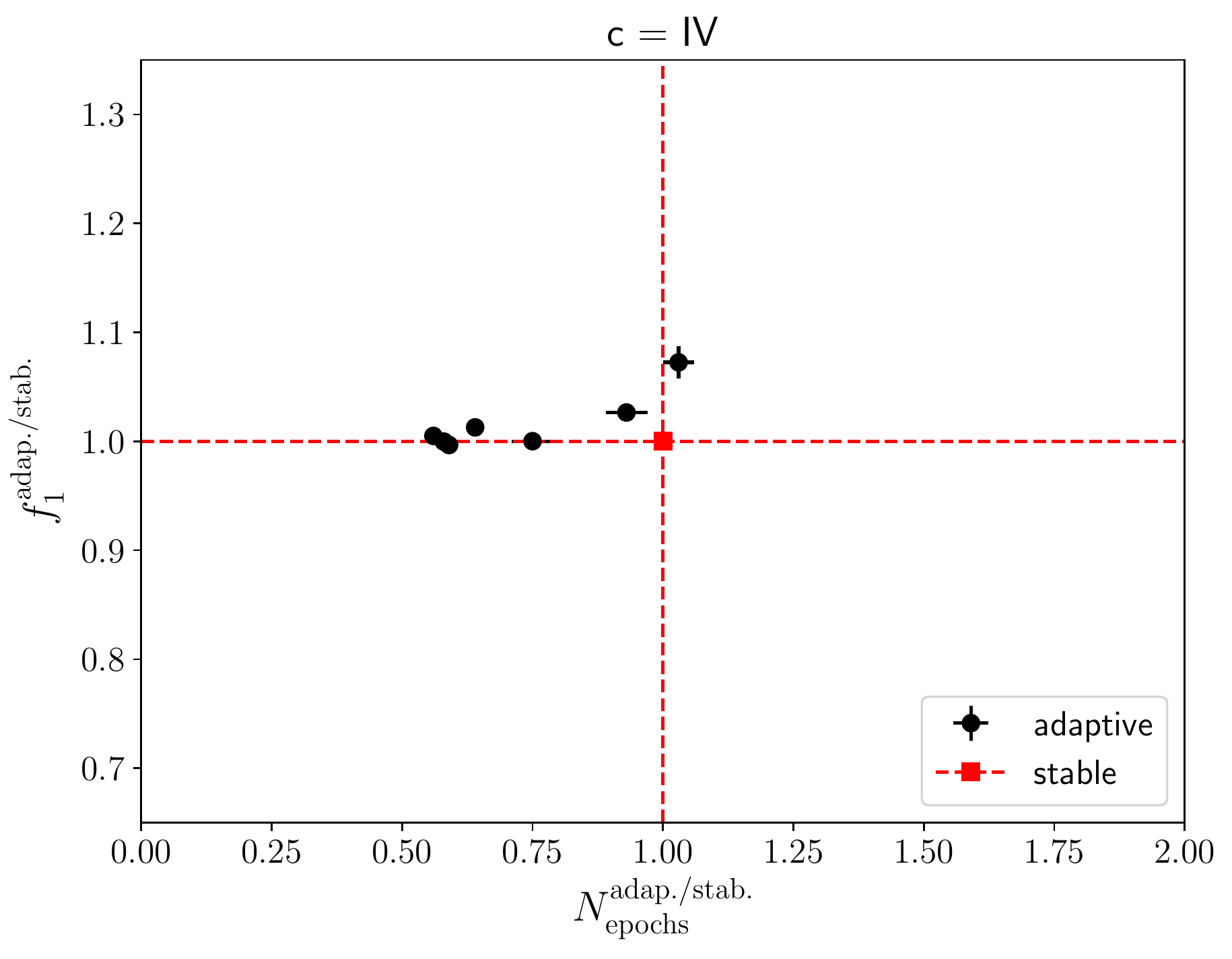}
    \end{minipage}
 \\ \vspace{2mm}
\scalebox{\tableScale}{
\begin{tabular}{c|c|c|c}
  $c$ & $x$ & $f_1^{\rm adap./orig.}$ & $N_{\rm epochs}^{\rm adap./orig.}$  \\ \hline
  \input{analysis/tex/5_app_F_mean_std_ratio_original_IV.tex}
 \end{tabular}
}
    \qquad
    \begin{minipage}{0.45\textwidth}
     \includegraphics[scale=\figScale]{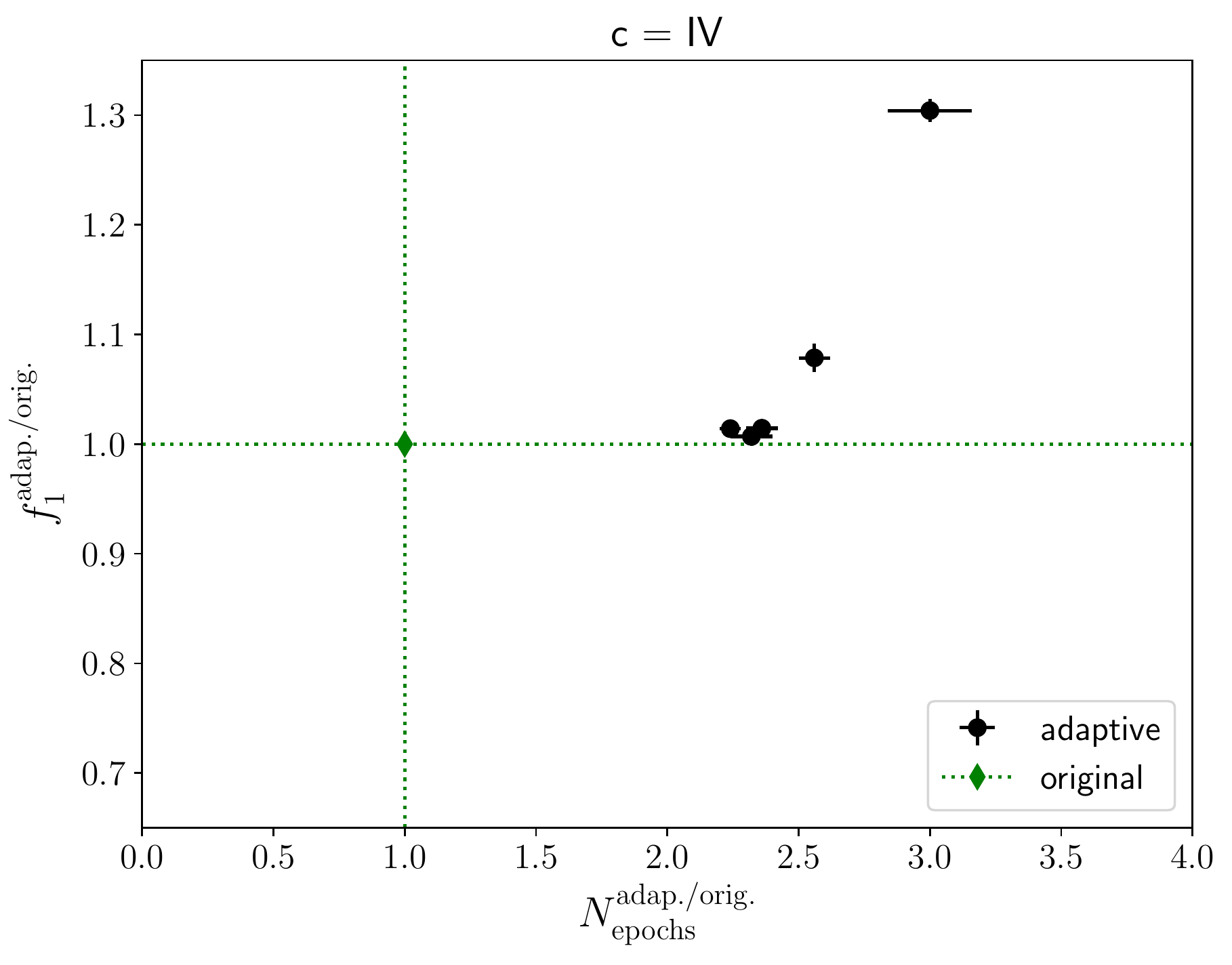}
    \end{minipage}
\\
\vspace{2mm}
\noindent\rule{\textwidth}{0.4pt}
\vspace{2mm}
\\
\scalebox{\tableScale}{
\begin{tabular}{c|c|c|c}
  $c$ & $x$ & $f_1^{\rm adap./stab.}$ & $N_{\rm epochs}^{\rm adap./stab.}$  \\ \hline
  \input{analysis/tex/5_app_F_mean_std_ratio_stable_V.tex}
 \end{tabular}
}
    \qquad
    \begin{minipage}{0.45\textwidth}
     \includegraphics[scale=\figScale]{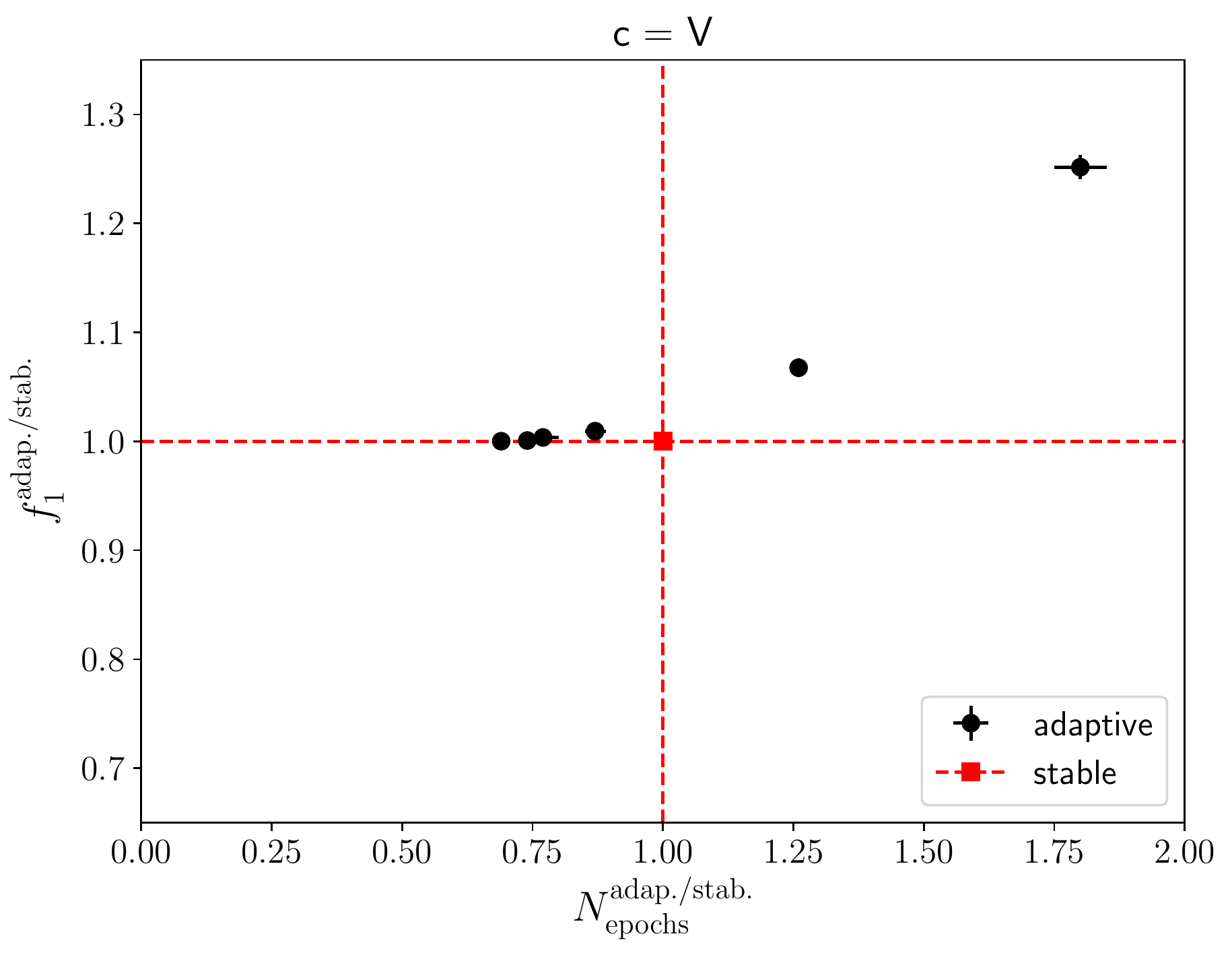}
    \end{minipage}
 \\ \vspace{2mm}
\scalebox{\tableScale}{
\begin{tabular}{c|c|c|c}
  $c$ & $x$ & $f_1^{\rm adap./orig.}$ & $N_{\rm epochs}^{\rm adap./orig.}$  \\ \hline
  \input{analysis/tex/5_app_F_mean_std_ratio_original_V.tex}
 \end{tabular}
}
    \qquad
    \begin{minipage}{0.45\textwidth}
     \includegraphics[scale=\figScale]{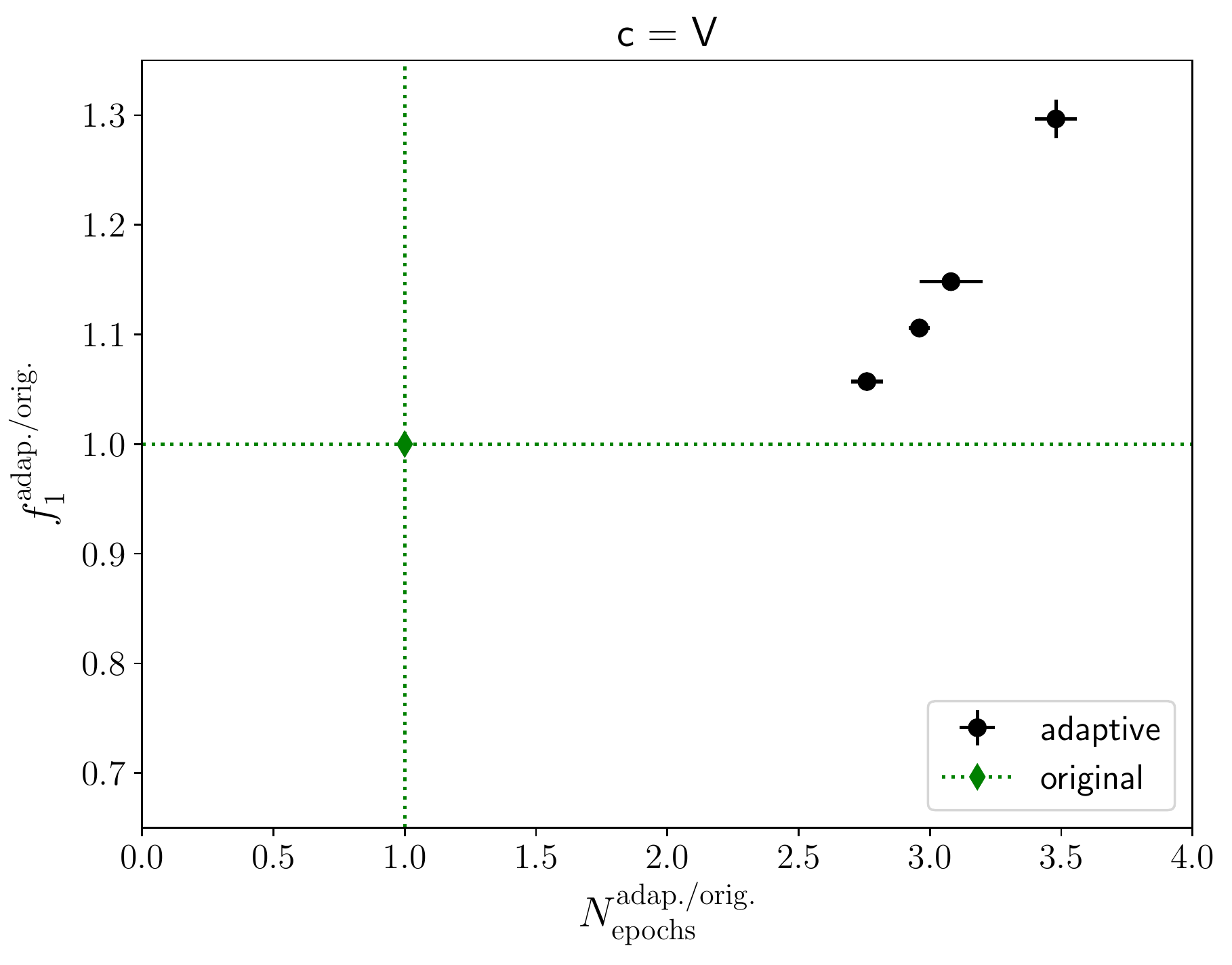}
    \end{minipage}
    \caption{Relative peformance $f_1^{\rm adap./a^\prime}$ (cf.~Eq.~(\ref{eq:f1ratio})) against the epoch ratio $N_{\rm epochs}^{\rm adap./a^\prime}$ (cf.~Eq.~(\ref{eq:epochsratio})) of the adaptive fine-tuning approach with respect to the fixed epoch approach variants $a \in \{ \text{original, stable} \}$, for different dataset scaling factors $x \in X$ and the dataset-model combination $c = \text{IV}$ (top) and $c = \text{V}$ (bottom).  The relative computational cost $R^c$ approximately equals the epoch ratio times a factor $\alpha^\text{IV} = 1.06$ and $\alpha^\text{V} = 1.12$ (cf.~Eq.~(\ref{eq:computational_effort_adaptive_vs_fixed}) and Tab.~\ref{tab:experiments_overview}),  respectively.  See Fig.~\ref{tab:result_training_epochs_ratio} for further details.
}
 \label{fig:result_training_epochs_ratio_ALL_2}
\end{figure}

In Sec.~\ref{sec:results_fine_tuning}, we discussed our results for the different fine-tuning approaches $a \in A$ and the specific dataset-model combination $c = \text{I}$.
In this appendix, we show the results for all dataset-model combinations $c \in C$.
More specifically,  in 
Tab.~\ref{tab:experiments_optimization_ALL}, 
Tab.~\ref{tab:experiments_optimization_maximum_ALL} and
Tab.~\ref{tab:stability_estimator},
we list 
$f_1$, the maximum values $f_{1,  \rm max}$ and the average relative uncertainties, respectively.
In addition,  the ratios $f_1^{\rm adap./a^\prime}$ with $a^\prime \in \{ \text{original, stable} \}$ for all $c \in C ~\textbackslash \{ \text{I} \}$ are illustrated in Fig.~\ref{fig:result_training_epochs_ratio_ALL_1} and Fig.~\ref{fig:result_training_epochs_ratio_ALL_2}. 
We note that the patterns found for $c = \text{I}$, in particular Eq.~(\ref{eq:result_performance}), hold for generally for all $c \in C$.

\begin{table}[ht!]
 \centering
\scalebox{\tableScale}{
  \begin{tabular}{c|c|c|c|c|c|c|c|c}
  & & \multicolumn{2}{c|}{original} & \multicolumn{2}{c|}{stable} & \multicolumn{3}{c}{adaptive} \\ \cline{3-9}
  \multirow{2}{*}{$c$} & \multirow{2}{*}{$x$} & \multirow{2}{*}{cv} & \multirow{2}{*}{$f_1$} & \multirow{2}{*}{cv} & \multirow{2}{*}{$f_1$} & \multirow{2}{*}{cv} & \multirow{2}{*}{$f_1$} & \multirow{2}{*}{$N_{\rm epochs}$} \\ 
  & & & & & & & \\ \hline
 \input{analysis/tex/5_app_F_mean_std.tex}
 \end{tabular}
}
 \caption{Comparison of fine-tuning results for the different fine-tuning approaches $a \in A$,  dataset-model combinations $c \in C$ and dataset scaling factors $x \in X$.  See Tab.~\ref{tab:experiments_optimization} for details.}
 \label{tab:experiments_optimization_ALL}
\end{table}

\begin{table}[ht!]
 \centering
\scalebox{\tableScale}{
 \begin{tabular}{c|c||c|c|c||c|c}
  $c$ & $x$ & $f_{1, \rm max}^\text{orig.}$ & $f_{1, \rm max}^\text{stab.}$ & $f_{1, \rm max}^\text{adap.}$ & $f_{1, \rm max}^\text{adap./orig.}$ & $f_{1, \rm max}^\text{adap./stab.}$ \\ \hline
  \input{analysis/tex/5_app_F_max.tex}
 \end{tabular}
}
 \caption{
Comparison of the maximum values $f_{1, \rm max}$ for different fine-tuning approaches $a \in A$ and dataset scaling factors $x \in X$.  The results show the performance on the test set for the run with the best performance on the validation set, and the best $f_{1, \rm max}$ of each row is highlighted in bold. The two columns on the right correspond to ratios of $f_{1, \rm max}$, analogous to Eq.~(\ref{eq:f1ratio}). 
}
 \label{tab:experiments_optimization_maximum_ALL}
\end{table}

\begin{table}[ht!]
\centering
  \floatbox[{\capbeside\thisfloatsetup{capbesideposition={left, center},capbesidewidth=8.5cm}}]{table}[\FBwidth]
{
\scalebox{\tableScale}{
 \begin{tabular}{c|c|c|c|c}
  $c$ & $\widetilde x$ & orig. & stab. & adap.   \\ \hline
   \input{analysis/tex/5_app_F_stability.tex}
 \end{tabular}
}
}
{
\caption{Average relative uncertainty $u^{a}(c)$ for the different approaches $a \in A$, computed according to Eq.~(\ref{eq:stability_estimator}) for each dataset-model combination $c \in C$.  The different choices of $\widetilde x$ take into account the dataset scaling factors $x$ for which all runs converge irrespective of $a$ and $c$, cf.~Eq.~(\ref{eq:stability_estimator_threshold}). The last row of each table shows the global average relative uncertainty $\overline{u}^{a}$, cf.~Eq.~(\ref{eq:global_stability_estimator}). }
 \label{tab:stability_estimator}
}
\end{table}

%% file: analysis/tex/5_app_F_mean_std.tex
\multirow{11}{*}{I}
 & 0.005 & 0 & --- &  & 0.3267(249) &  &{ \bf 0.6112(97) }& 49.4(1.9)  \\
 & 0.01 & 2 & --- &  & 0.6080(87) &  &{ \bf 0.7566(44) }& 35.2(3.0)  \\
 & 0.015 &  & 0.0762(256) &  & 0.7483(95) &  &{ \bf 0.8145(12) }& 25.2(1.1)  \\
 & 0.02 &  & 0.3377(295) &  & 0.8270(28) &  &{ \bf 0.8389(28) }& 22.6(7)  \\
 & 0.05 &  & 0.7846(116) &  & 0.8804(17) &  &{ \bf 0.8838(14) }& 17.6(1.2)  \\
 & 0.1 &  & 0.8801(22) &  & 0.8937(12) &  & 0.8942(1) & 15.4(7)  \\
 & 0.2 &  & 0.9000(8) &  & 0.9066(2) &  & 0.9063(4) & 13.4(9)  \\
 & 0.4 &  & 0.9096(7) &  & 0.9118(11) &  & 0.9111(6) & 11.2(2)  \\
 & 0.6 &  & 0.9136(9) &  & 0.9157(7) &  & 0.9155(9) & 11.6(5)  \\
 & 0.8 &  & 0.9169(5) &  & 0.9202(4) &  & 0.9201(4) & 11.4(4)  \\
 & 1.0 &  & 0.9175(5) &  & 0.9195(7) &  &{ \bf 0.9214(8) }& 11.2(2)  \\ \hline
\multirow{11}{*}{II}
 & 0.005 & 0 & --- &  & 0.1987(263) &  &{ \bf 0.6631(72) }& 42.8(2.9)  \\
 & 0.01 & 0 & --- &  & 0.5923(6) &  &{ \bf 0.7404(46) }& 29.8(1.2)  \\
 & 0.015 & 0 & --- &  & 0.7290(69) &  &{ \bf 0.7706(31) }& 25.4(7)  \\
 & 0.02 &  & 0.1245(418) &  & 0.7805(38) &  & 0.7853(41) & 23.4(7)  \\
 & 0.05 &  & 0.7108(64) &  & 0.8468(11) &  &{ \bf 0.8527(6) }& 19.6(1.1)  \\
 & 0.1 &  & 0.8465(11) &  & 0.8684(8) &  & 0.8692(11) & 16.0(5)  \\
 & 0.2 &  & 0.8798(6) &  & 0.8849(9) &  & 0.8837(8) & 14.0(3)  \\
 & 0.4 &  & 0.8949(3) &  & 0.8963(6) &  & 0.8955(8) & 12.6(2)  \\
 & 0.6 &  & 0.9011(2) &  &{ \bf 0.9036(4) }&  & 0.9026(7) & 11.6(2)  \\
 & 0.8 &  & 0.9015(3) &  & 0.9042(6) &  & 0.9041(6) & 11.4(2)  \\
 & 1.0 &  & 0.9028(8) &  & 0.9037(9) &  &{ \bf 0.9055(6) }& 11.0(0)  \\ \hline
\multirow{11}{*}{III}
 & 0.005 & 0 & --- & 0 & --- & 1 & --- & ---  \\
 & 0.01 & 0 & --- & 0 & --- & 1 & --- & ---  \\
 & 0.015 & 0 & --- & 0 & --- &  &{ \bf 0.3682(294) }& 35.6(3.3)  \\
 & 0.02 & 0 & --- &  & 0.2430(84) &  &{ \bf 0.5869(102) }& 42.0(2.5)  \\
 & 0.05 & 0 & --- &  & 0.6864(103) &  &{ \bf 0.7417(67) }& 22.4(1.5)  \\
 & 0.1 &  & 0.5786(111) &  & 0.8046(35) &  &{ \bf 0.8108(29) }& 16.2(5)  \\
 & 0.2 &  & 0.7913(63) &  & 0.8456(22) &  & 0.8456(19) & 12.8(2)  \\
 & 0.4 &  & 0.8494(22) &  & 0.8621(15) &  & 0.8621(13) & 11.4(2)  \\
 & 0.6 &  & 0.8620(2) &  & 0.8609(18) &  & 0.8609(24) & 11.0(0)  \\
 & 0.8 &  & 0.8657(21) &  & 0.8662(11) &  &{ \bf 0.8691(17) }& 11.6(2)  \\
 & 1.0 &  & 0.8695(12) &  & 0.8700(2) &  &{ \bf 0.8717(14) }& 10.8(2)  \\ \hline
\multirow{11}{*}{IV}
 & 0.005 & 0 & --- & 0 & --- & 0 & --- & ---  \\
 & 0.01 & 0 & --- & 0 & --- & 2 & --- & ---  \\
 & 0.015 & 0 & --- & 0 & --- &  &{ \bf 0.3798(81) }& 34.0(1.1)  \\
 & 0.02 & 0 & --- & 3 & --- &  &{ \bf 0.4566(187) }& 32.6(1.5)  \\
 & 0.05 & 0 & --- &  & 0.5671(49) &  &{ \bf 0.6082(78) }& 20.6(6)  \\
 & 0.1 &  & 0.2217(391) &  & 0.6576(39) &  &{ \bf 0.6750(31) }& 18.6(8)  \\
 & 0.2 &  & 0.5733(7) &  & 0.7476(18) &  & 0.7477(32) & 15.0(8)  \\
 & 0.4 &  & 0.7262(119) &  & 0.7734(33) &  &{ \bf 0.7833(21) }& 12.8(3)  \\
 & 0.6 &  & 0.7810(55) &  & 0.7950(17) &  & 0.7923(26) & 11.8(3)  \\
 & 0.8 &  & 0.7945(18) &  & 0.8017(14) &  &{ \bf 0.8057(14) }& 11.2(2)  \\
 & 1.0 &  & 0.8011(19) &  & 0.8069(33) &  & 0.8067(34) & 11.6(4)  \\ \hline
\multirow{11}{*}{V}
 & 0.005 & 3 & --- & 2 & --- &  &{ \bf 0.5642(84) }& 191.2(20.5)  \\
 & 0.01 & 3 & --- & 2 & --- &  &{ \bf 0.6331(97) }& 149.6(14.1)  \\
 & 0.015 & 3 & --- & 4 & --- &  &{ \bf 0.6645(56) }& 125.8(6.8)  \\
 & 0.02 & 3 & --- & 4 & --- &  &{ \bf 0.6586(39) }& 110.8(3.8)  \\
 & 0.05 & 3 & --- & 4 & --- &  &{ \bf 0.7077(42) }& 62.6(2.0)  \\
 & 0.1 & 3 & --- &  & 0.5838(81) &  &{ \bf 0.7307(29) }& 36.0(1.0)  \\
 & 0.2 &  & 0.2073(716) &  & 0.7135(4) &  &{ \bf 0.7617(54) }& 25.2(2)  \\
 & 0.4 &  & 0.5998(123) &  & 0.7705(49) &  &{ \bf 0.7777(48) }& 17.4(4)  \\
 & 0.6 &  & 0.6844(26) &  & 0.7830(44) &  & 0.7858(38) & 15.4(6)  \\
 & 0.8 &  & 0.7169(3) &  & 0.7922(29) &  & 0.7928(3) & 14.8(2)  \\
 & 1.0 &  & 0.7558(32) &  & 0.7988(9) &  & 0.7989(15) & 13.8(3)

%% file: analysis/tex/5_app_F_max.tex
\multirow{11}{*}{I}
 & 0.005 & --- & 0.3591 &{ \bf 0.6522 }& --- & 1.816  \\
 & 0.01 & --- & 0.6292 &{ \bf 0.7594 }& --- & 1.207  \\
 & 0.015 & 0.1764 & 0.7759 &{ \bf 0.8197 }& 4.647 & 1.056  \\
 & 0.02 & 0.4463 & 0.8179 &{ \bf 0.8428 }& 1.888 & 1.030  \\
 & 0.05 & 0.8189 & 0.8858 &{ \bf 0.8870 }& 1.083 & 1.001  \\
 & 0.1 & 0.8808 &{ \bf 0.8964 }& 0.8916 & 1.012 & 0.995  \\
 & 0.2 & 0.9022 & 0.9063 &{ \bf 0.9072 }& 1.006 & 1.001  \\
 & 0.4 & 0.9079 &{ \bf 0.9129 }& 0.9091 & 1.001 & 0.996  \\
 & 0.6 & 0.9126 & 0.9154 &{ \bf 0.9161 }& 1.004 & 1.001  \\
 & 0.8 & 0.9177 &{ \bf 0.9199 }& 0.9197 & 1.002 & 1.000  \\
 & 1.0 & 0.9165 & 0.9210 & 0.9210 & 1.005 & 1.000  \\ \hline
\multirow{11}{*}{II}
 & 0.005 & --- & 0.1468 &{ \bf 0.6414 }& --- & 4.369  \\
 & 0.01 & --- & 0.6013 &{ \bf 0.7559 }& --- & 1.257  \\
 & 0.015 & --- & 0.7375 &{ \bf 0.7610 }& --- & 1.032  \\
 & 0.02 & 0.2806 & 0.7646 &{ \bf 0.7804 }& 2.781 & 1.021  \\
 & 0.05 & 0.7134 & 0.8502 &{ \bf 0.8549 }& 1.198 & 1.006  \\
 & 0.1 & 0.8475 & 0.8662 &{ \bf 0.8724 }& 1.029 & 1.007  \\
 & 0.2 & 0.8817 & 0.8810 &{ \bf 0.8834 }& 1.002 & 1.003  \\
 & 0.4 & 0.8957 & 0.8940 &{ \bf 0.8961 }& 1.000 & 1.002  \\
 & 0.6 & 0.9014 & 0.9038 &{ \bf 0.9041 }& 1.003 & 1.000  \\
 & 0.8 & 0.9008 &{ \bf 0.9048 }& 0.9031 & 1.003 & 0.998  \\
 & 1.0 &{ \bf 0.9040 }& 0.9017 & 0.9032 & 0.999 & 1.002  \\ \hline
\multirow{11}{*}{III}
 & 0.005 & --- & --- & --- & --- & ---  \\
 & 0.01 & --- & --- & --- & --- & ---  \\
 & 0.015 & --- & --- &{ \bf 0.4106 }& --- & ---  \\
 & 0.02 & --- & 0.2363 &{ \bf 0.6101 }& --- & 2.582  \\
 & 0.05 & --- & 0.6787 &{ \bf 0.7466 }& --- & 1.100  \\
 & 0.1 & 0.5949 & 0.7908 &{ \bf 0.8190 }& 1.377 & 1.036  \\
 & 0.2 & 0.7841 &{ \bf 0.8500 }& 0.8385 & 1.069 & 0.986  \\
 & 0.4 & 0.8571 &{ \bf 0.8651 }& 0.8620 & 1.006 & 0.996  \\
 & 0.6 & 0.8642 & 0.8582 &{ \bf 0.8663 }& 1.002 & 1.009  \\
 & 0.8 & 0.8668 & 0.8677 &{ \bf 0.8709 }& 1.005 & 1.004  \\
 & 1.0 & 0.8743 & 0.8717 &{ \bf 0.8774 }& 1.004 & 1.007  \\ \hline
\multirow{11}{*}{IV}
 & 0.005 & --- & --- & --- & --- & ---  \\
 & 0.01 & --- & --- & --- & --- & ---  \\
 & 0.015 & --- & --- &{ \bf 0.3791 }& --- & ---  \\
 & 0.02 & --- & --- &{ \bf 0.4286 }& --- & ---  \\
 & 0.05 & --- & 0.5839 &{ \bf 0.6204 }& --- & 1.063  \\
 & 0.1 & 0.3357 & 0.6601 &{ \bf 0.6766 }& 2.015 & 1.025  \\
 & 0.2 & 0.5769 &{ \bf 0.7526 }& 0.7511 & 1.302 & 0.998  \\
 & 0.4 & 0.7679 & 0.7753 &{ \bf 0.7783 }& 1.014 & 1.004  \\
 & 0.6 & 0.7946 & 0.7934 &{ \bf 0.7971 }& 1.003 & 1.005  \\
 & 0.8 & 0.8004 & 0.7960 &{ \bf 0.8027 }& 1.003 & 1.008  \\
 & 1.0 & 0.7996 & 0.8056 &{ \bf 0.8135 }& 1.017 & 1.010  \\ \hline
\multirow{11}{*}{V}
 & 0.005 & --- & --- &{ \bf 0.5658 }& --- & ---  \\
 & 0.01 & --- & --- &{ \bf 0.6667 }& --- & ---  \\
 & 0.015 & --- & --- &{ \bf 0.6624 }& --- & ---  \\
 & 0.02 & --- & --- &{ \bf 0.6512 }& --- & ---  \\
 & 0.05 & --- & --- &{ \bf 0.7061 }& --- & ---  \\
 & 0.1 & --- & 0.5971 &{ \bf 0.7352 }& --- & 1.231  \\
 & 0.2 & 0.4219 & 0.7154 &{ \bf 0.7390 }& 1.752 & 1.033  \\
 & 0.4 & 0.6249 &{ \bf 0.7857 }& 0.7577 & 1.213 & 0.964  \\
 & 0.6 & 0.6855 & 0.7878 &{ \bf 0.7983 }& 1.165 & 1.013  \\
 & 0.8 & 0.7068 & 0.7840 &{ \bf 0.7913 }& 1.120 & 1.009  \\
 & 1.0 & 0.7480 &{ \bf 0.8021 }& 0.8014 & 1.071 & 0.999

%% file: analysis/tex/5_app_F_stability.tex
 I & 0.015 & 0.0494 & 0.0025 & 0.0012  \\
 II & 0.02 & 0.0436 & 0.0014 & 0.0014  \\
 III & 0.1 & 0.0060 & 0.0024 & 0.0023  \\
 IV & 0.1 & 0.0361 & 0.0034 & 0.0035  \\
 V & 0.2 & 0.0756 & 0.0045 & 0.0048  \\ \hline
 glob. &  & 0.0421 & 0.0028 & 0.0026

%% file: analysis/tex/5_app_F_mean_std_ratio_stable_II.tex
\multirow{11}{*}{II}
 & 0.005 & 3.337(95) & 2.14(14)  \\
 & 0.01 & 1.250(11) & 1.49(6)  \\
 & 0.015 & 1.057(8) & 1.27(3)  \\
 & 0.02 & 1.006(6) & 1.17(3)  \\
 & 0.05 & 1.007(1) & 0.98(6)  \\
 & 0.1 & 1.001(1) & 0.80(3)  \\
 & 0.2 & 0.999(1) & 0.70(1)  \\
 & 0.4 & 0.999(1) & 0.63(1)  \\
 & 0.6 & 0.999(1) & 0.58(1)  \\
 & 0.8 & 1.000(1) & 0.57(1)  \\
 & 1.0 & 1.002(1) & 0.55(0)

%% file: analysis/tex/5_app_F_mean_std_ratio_original_II.tex
\multirow{11}{*}{II}
 & 0.005 & --- & ---  \\
 & 0.01 & --- & ---  \\
 & 0.015 & --- & ---  \\
 & 0.02 & 6.308(266) & 4.68(14)  \\
 & 0.05 & 1.200(8) & 3.92(22)  \\
 & 0.1 & 1.027(2) & 3.20(1)  \\
 & 0.2 & 1.004(1) & 2.80(6)  \\
 & 0.4 & 1.001(1) & 2.52(4)  \\
 & 0.6 & 1.002(1) & 2.32(4)  \\
 & 0.8 & 1.003(1) & 2.28(4)  \\
 & 1.0 & 1.003(1) & 2.20(0)

%% file: analysis/tex/5_app_F_mean_std_ratio_stable_III.tex
\multirow{11}{*}{III}
 & 0.005 & --- & ---  \\
 & 0.01 & --- & ---  \\
 & 0.015 & --- & ---  \\
 & 0.02 & 2.415(47) & 2.10(12)  \\
 & 0.05 & 1.081(15) & 1.12(7)  \\
 & 0.1 & 1.008(5) & 0.81(3)  \\
 & 0.2 & 1.000(3) & 0.64(1)  \\
 & 0.4 & 1.000(2) & 0.57(1)  \\
 & 0.6 & 1.000(3) & 0.55(0)  \\
 & 0.8 & 1.003(2) & 0.58(1)  \\
 & 1.0 & 1.002(3) & 0.54(1)

%% file: analysis/tex/5_app_F_mean_std_ratio_original_III.tex
\multirow{11}{*}{III}
 & 0.005 & --- & ---  \\
 & 0.01 & --- & ---  \\
 & 0.015 & --- & ---  \\
 & 0.02 & --- & ---  \\
 & 0.05 & --- & ---  \\
 & 0.1 & 1.401(16) & 3.24(1)  \\
 & 0.2 & 1.069(7) & 2.56(4)  \\
 & 0.4 & 1.015(3) & 2.28(4)  \\
 & 0.6 & 0.999(3) & 2.20(0)  \\
 & 0.8 & 1.004(3) & 2.32(4)  \\
 & 1.0 & 1.003(2) & 2.16(4)

%% file: analysis/tex/5_app_F_mean_std_ratio_stable_IV.tex
\multirow{11}{*}{IV}
 & 0.005 & --- & ---  \\
 & 0.01 & --- & ---  \\
 & 0.015 & --- & ---  \\
 & 0.02 & --- & ---  \\
 & 0.05 & 1.072(15) & 1.03(3)  \\
 & 0.1 & 1.026(6) & 0.93(4)  \\
 & 0.2 & 1.000(5) & 0.75(4)  \\
 & 0.4 & 1.013(4) & 0.64(1)  \\
 & 0.6 & 0.997(4) & 0.59(1)  \\
 & 0.8 & 1.005(2) & 0.56(1)  \\
 & 1.0 & 1.000(5) & 0.58(2)

%% file: analysis/tex/5_app_F_mean_std_ratio_original_IV.tex
\multirow{11}{*}{IV}
 & 0.005 & --- & ---  \\
 & 0.01 & --- & ---  \\
 & 0.015 & --- & ---  \\
 & 0.02 & --- & ---  \\
 & 0.05 & --- & ---  \\
 & 0.1 & 3.045(12) & 3.72(16)  \\
 & 0.2 & 1.304(11) & 3.00(16)  \\
 & 0.4 & 1.079(13) & 2.56(6)  \\
 & 0.6 & 1.014(6) & 2.36(6)  \\
 & 0.8 & 1.014(3) & 2.24(4)  \\
 & 1.0 & 1.007(5) & 2.32(8)

%% file: analysis/tex/5_app_F_mean_std_ratio_stable_V.tex
\multirow{11}{*}{V}
 & 0.005 & --- & ---  \\
 & 0.01 & --- & ---  \\
 & 0.015 & --- & ---  \\
 & 0.02 & --- & ---  \\
 & 0.05 & --- & ---  \\
 & 0.1 & 1.252(11) & 1.80(5)  \\
 & 0.2 & 1.068(9) & 1.26(1)  \\
 & 0.4 & 1.009(8) & 0.87(2)  \\
 & 0.6 & 1.004(7) & 0.77(3)  \\
 & 0.8 & 1.001(5) & 0.74(1)  \\
 & 1.0 & 1.000(2) & 0.69(1)

%% file: analysis/tex/5_app_F_mean_std_ratio_original_V.tex
\multirow{11}{*}{V}
 & 0.005 & --- & ---  \\
 & 0.01 & --- & ---  \\
 & 0.015 & --- & ---  \\
 & 0.02 & --- & ---  \\
 & 0.05 & --- & ---  \\
 & 0.1 & --- & ---  \\
 & 0.2 & 3.674(264) & 5.04(4)  \\
 & 0.4 & 1.297(18) & 3.48(8)  \\
 & 0.6 & 1.148(6) & 3.08(12)  \\
 & 0.8 & 1.106(5) & 2.96(4)  \\
 & 1.0 & 1.057(4) & 2.76(6)

%% file: tex/appendix_variations.tex
\begin{figure}[ht!]
 \centering
 \includegraphics[scale=\figScaleThree]{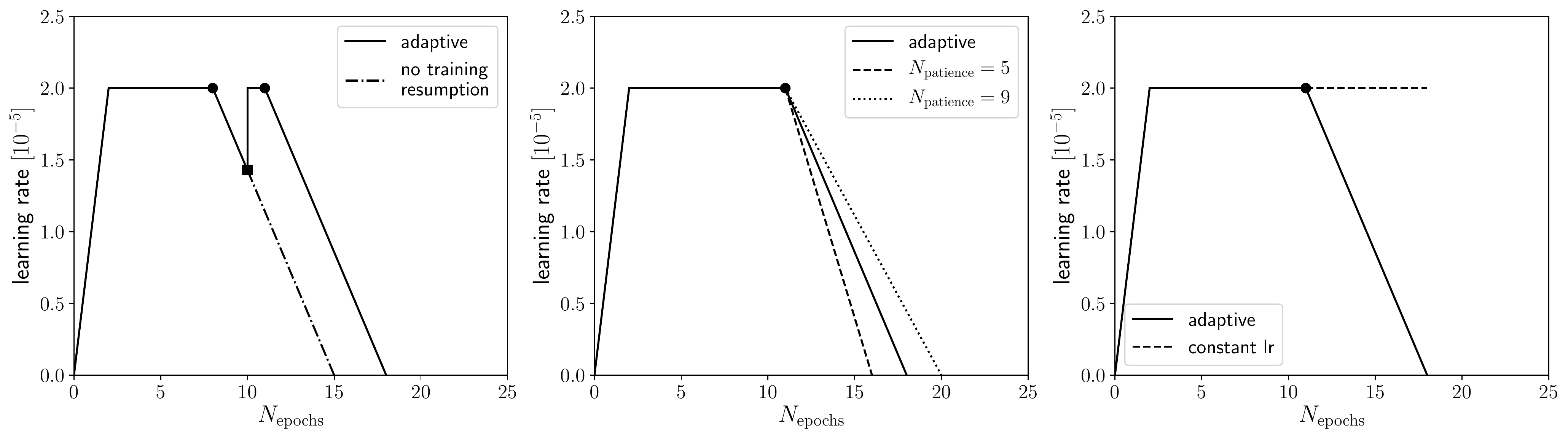} 
 \caption{Comparison of the standard adaptive approach to some variations.  Circles indicate the beginning of an early stopping stage.  \textit{Left:} adaptive fine-tuning without training resumption, where early stopping is executed irrespective of the validation loss.  \textit{Center:} adaptive fine-tuning with $N_{\rm patience} \in \{5, 7, 9\}$. \textit{Right:} adaptive fine-tuning where the learning rate is held constant instead of linearly decreased.}
 \label{fig:escd_variations}
\end{figure}

In this appendix, we study variations of the adaptive fine-tuning approach. 
Firstly, in Sec.~\ref{app:escd_variation_ablation}, we consider simplifications in order to verify the relevance of training resumption and the cool-down phase. 
Next, we vary the length $N_{\rm patience}$ of the cool-down phase (Sec.~\ref{app:escd_variation_number_cool_down}).
Finally, we compare our results to an early stopping approach where the learning rate is kept constant, in Sec.~\ref{app:escd_variation_early_stopping}.
The different variations are illustrated in Fig.~\ref{fig:escd_variations}.
For brevity, we restrict our studies to the dataset-model combinations $c \in \{ \text{II, III, IV} \}$.

\subsection{Simplification \label{app:escd_variation_ablation}}
\begin{figure}[t]
\begin{floatrow}
\ffigbox{%
   \includegraphics[scale=\figScaleThree]{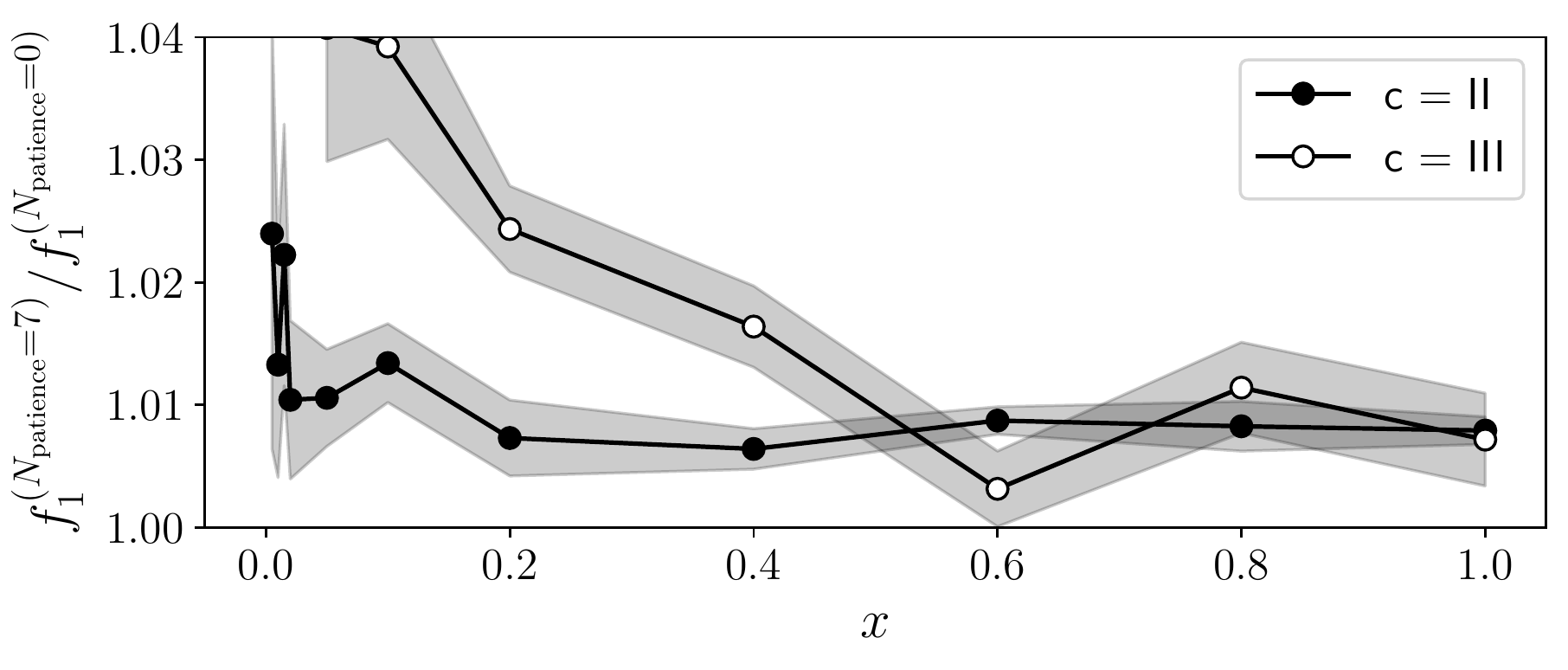}
}{%
  \caption{Relative performance of the two simplification variants without training resumption, $f_1^{(N_{\rm patience}=7)} / f_1^{(N_{\rm patience}=0)}$, as a function of the dataset scaling factor $x$.}
  \label{fig:experiments_optimization_variation_ablation_2}
}
\capbtabbox{%
\scalebox{\tableScale}{
 \begin{tabular}{c|c|c|c|c}
 \multicolumn{2}{c|}{} & \multicolumn{2}{c|}{variant} & adaptive \\
 \multicolumn{2}{c|}{training res.} & \multicolumn{2}{c|}{no} & yes \\ \hline
  $c$  & $\widetilde x$ & $N_{\rm patience}=0$ & $N_{\rm patience}=7$ & $N_{\rm patience}=7$ \\ \hline
 \input{analysis/tex/2_variation_simplification_app_G1_stability.tex}
 \linebreak \vspace{8mm}
 \end{tabular}
}
}{%
 \caption{Average relative uncertainty $u^{a}(c)$ for $c \in \{ \text{II}, \text{III} \}$ and the adaptive variants without training resumption. See Tab.~\ref{tab:stability_estimator} for further details.}
 \label{tab:experiments_optimization_variation_ablation_2}
}
\end{floatrow}
\end{figure}

We conduct some experiments while dropping elements of the adaptive approach, see the left panel of Fig.~\ref{fig:escd_variations}.  Firstly, we consider immediate early stopping, where we drop both the linear cool-down phase and the training resumption.  Subsequently, we employ the cool-down phase and drop only the possibility to resume the training.
The results can be found in Tab.~\ref{tab:experiments_optimization_variation_ablation}.
\begin{table}[ht]
 \centering
\scalebox{\tableScale}{
\begin{tabular}{c|c|c|c|c|c|c|c|c|c|c}
                       & & \multicolumn{6}{c|}{variant} & \multicolumn{3}{c}{adaptive}  \\
                       & & \multicolumn{3}{c|}{$N_{\rm patience = 0}$} & \multicolumn{3}{c|}{$N_{\rm patience = 7}$} & \multicolumn{3}{c}{$N_{\rm patience = 7}$}  \\
   & & \multicolumn{3}{c|}{} & \multicolumn{3}{c|}{no training resumption} & \multicolumn{3}{c}{training resumption} \\ \cline{3-11}
  \multirow{2}{*}{$c$} & \multirow{2}{*}{$x$} & \multirow{2}{*}{cv} & \multirow{2}{*}{$f_1$} & \multirow{2}{*}{$N_{\rm epochs}$} & \multirow{2}{*}{cv} & \multirow{2}{*}{$f_1$} & \multirow{2}{*}{$N_{\rm epochs}$} & \multirow{2}{*}{cv} & \multirow{2}{*}{$f_1$} & \multirow{2}{*}{$N_{\rm epochs}$}  \\ 
  & & & & & & & & & & \\ \hline
 \hline
 \input{analysis/tex/2_variation_simplification_app_G1_mean_std.tex}
 \end{tabular}
}
 \caption{Comparison of fine-tuning results for simplified versions of the adaptive approach,  where either early stopping is applied immediately without patience, or the resumption of training in case of validation loss improvement is dropped. }
 \label{tab:experiments_optimization_variation_ablation}
\end{table}
In addition, Fig.~\ref{fig:experiments_optimization_variation_ablation_2} shows the relative performance of the two simplification variants, while the average relative uncertainties are compared in Tab.~\ref{tab:experiments_optimization_variation_ablation_2}.

We observe that that additional linear cool-down epochs $N_{\rm patience}$ have a positive impact on both model performance and stability. Moreover, we find that training resumption improves convergence for very small datasets ($c = \text{III}$ and $x \leq 0.02$).

\subsection{Varying length of the cool-down phase \label{app:escd_variation_number_cool_down}}

\begin{figure}[ht!]
 \centering
 \includegraphics[scale=\figScaleThree]{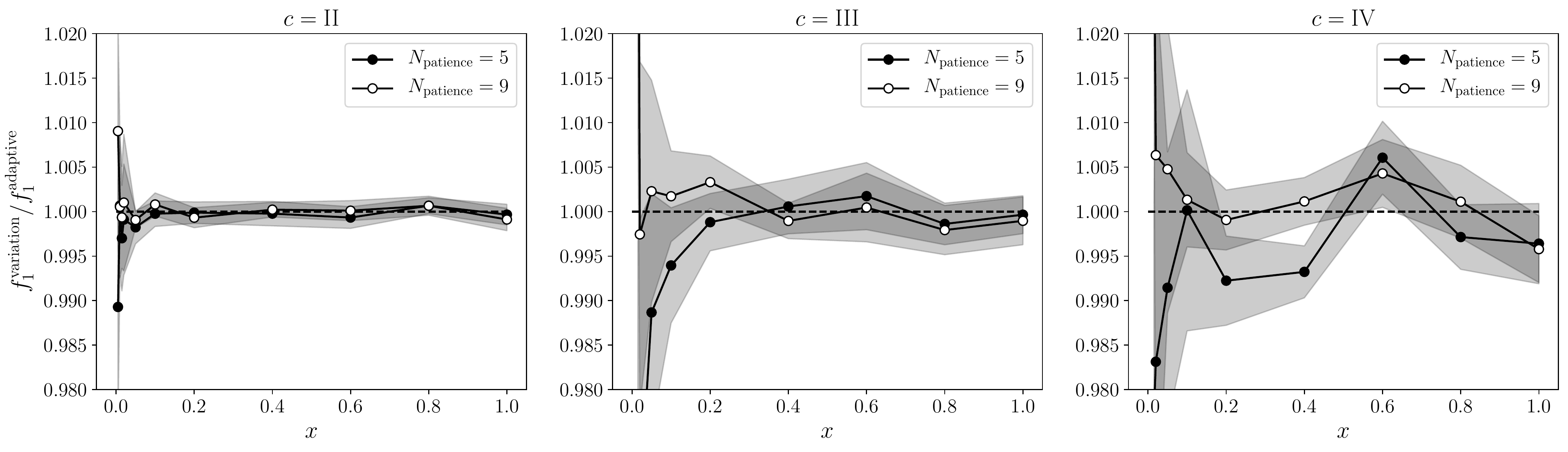}
 \caption{Relative performance of the two variants with modified patience ($N_{\rm patience} \in \{ 5, 9 \}$) with respect to the standard adaptive approach ($N_{\rm patience}=7$), as a function of the dataset scaling factor $x$. The results are shown separately for the dataset-model combinations $c \in \{ \text{II, III, IV} \}$.}
 \label{fig:experiments_optimization_variation_number_of_epochs_2}
\end{figure}

In Sec.~\ref{sec:adaptive},  the parameter $N_{\rm patience}$, which is the number of epochs that training is continued for after an increase in validation loss is detected, was set to a fixed value in Eq.~(\ref{eq:Npatience}).
Here, we motivate our choice after conducting some experiments using other values of $N_{\rm patience}$, see the center panel of Fig.~\ref{fig:escd_variations}.
The bare results can be found in Tab.~\ref{tab:experiments_optimization_variation_number_of_epochs}, 
and the performances for the different variants are compared in Fig.~\ref{fig:experiments_optimization_variation_number_of_epochs_2}.
\begin{table}[ht]
 \centering
\scalebox{\tableScale}{
 \begin{tabular}{c|c|c|c|c|c|c|c|c|c|c}
  & & \multicolumn{3}{c|}{variant} & \multicolumn{3}{c|}{adaptive} & \multicolumn{3}{c}{variant}  \\ 
   & & \multicolumn{3}{c|}{$N_{\rm patience} = 5$} & \multicolumn{3}{c|}{$N_{\rm patience} = 7$} & \multicolumn{3}{c}{$N_{\rm patience} = 9$} \\ \cline{3-11}
  \multirow{2}{*}{$c$} & \multirow{2}{*}{$x$} & \multirow{2}{*}{cv} & \multirow{2}{*}{$f_1$} & \multirow{2}{*}{$N_{\rm epochs}$} & \multirow{2}{*}{cv} & \multirow{2}{*}{$f_1$} & \multirow{2}{*}{$N_{\rm epochs}$} & \multirow{2}{*}{cv} & \multirow{2}{*}{$f_1$} & \multirow{2}{*}{$N_{\rm epochs}$} \\ 
  & & & & & & & & & & \\ \hline
 \input{analysis/tex/3_variation_cooldown_app_G2_mean_std.tex}
 \end{tabular}
}
 \caption{Fine-tuning results for variants of the adaptive approach with different $N_{\rm patience}$.}
 \label{tab:experiments_optimization_variation_number_of_epochs}
\end{table}

\begin{figure}[ht!]
\begin{floatrow}
\capbtabbox{%
\scalebox{\tableScale}{
 \begin{tabular}{c|c|c|c|c}
  & & variant & adaptive & variant \\
  $c$  & $\widetilde x$ & $N_{\rm patience}=5$ & $N_{\rm patience}=7$  & $N_{\rm patience}=9$ \\ \hline
  \input{analysis/tex/3_variation_cooldown_app_G2_stability.tex}
\end{tabular}
}
}{%
 \caption{Average relative uncertainty $u^{a}(c)$ for $c \in \{ \text{II}, \text{III}, \text{IV} \}$ and the adaptive variants with $N_{\rm patience} \in \{ 5, 7, 9 \}$. See Tab.~\ref{tab:stability_estimator} for further details.}
 \label{tab:experiments_optimization_variation_number_of_epochs_2}
}
\capbtabbox{%
\scalebox{\tableScale}{
 \begin{tabular}{c|c|c|c}
  & variant & adaptive & variant \\
  $c$ & $N_{\rm patience} = 5$ & $N_{\rm patience} = 7$ & $N_{\rm patience} = 9$  \\ \hline
 \input{analysis/tex/3_variation_cooldown_app_G2_f1slope.tex}
 \end{tabular}
}
}{%
  \caption{Gain $G(c; N_{\rm patience})$ for different $c \in \{ \text{II}, \text{III}, \text{IV} \}$ and $N_{\rm patience} \in \{ 5, 7, 9 \}$, in units of $10^{-5}$.  The last row shows the global values that are averaged over $c$. Note that in order to get $G(c; 5)$, we conducted experiments for $N_{\rm patience} = 3$. However, we do not show those results explicitly for the sake of brevity.}
 \label{tab:experiments_optimization_variation_number_of_epochs_f1slope}
}
\end{floatrow}
\end{figure}

In addition,  Tab.~\ref{tab:experiments_optimization_variation_number_of_epochs_2} shows the average relative uncertainties of the different variants.
The results indicate that an increase of $N_{\rm patience}$ has a stabilizing effect for very small datasets. 
For larger dataset scaling factors $x$ though, the effect of varying $N_{\rm patience}$ is small and ambiguous.  
This is a wanted effect, which we attribute to the fact that the early stopping mechanism ensures that the number of training epochs is of an appropriate size generally, while $N_{\rm patience}$ has little impact since it only affects the very end of the training process. Hence, the need for manually adjusting this parameter is reduced.  
In order to pick a reasonable parameter $N_{\rm patience}$ heuristically, we compute the gain
\begin{align}
 G(c; N_{\rm patience}) &:= \sum_{x \in X} x \cdot \left[ f_1(c; x; N_{\rm patience}) - f_1(c; x; N_{\rm patience} - 2) \right]
 \label{eq:gain}
\end{align}
to move from $N_{\rm patience} - 2$ to $N_{\rm patience}$ for each dataset-model combination $c$.
We choose $x$ as weights in the sum to account for the fact that the set $X$ (see Eq.~(\ref{eq:parameter_c})) places emphasis on rather small dataset scaling factors, while the larger values are more important for our considerations.
The results for the gain can be found in Tab.~\ref{tab:experiments_optimization_variation_number_of_epochs_f1slope}.
Increasing $N_{\rm patience}$ up to $7$ has a positive impact on performance for all $c$.  Above this value, we observe a negative or negligible effect on the performance.  Moreover, $N_{\rm patience} = 7$ leads to the best results for $x = 1.0$, i.e. when the full datasets are used.

\subsection{Early stopping with a constant learning rate \label{app:escd_variation_early_stopping}}

\begin{figure}[ht]
 \centering
 \includegraphics[scale=\figScale]{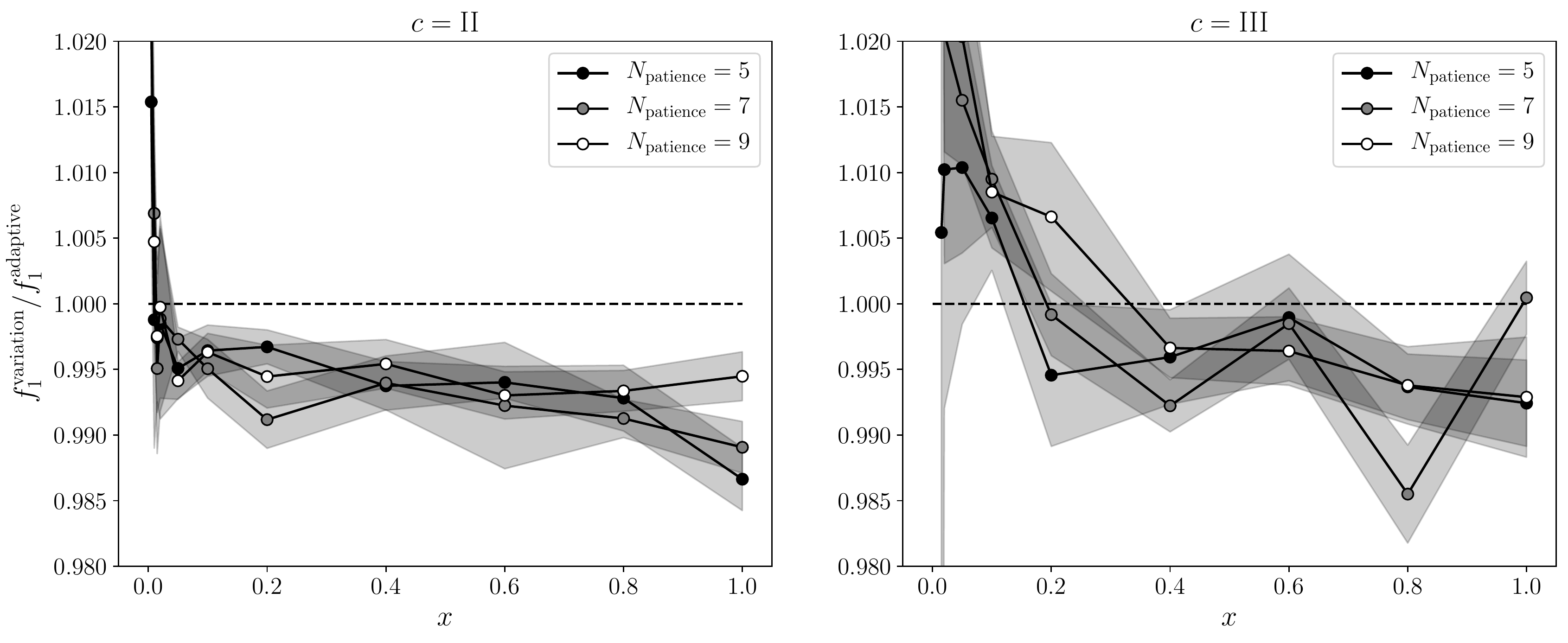}
 \caption{Ratio of the $f_1$ score for the alternative approach with a constant learning rate (see Tab.~\ref{tab:experiments_optimization_ablation_2}) and the $f_1$ score for adaptive approach with a hybrid learning rate (see Tab.~\ref{tab:experiments_optimization_variation_number_of_epochs}).}
 \label{fig:experiments_optimization_variation_constant}
\end{figure}

In App.~\ref{app:escd_variation_ablation}, we have seen that a linearly decreasing cool-down benefits the model performance. However, one could just as well keep the learning rate constant also after early stopping is triggered, see the right panel of Fig.~\ref{fig:escd_variations}.
We repeat a few experiments with this alternative approach, again while varying $N_{\rm patience} \in \{ 5, 7, 9 \}$ (cf.~App.~\ref{app:escd_variation_number_cool_down}). The results are shown in Tab.~\ref{tab:experiments_optimization_ablation_2}. In Fig.~\ref{fig:experiments_optimization_variation_constant}, we compare them visually to the corresponding results of the adaptive approach.
\begin{table}[ht]
 \centering
\scalebox{\tableScale}{
 \begin{tabular}{c|c|c|c|c|c|c|c|c|c|c}
  & & \multicolumn{3}{c|}{variant} & \multicolumn{3}{c|}{adaptive} & \multicolumn{3}{c}{variant} \\
  & & \multicolumn{3}{c|}{$N_{\rm patience} = 5$} & \multicolumn{3}{c|}{$N_{\rm patience} = 7$} & \multicolumn{3}{c}{$N_{\rm patience} = 9$} \\ \cline{3-11}
  \multirow{2}{*}{$c$} & \multirow{2}{*}{$x$} & \multirow{2}{*}{cv} & \multirow{2}{*}{$f_1$} & \multirow{2}{*}{$N_{\rm epochs}$} & \multirow{2}{*}{cv} & \multirow{2}{*}{$f_1$} & \multirow{2}{*}{$N_{\rm epochs}$} & \multirow{2}{*}{cv} & \multirow{2}{*}{$f_1$} & \multirow{2}{*}{$N_{\rm epochs}$}  \\	
  & & & & & & & & & & \\ \hline
 \input{analysis/tex/4_variation_cooldown_app_G3_constant.tex}
 \end{tabular}
}
 \caption{Results for the alternative approach with a constant learning rate.}
 \label{tab:experiments_optimization_ablation_2}
\end{table}
We find that using a constant learning rate has a moderate positive impact on performance and stability for very small datasets, with $x \lesssim 0.015$ for $c = \text{II}$ and $x \lesssim 0.1$ for $c = \text{III}$. However, for all other datasets sizes, the effect is the opposite.  We assign a higher value to the latter and thus conclude that the adaptive approach with a linearly decaying learning rate is preferable.

%% file: analysis/tex/2_variation_simplification_app_G1_mean_std.tex
\multirow{11}{*}{II}
 & 0.005 &  & 0.6341(111) & 30.4(2.8) &  & 0.6493(85) & 37.4(2.8) &  &{ \bf 0.6631(72) }& 42.8(2.9)  \\
 & 0.01 &  & 0.7306(61) & 18.2(8) &  & 0.7403(5) & 25.2(8) &  & 0.7404(46) & 29.8(1.2)  \\
 & 0.015 &  & 0.7508(1) & 14.6(9) &  & 0.7675(23) & 21.6(9) &  & 0.7706(31) & 25.4(7)  \\
 & 0.02 &  & 0.7780(46) & 13.2(3) &  & 0.7861(35) & 20.2(3) &  & 0.7853(41) & 23.4(7)  \\
 & 0.05 &  & 0.8419(35) & 9.4(5) &  & 0.8508(15) & 16.4(5) &  &{ \bf 0.8527(6) }& 19.6(1.1)  \\
 & 0.1 &  & 0.8576(3) & 7.0(6) &  & 0.8691(9) & 14.0(6) &  & 0.8692(11) & 16.0(5)  \\
 & 0.2 &  & 0.8773(3) & 5.6(4) &  & 0.8837(6) & 12.6(4) &  & 0.8837(8) & 14.0(3)  \\
 & 0.4 &  & 0.8902(15) & 5.2(3) &  & 0.8959(6) & 12.2(3) &  & 0.8955(8) & 12.6(2)  \\
 & 0.6 &  & 0.8948(8) & 4.6(2) &  & 0.9026(7) & 11.6(2) &  & 0.9026(7) & 11.6(2)  \\
 & 0.8 &  & 0.8967(19) & 4.4(2) &  & 0.9041(6) & 11.4(2) &  & 0.9041(6) & 11.4(2)  \\
 & 1.0 &  & 0.8984(9) & 4.0(0) &  & 0.9055(6) & 11.0(0) &  & 0.9055(6) & 11.0(0)  \\ \hline
\multirow{11}{*}{III}
 & 0.005 & 0 & --- & --- & 0 & --- & --- & 1 & --- & ---  \\
 & 0.01 & 0 & --- & --- & 0 & --- & --- & 1 & --- & ---  \\
 & 0.015 & 0 & --- & --- & 0 & --- & --- &  &{ \bf 0.3682(294) }& 35.6(3.3)  \\
 & 0.02 & 0 & --- & --- & 0 & --- & --- &  &{ \bf 0.5869(102) }& 42.0(2.5)  \\
 & 0.05 &  & 0.6993(61) & 11.8(4) &  & 0.7278(62) & 18.8(4) &  &{ \bf 0.7417(67) }& 22.4(1.5)  \\
 & 0.1 &  & 0.7798(62) & 8.2(3) &  & 0.8104(31) & 15.2(3) &  & 0.8108(29) & 16.2(5)  \\
 & 0.2 &  & 0.8255(26) & 5.8(2) &  & 0.8456(19) & 12.8(2) &  & 0.8456(19) & 12.8(2)  \\
 & 0.4 &  & 0.8482(29) & 4.4(2) &  & 0.8621(13) & 11.4(2) &  & 0.8621(13) & 11.4(2)  \\
 & 0.6 &  & 0.8582(13) & 4.0(0) &  & 0.8609(24) & 11.0(0) &  & 0.8609(24) & 11.0(0)  \\
 & 0.8 &  & 0.8593(31) & 4.6(2) &  & 0.8691(17) & 11.6(2) &  & 0.8691(17) & 11.6(2)  \\
 & 1.0 &  & 0.8655(34) & 3.8(2) &  & 0.8717(14) & 10.8(2) &  & 0.8717(14) & 10.8(2)

%% file: analysis/tex/2_variation_simplification_app_G1_stability.tex
 II & 0.02 & 0.0028 & 0.0013 & 0.0014  \\
 III & 0.1 & 0.0039 & 0.0023 & 0.0023  \\ \hline
 glob. &  & 0.0034 & 0.0018 & 0.0018

%% file: analysis/tex/3_variation_cooldown_app_G2_mean_std.tex
\multirow{11}{*}{II}
 & 0.005 &  & 0.6560(81) & 39.6(3.1) &  & 0.6631(72) & 42.8(2.9) &  & 0.6691(73) & 49.2(4.6)  \\
 & 0.01 &  & 0.7409(47) & 27.8(1.2) &  & 0.7404(46) & 29.8(1.2) &  & 0.7408(48) & 31.6(1.2)  \\
 & 0.015 &  & 0.7683(39) & 23.8(1.0) &  & 0.7706(31) & 25.4(7) &  & 0.7701(37) & 27.4(7)  \\
 & 0.02 &  & 0.7846(37) & 21.0(8) &  & 0.7853(41) & 23.4(7) &  & 0.7861(51) & 25.4(7)  \\
 & 0.05 &  & 0.8512(15) & 17.4(1.7) &  & 0.8527(6) & 19.6(1.1) &  & 0.8519(6) & 22.4(1.6)  \\
 & 0.1 &  & 0.8690(8) & 14.4(5) &  & 0.8692(11) & 16.0(5) &  & 0.8699(6) & 18.0(5)  \\
 & 0.2 &  & 0.8836(8) & 12.0(3) &  & 0.8837(8) & 14.0(3) &  & 0.8831(7) & 16.0(3)  \\
 & 0.4 &  & 0.8953(1) & 10.6(2) &  & 0.8955(8) & 12.6(2) &  & 0.8957(2) & 14.6(2)  \\
 & 0.6 &  & 0.9020(9) & 9.6(2) &  & 0.9026(7) & 11.6(2) &  & 0.9027(8) & 13.6(2)  \\
 & 0.8 &  & 0.9047(6) & 9.4(2) &  & 0.9041(6) & 11.4(2) &  & 0.9047(8) & 13.4(2)  \\
 & 1.0 &  & 0.9052(9) & 9.0(0) &  & 0.9055(6) & 11.0(0) &  & 0.9047(1) & 13.0(0)  \\ \hline
\multirow{11}{*}{III}
 & 0.005 & 0 & --- & --- & 1 & --- & --- & 1 & --- & ---  \\
 & 0.01 & 0 & --- & --- & 1 & --- & --- & 1 & --- & ---  \\
 & 0.015 &  & 0.3295(362) & 29.0(3.4) &  & 0.3682(294) & 35.6(3.3) &  & 0.3891(133) & 42.4(3.6)  \\
 & 0.02 &  & 0.5674(181) & 37.6(3.2) &  & 0.5869(102) & 42.0(2.5) &  & 0.5854(97) & 43.8(2.4)  \\
 & 0.05 &  & 0.7333(86) & 18.6(4) &  & 0.7417(67) & 22.4(1.5) &  & 0.7434(78) & 24.4(1.5)  \\
 & 0.1 &  & 0.8059(47) & 14.2(5) &  & 0.8108(29) & 16.2(5) &  & 0.8122(34) & 18.2(5)  \\
 & 0.2 &  & 0.8446(22) & 10.8(2) &  & 0.8456(19) & 12.8(2) &  &{ \bf 0.8484(19) }& 14.8(2)  \\
 & 0.4 &  & 0.8626(24) & 9.4(2) &  & 0.8621(13) & 11.4(2) &  & 0.8612(13) & 13.4(2)  \\
 & 0.6 &  & 0.8624(25) & 9.0(0) &  & 0.8609(24) & 11.0(0) &  & 0.8613(26) & 13.0(0)  \\
 & 0.8 &  & 0.8679(14) & 9.6(2) &  & 0.8691(17) & 11.6(2) &  & 0.8673(19) & 13.6(2)  \\
 & 1.0 &  & 0.8714(14) & 8.8(2) &  & 0.8717(14) & 10.8(2) &  & 0.8708(2) & 12.8(2)  \\ \hline
\multirow{11}{*}{IV}
 & 0.005 & 0 & --- & --- & 0 & --- & --- & 0 & --- & ---  \\
 & 0.01 & 2 & --- & --- & 2 & --- & --- & 2 & --- & ---  \\
 & 0.015 &  & 0.3711(78) & 33.4(2.6) &  & 0.3798(81) & 34.0(1.1) &  &{ \bf 0.3932(52) }& 42.4(3.3)  \\
 & 0.02 &  & 0.4489(179) & 30.8(1.7) &  & 0.4566(187) & 32.6(1.5) &  & 0.4595(188) & 34.6(1.5)  \\
 & 0.05 &  & 0.6030(8) & 18.6(6) &  & 0.6082(78) & 20.6(6) &  & 0.6111(86) & 22.6(6)  \\
 & 0.1 &  & 0.6751(89) & 17.2(1.5) &  & 0.6750(31) & 18.6(8) &  & 0.6759(29) & 19.2(5)  \\
 & 0.2 &  & 0.7419(29) & 13.0(8) &  & 0.7477(32) & 15.0(8) &  & 0.7470(8) & 17.8(8)  \\
 & 0.4 &  & 0.7780(16) & 10.8(3) &  & 0.7833(21) & 12.8(3) &  & 0.7842(13) & 14.8(3)  \\
 & 0.6 &  & 0.7971(25) & 9.8(3) &  & 0.7923(26) & 11.8(3) &  & 0.7957(22) & 13.8(3)  \\
 & 0.8 &  & 0.8034(27) & 9.2(2) &  & 0.8057(14) & 11.2(2) &  & 0.8066(31) & 13.6(5)  \\
 & 1.0 &  & 0.8038(24) & 9.6(4) &  & 0.8067(34) & 11.6(4) &  & 0.8033(13) & 13.6(4)

%% file: analysis/tex/3_variation_cooldown_app_G2_stability.tex
 II & 0.02 & 0.0015 & 0.0014 & 0.0015  \\
 III & 0.1 & 0.0029 & 0.0023 & 0.0026  \\
 IV & 0.1 & 0.0048 & 0.0035 & 0.0025  \\ \hline
 glob. &  & 0.0031 & 0.0024 & 0.0022

%% file: analysis/tex/3_variation_cooldown_app_G2_f1slope.tex
II & 0.87 & 0.38 & -0.19  \\
III & 3.22 & 2.77 & -1.72  \\
IV & 13.16 & 7.01 & 0.09  \\ \hline
glob. & 5.75 & 3.38 & -0.60

%% file: analysis/tex/4_variation_cooldown_app_G3_constant.tex
\multirow{11}{*}{II}
 & 0.005 &  & 0.6733(45) & 44.4(2.7) &  & 0.6774(54) & 50.2(3.6) &  & 0.6781(49) & 52.2(3.6)  \\
 & 0.01 &  & 0.7395(64) & 26.6(1.0) &  & 0.7455(44) & 28.6(1.0) &  & 0.7439(45) & 30.6(1.0)  \\
 & 0.015 &  & 0.7686(29) & 23.2(7) &  & 0.7668(44) & 25.2(7) &  & 0.7687(38) & 27.2(7)  \\
 & 0.02 &  & 0.7841(47) & 21.8(1.0) &  & 0.7844(45) & 23.8(1.0) &  & 0.7851(44) & 25.8(1.0)  \\
 & 0.05 &  & 0.8485(19) & 17.6(1.1) &  & 0.8504(6) & 19.6(1.1) &  & 0.8477(11) & 21.6(1.1)  \\
 & 0.1 &  & 0.8661(14) & 14.0(5) &  & 0.8649(17) & 16.0(5) &  & 0.8660(8) & 18.0(5)  \\
 & 0.2 &  &{ \bf 0.8808(9) }& 12.0(3) &  & 0.8759(18) & 14.0(3) &  & 0.8788(2) & 16.0(3)  \\
 & 0.4 &  & 0.8899(15) & 10.8(3) &  & 0.8901(17) & 12.8(3) &  & 0.8914(15) & 14.8(3)  \\
 & 0.6 &  & 0.8972(9) & 9.6(2) &  & 0.8956(43) & 11.6(2) &  & 0.8963(15) & 13.6(2)  \\
 & 0.8 &  & 0.8976(22) & 9.4(2) &  & 0.8962(12) & 11.4(2) &  & 0.8981(13) & 13.4(2)  \\
 & 1.0 &  & 0.8934(21) & 9.0(0) &  & 0.8956(17) & 11.0(0) &  &{ \bf 0.9005(16) }& 13.0(0)  \\ \hline
\multirow{11}{*}{III}
 & 0.005 & 1 & --- & --- & 1 & --- & --- & 1 & --- & ---  \\
 & 0.01 & 2 & --- & --- & 3 & --- & --- & 2 & --- & ---  \\
 & 0.015 &  & 0.3702(307) & 33.0(4.5) &  & 0.4156(187) & 40.4(2.4) &  & 0.4251(151) & 43.8(2.7)  \\
 & 0.02 &  & 0.5929(88) & 39.8(2.4) &  & 0.5990(83) & 41.8(2.4) &  & 0.6046(9) & 45.4(2.6)  \\
 & 0.05 &  & 0.7494(73) & 21.8(1.9) &  & 0.7532(7) & 23.8(1.9) &  & 0.7568(52) & 25.8(1.9)  \\
 & 0.1 &  & 0.8161(22) & 15.4(9) &  & 0.8185(18) & 17.4(9) &  & 0.8177(25) & 19.4(9)  \\
 & 0.2 &  & 0.8410(43) & 10.8(2) &  & 0.8449(21) & 12.8(2) &  &{ \bf 0.8512(45) }& 14.8(2)  \\
 & 0.4 &  & 0.8586(29) & 9.4(2) &  & 0.8554(13) & 11.4(2) &  & 0.8592(16) & 13.4(2)  \\
 & 0.6 &  & 0.8600(36) & 9.0(0) &  & 0.8596(11) & 11.0(0) &  & 0.8578(9) & 13.0(0)  \\
 & 0.8 &  & 0.8636(16) & 9.6(2) &  & 0.8565(29) & 11.6(2) &  & 0.8637(21) & 13.6(2)  \\
 & 1.0 &  & 0.8651(26) & 8.8(2) &  &{ \bf 0.8721(21) }& 10.8(2) &  & 0.8655(38) & 12.8(2)

%% file: tex/appendix_finetuning_ablation.tex
As we have seen in Sec.~\ref{sec:results_fine_tuning}, the $f_1$ performance using the adaptive approach is greater or equal to the performance using the fixed epoch variants (cf.~Eq.~(\ref{eq:result_performance})).  
Recall that the main differences of the adaptive approach with respect to the fixed epoch approach are the early stopping mechanism, the adaptive number $N_{\rm epochs}$ of training epochs and the hybrid learning rate schedule that employs a constant learning rate followed by a linearly decaying cool-down phase at the end of the training process.
 
\begin{figure}[ht]
 \centering
 \includegraphics[scale=\figScaleThree]{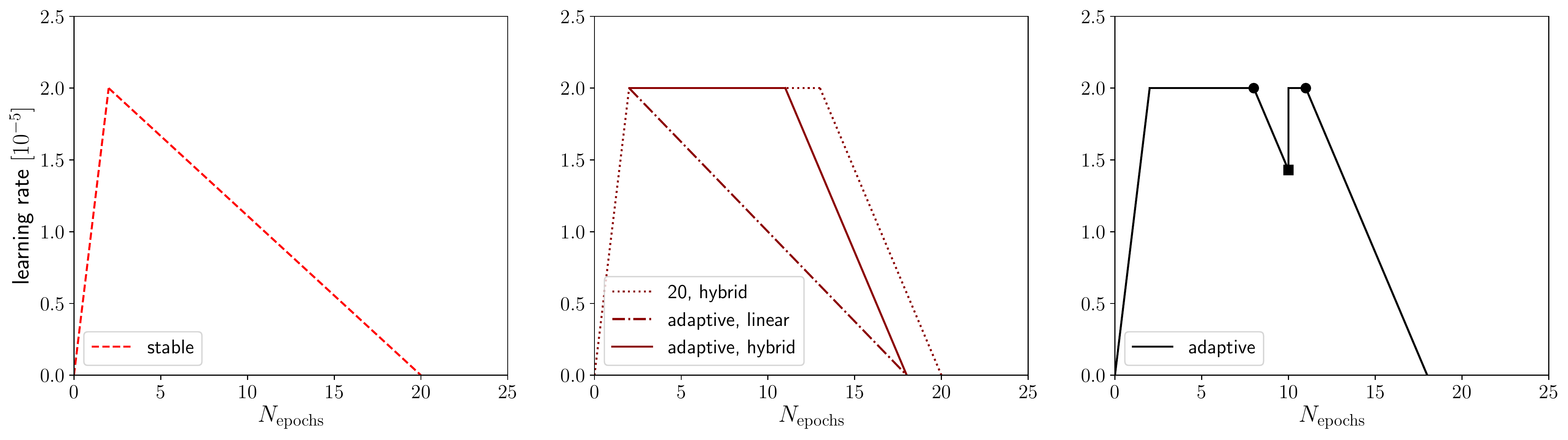}
 \caption{Ablation variants (center) compared to the stable (left) and adaptive (right) learning rate schedules.  Circles indicate the beginning of an early stopping stage, while the square represents training resumption. }
 \label{fig:experiments_optimization_ablation}
\end{figure}

In this appendix, we aim to disentangle the impact of these features and study how they affect the performance individually. To this end, we rerun the model training with three different setups that use different subsets of the aforementioned feature that make the adaptive approach distinct. 
We start from the stable fine-tuning approach. 
Then, firstly, we replace the linearly decaying learning rate schedule by the hybrid learning rate schedule of the adaptive approach. 
Secondly,  instead of training for $N_{\rm epochs} = 20$ epochs, we use the number of training epochs we found previously using the adaptive approach, rounded to the nearest integer. 
Finally, we combine those two changes, i.e.  we use both a hybrid learning rate schedule and an adaptive number of training epochs. This last variant can be considered a static version of the adaptive approach that misses only the early stopping mechanism. 
The different ablation variants are illustrated in Fig.~\ref{fig:experiments_optimization_ablation}.

Note, however, that the three ablation variants are no serious alternatives in practice, for different reasons. The variant with a fixed number of training epochs and a hybrid learning rate schedule contains two parameters with regard to the number of training epochs, $N_{\rm epochs} = 20$ and $N_{\rm patience} = 7$, which is a significant disadvantage compared to the other approaches. 
The ablation variants that use the number of training epochs from the adaptive approach cannot be applied in practice, as the number of training epochs is a priori unknown.  
Hence, it has to be emphasized that the reason we study the ablation variants here is mostly to understand the different individual effects of the three features that differentiate the adaptive approach from the other approaches.  
This is also why we discuss them separately in this section instead of treating them as yet additional variants of the adaptive approach like the ones presented in Sec.~\ref{app:escd_variations}. 

\begin{figure}[p]
 \centering
 \includegraphics[scale=\figScaleThree]{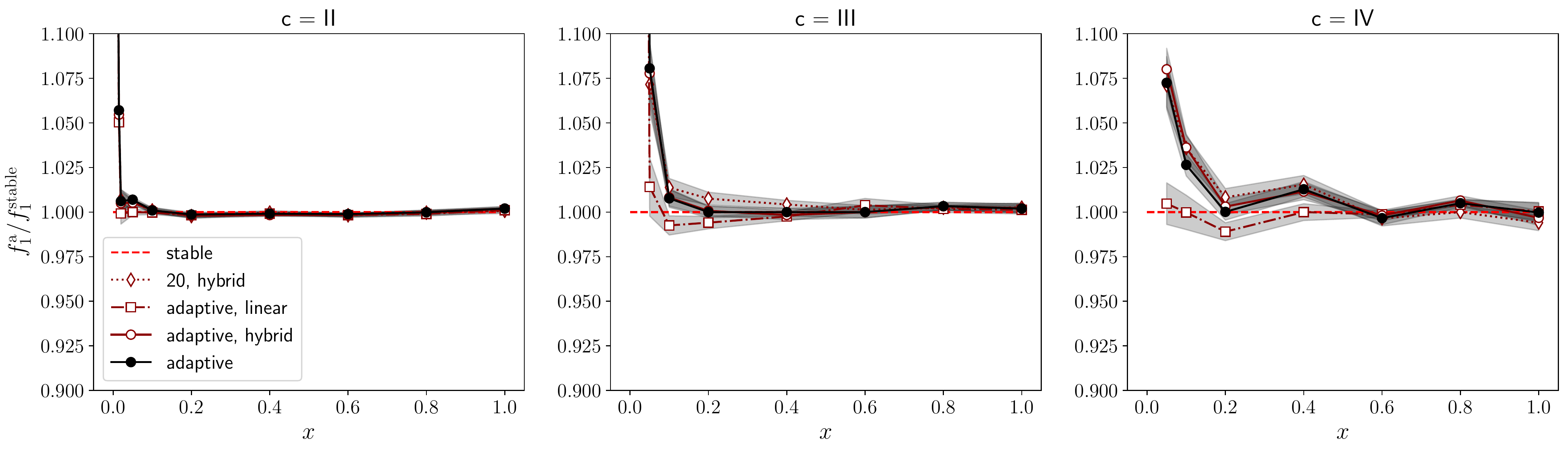}
 \caption{Relative $f_1$ score of the different ablation variants and the adaptive approach with respect to the stable approach. Note that no results are shown for small $x$ and $c \in \{ \text{III},  \text{IV} \}$ as the stable approach does not converge. The ablation variants, however, show significantly improved behavior in those cases. }
 \label{fig:experiments_optimization_ablation_relative}
\end{figure}

The fine-tuning results of the ablation variants for $c \in \{ \text{II},  \text{III}, \text{IV} \}$ are listed in Tab.~\ref{tab:experiments_optimization_ablation}. In Fig.~\ref{fig:experiments_optimization_ablation_relative}, we compare the $f_1$ score for the ablation variants and the adaptive approach with the stable approach.
\begin{table}[p]
 \centering
\scalebox{\tableScale}{
  \begin{tabular}{c|c||c|c|c|c|c|c}
  \multicolumn{2}{c||}{approach} & stable & \multicolumn{3}{c|}{ablation} & \multicolumn{2}{c}{adaptive} \\ \hline
  \multicolumn{2}{c||}{early stopping} & no & no & no & no & \multicolumn{2}{c}{yes} \\
  \multicolumn{2}{c||}{training epochs} & 20 & 20 & adaptive & adaptive & \multicolumn{2}{c}{adaptive} \\
  \multicolumn{2}{c||}{learning rate schedule} & linear & hybrid & linear & hybrid & \multicolumn{2}{c}{hybrid} \\ \hline \hline
  \multirow{2}{*}{$c$} & \multirow{2}{*}{$x$} & \multirow{2}{*}{$f_1$} & \multirow{2}{*}{$f_1$} & \multirow{2}{*}{$f_1$} & \multirow{2}{*}{$f_1$} & \multirow{2}{*}{$f_1$} & \multirow{2}{*}{$N_{\rm epochs}$} \\ 
  & & & & & & \\ \hline    
  \input{analysis/tex/6_app_H.tex}
 \end{tabular}
}
 \caption{Fine-tuning results for the ablations variants.  We abstain from showing the number of converged runs here for the sake of brevity.  Note that $N_{\rm epochs}$ is the number of epochs found using the adaptive approach.  The ablation variants with adaptive training epochs use the same numbers, rounded to the nearest integer. More detailed results for the stable and adaptive fine-tuning approaches can be found in Tab.~\ref{tab:experiments_optimization_ALL}. }
 \label{tab:experiments_optimization_ablation}
\end{table}
We observe the following patterns:
If we replace the linear learning rate schedule of the stable approach with a hybrid learning rate schedule while keeping the number of training epochs fixed, we find that the $f_1$ score benefits for small and moderately large datasets ($x^{\rm II} \leq 0.05$,  $x^{\rm III} \leq 0.4$, $x^{\rm IV} \leq 0.4$). 
The ablation variant with a purely linear decay rate but varying $N_{\rm epochs}$ leads to better results for very small datasets ($x^{\rm II} \leq 0.015$,  $x^{\rm III} \leq 0.02$, $x^{\rm IV} \leq 0.02$), those that benefit from $N_{\rm epochs} > 20$.
The combination of an adaptive number of training epochs with a hybrid learning rate schedule combines these two positive effects and significantly improves the results for small dataset scaling factors $x$, while maintaining the performance of the stable approach for large datasets.
In fact, the difference between this ablation variant and the full adaptive approach (using early stopping) is very small compared to the associated statistical uncertainties. 

We conclude that the hybrid learning rate schedule has a positive impact on the performance especially for small datasets.  However, we stress again that only our adaptive approach allows to use it without introducing an additional parameter.
Furthermore, we find that the adaptive number of training epochs has two effects. It improves the performance for very small datasets, while making the training more efficient for larger datasets.

%% file: analysis/tex/6_app_H.tex
\multirow{11}{*}{II}
 & 0.005 & 0.1987(263) & 0.4840(277) & 0.6020(33) & 0.6632(27) & 0.6631(72) & 42.8(2.9)  \\
 & 0.01 & 0.5923(6) & 0.7231(18) & 0.7258(16) & 0.7441(45) & 0.7404(46) & 29.8(1.2)  \\
 & 0.015 & 0.7290(69) & 0.7670(23) & 0.7657(26) & 0.7688(46) & 0.7706(31) & 25.4(7)  \\
 & 0.02 & 0.7805(38) & 0.7860(33) & 0.7799(36) & 0.7844(42) & 0.7853(41) & 23.4(7)  \\
 & 0.05 & 0.8468(11) & 0.8513(5) & 0.8468(11) & 0.8513(5) &{ \bf 0.8527(6) }& 19.6(1.1)  \\
 & 0.1 & 0.8684(8) & 0.8692(8) & 0.8682(7) & 0.8694(12) & 0.8692(11) & 16.0(5)  \\
 & 0.2 & 0.8849(9) & 0.8829(8) & 0.8833(6) & 0.8831(8) & 0.8837(8) & 14.0(3)  \\
 & 0.4 & 0.8963(6) & 0.8961(8) & 0.8955(9) & 0.8948(4) & 0.8955(8) & 12.6(2)  \\
 & 0.6 &{ \bf 0.9036(4) }& 0.9020(1) & 0.9021(9) & 0.9021(7) & 0.9026(7) & 11.6(2)  \\
 & 0.8 & 0.9042(6) & 0.9039(12) & 0.9033(1) & 0.9042(9) & 0.9041(6) & 11.4(2)  \\
 & 1.0 & 0.9037(9) & 0.9043(1) & 0.9047(5) & 0.9055(6) & 0.9055(6) & 11.0(0)  \\ \hline
\multirow{11}{*}{III}
 & 0.005 & --- & --- & --- & --- & --- & ---  \\
 & 0.01 & --- & --- & --- & --- & --- & ---  \\
 & 0.015 & --- & 0.2090(259) & 0.2634(172) & 0.3878(123) & 0.3682(294) & 35.6(3.3)  \\
 & 0.02 & 0.2430(84) & 0.4341(84) & 0.5226(9) & 0.5899(68) & 0.5869(102) & 42.0(2.5)  \\
 & 0.05 & 0.6864(103) & 0.7355(75) & 0.6961(86) & 0.7398(92) & 0.7417(67) & 22.4(1.5)  \\
 & 0.1 & 0.8046(35) &{ \bf 0.8157(29) }& 0.7986(33) & 0.8112(27) & 0.8108(29) & 16.2(5)  \\
 & 0.2 & 0.8456(22) &{ \bf 0.8520(24) }& 0.8406(23) & 0.8462(15) & 0.8456(19) & 12.8(2)  \\
 & 0.4 & 0.8621(15) &{ \bf 0.8658(15) }& 0.8600(17) & 0.8607(18) & 0.8621(13) & 11.4(2)  \\
 & 0.6 & 0.8609(18) & 0.8620(14) & 0.8642(29) & 0.8609(24) & 0.8609(24) & 11.0(0)  \\
 & 0.8 & 0.8662(11) & 0.8683(8) & 0.8678(15) & 0.8690(15) & 0.8691(17) & 11.6(2)  \\
 & 1.0 & 0.8700(2) & 0.8722(14) & 0.8711(1) & 0.8719(13) & 0.8717(14) & 10.8(2)  \\ \hline
\multirow{11}{*}{IV}
 & 0.005 & --- & --- & --- & --- & --- & ---  \\
 & 0.01 & --- & --- & --- & --- & --- & ---  \\
 & 0.015 & --- & --- & 0.2704(145) & 0.3792(99) & 0.3798(81) & 34.0(1.1)  \\
 & 0.02 & --- & 0.3224(155) & 0.3503(19) & 0.4556(162) & 0.4566(187) & 32.6(1.5)  \\
 & 0.05 & 0.5671(49) & 0.6074(62) & 0.5698(6) & 0.6125(6) & 0.6082(78) & 20.6(6)  \\
 & 0.1 & 0.6576(39) & 0.6812(38) & 0.6575(58) & 0.6814(42) & 0.6750(31) & 18.6(8)  \\
 & 0.2 & 0.7476(18) &{ \bf 0.7538(34) }& 0.7394(35) & 0.7499(3) & 0.7477(32) & 15.0(8)  \\
 & 0.4 & 0.7734(33) & 0.7853(3) & 0.7734(26) & 0.7823(24) & 0.7833(21) & 12.8(3)  \\
 & 0.6 & 0.7950(17) & 0.7922(31) & 0.7940(2) & 0.7935(19) & 0.7923(26) & 11.8(3)  \\
 & 0.8 & 0.8017(14) & 0.8017(24) & 0.8050(19) & 0.8069(18) & 0.8057(14) & 11.2(2)  \\
 & 1.0 & 0.8069(33) & 0.8021(24) & 0.8072(31) & 0.8043(21) & 0.8067(34) & 11.6(4)

%% file: tex/appendix_finetuning_additional.tex
\begin{figure}[ht!]
\begin{minipage}[t]{0.45\linewidth}
 \centering
 \strut\vspace*{-\baselineskip}\newline\newline\newline
\scalebox{\tableScale}{
\begin{tabular}{c|c|c|c|c}
  $c$  & $x$ & original & stable & adaptive \\ \hline
 \input{analysis/tex/tensorboard_learning_curves_cII_final.tex}
\end{tabular}%
}
\end{minipage}%
\hspace{0.02\linewidth}
\begin{minipage}[t]{0.45\linewidth}
 \centering
  \strut\vspace*{-\baselineskip}\newline
 \includegraphics[scale=\figScale]{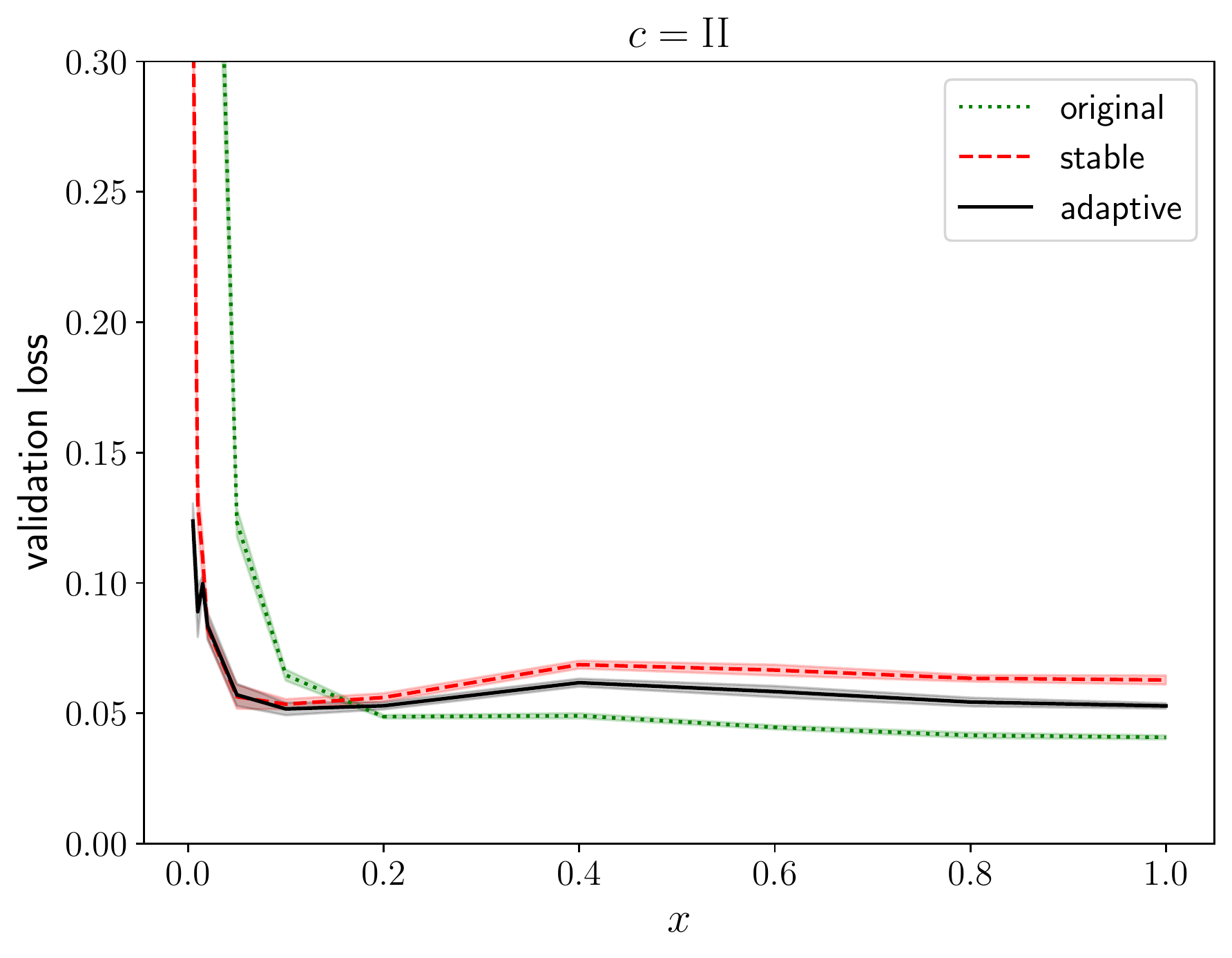}
\end{minipage}%
\caption{Validation loss for $c = \text{II}$ in dependence of $x \in X$ and for all $a \in A$.}
\label{fig:tensorboard_learning_curves_cII_final}
\end{figure}
\begin{figure}[ht!]
 \centering
 \includegraphics[scale=\figScale]{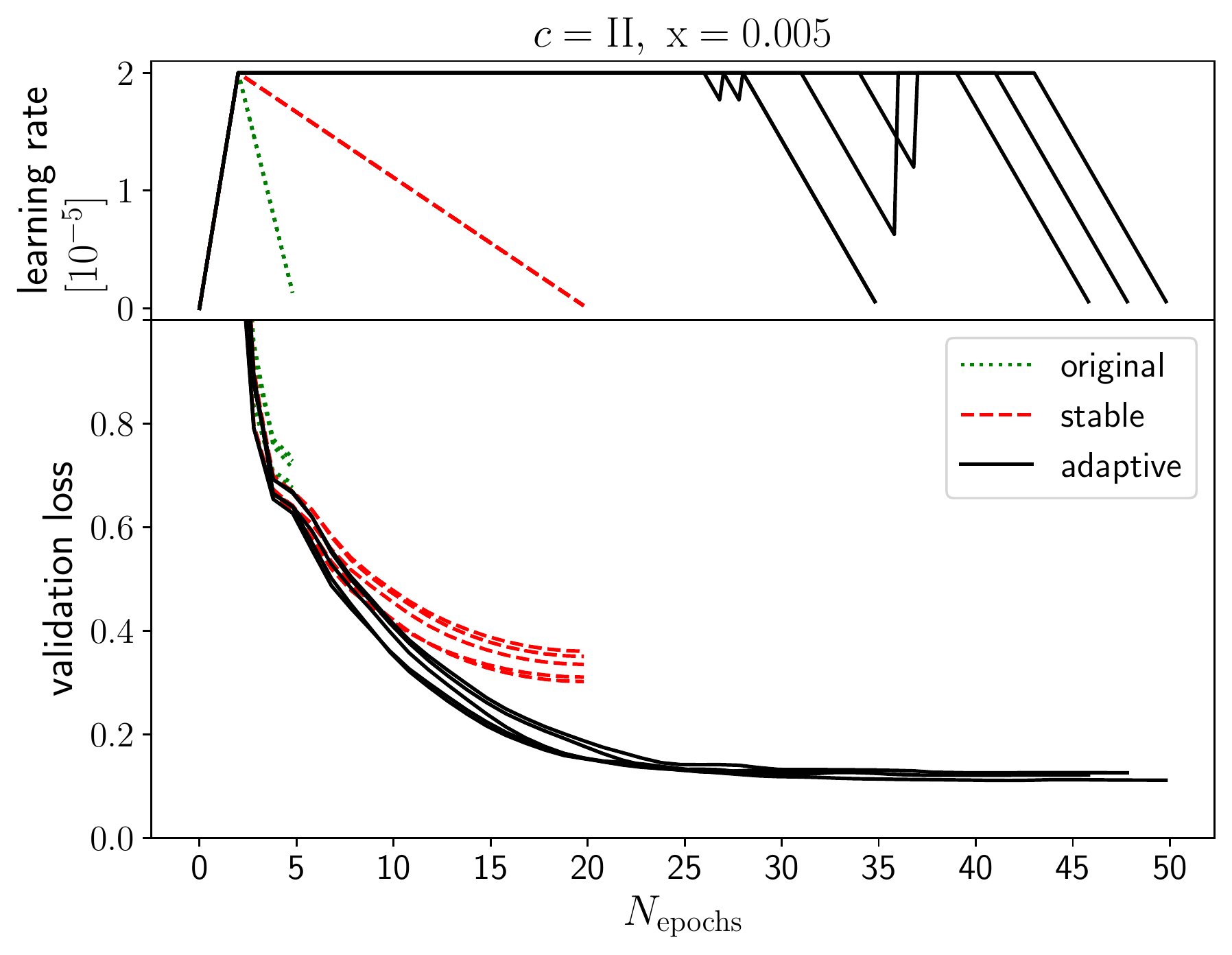}
 \hspace{5mm}
 \includegraphics[scale=\figScale]{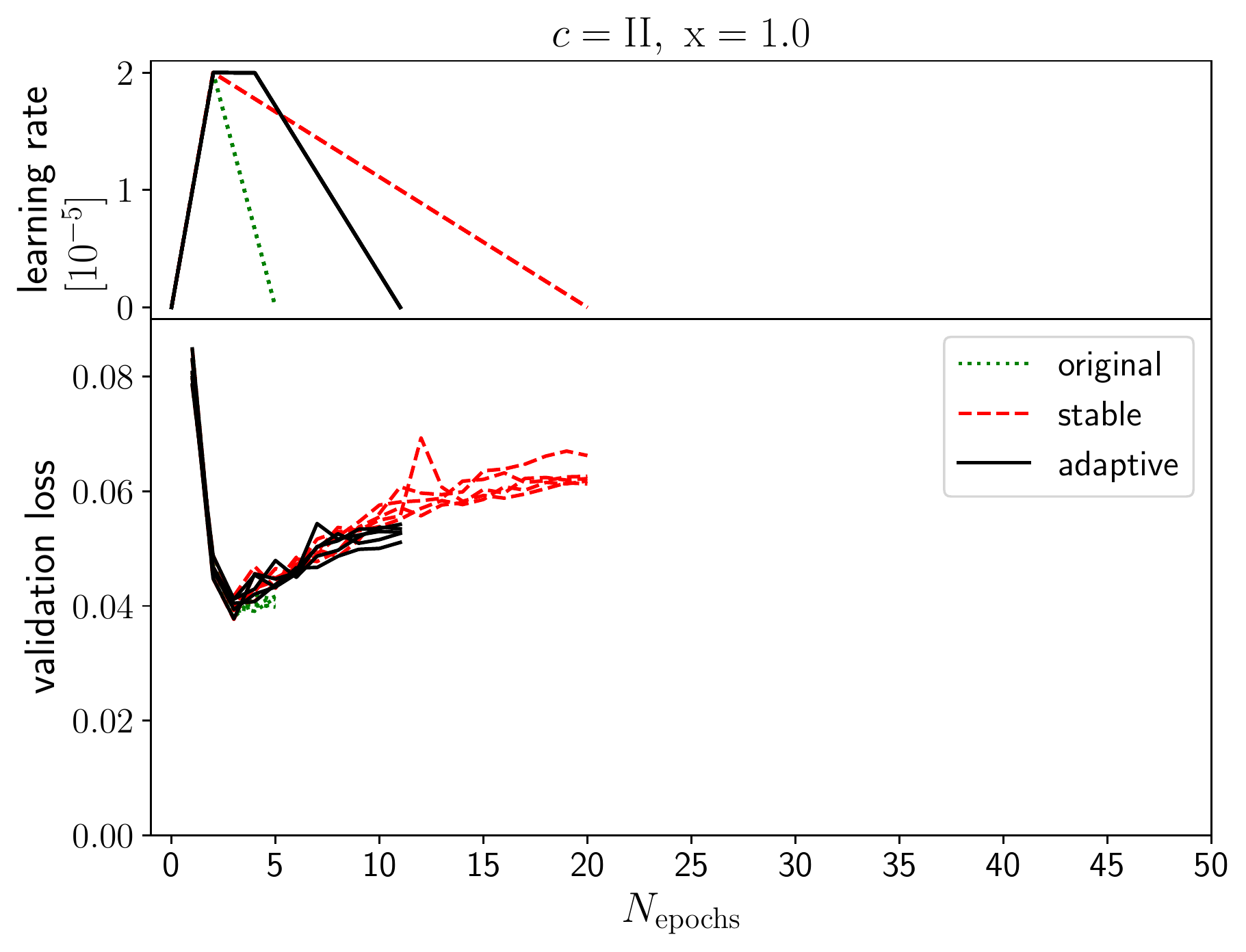}
 \caption{Learning curves for the different fine-tuning approaches $a \in A$, the dataset-model combination $c = \text{II}$ and the dataset scaling factor $x = 0.005$ (left panel) as well as $x = 1.0$ (right panel).  For each approach and each of the five runs, the learning rate (top) and validation loss (bottom) are plotted against the training epochs. Note that the scales for the validation loss are different.  For the adaptive approach and $x = 0.005$, we observe that training is often resumed during the cool-down phase as a new minimum of the validation loss is encountered (cf.~Sec.~\ref{subsec:adaptive_approach}). 
}
 \label{fig:experiments_learning_curves}
\end{figure}

In Fig.~\ref{fig:tensorboard_learning_curves_cII_final}, we display the validation loss obtained at the end of training, for $c = \text{II}$,  all $a \in A$ and all $x \in X$.   
The validation loss \textit{learning curves} for two representative examples, $x = 0.005$ and $x = 1.0$, are shown in Fig.~\ref{fig:experiments_learning_curves}.

We find that for a small dataset, the fixed epoch variants achieve a suboptimal validation loss as the training is stopped prematurely.  
In contrast, for a large dataset, all approaches encounter a minimum of the validation loss (which is located at around 3 training epochs for $x = 1.0$).  The stable variant continues training for several additional epochs after the minimum is reached, just like the adaptive fine-tuning approach does by design.  Comparing this behavior to the $f_1$ scores given in Tab.~\ref{tab:experiments_optimization_ALL}, for instance $f_1^{\rm adap.} > f_1^{\rm stab.} > f_1^{\rm orig.}$ for $x = 0.005$ and $f_1^{\rm adap.} \gtrsim f_1^{\rm stab.} > f_1^{\rm orig.}$ for $x = 1.0$, we infer that the model performance benefits from training beyond the validation loss minimum. This is also in accordance with the results of Sec.~\ref{app:escd_variation_ablation}, where we showed that the cool-down phase of the adaptive approach has a positive effect on the model performance.

%% file: analysis/tex/tensorboard_learning_curves_cII_final.tex
\multirow{11}{*}{II}
 & 0.005 & 0.7076(23) & 0.3319(225) & 0.1238(68)  \\
 & 0.01 & 0.5400(182) & 0.1299(89) & 0.0890(98)  \\
 & 0.015 & 0.4868(235) & 0.1096(84) & 0.0998(3)  \\
 & 0.02 & 0.5295(433) & 0.0825(37) & 0.0836(52)  \\
 & 0.05 & 0.1232(55) & 0.0564(46) & 0.0571(43)  \\
 & 0.1 & 0.0647(22) & 0.0535(2) & 0.0517(24)  \\
 & 0.2 & 0.0487(7) & 0.0561(16) & 0.0529(14)  \\
 & 0.4 & 0.0491(1) & 0.0687(15) & 0.0618(15)  \\
 & 0.6 & 0.0447(8) & 0.0666(21) & 0.0584(22)  \\
 & 0.8 & 0.0416(11) & 0.0634(13) & 0.0544(17)  \\
 & 1.0 & 0.0408(7) & 0.0628(18) & 0.0529(1)

%% file: tex/appendix_finetuning_dependency.tex
Given a dataset-model combination $c \in C$, the fine-tuning performance $f_1$ depends on the dataset scaling factor $x \in X$ as well as the fine-tuning approach $a \in A$. The latter impacts the results mainly through the number of training epochs $N_{\rm epochs}$, as we have seen in App.~\ref{app:finetuning_ablation}.  

\begin{figure}[ht]
 \centering
 \includegraphics[scale=\figScale]{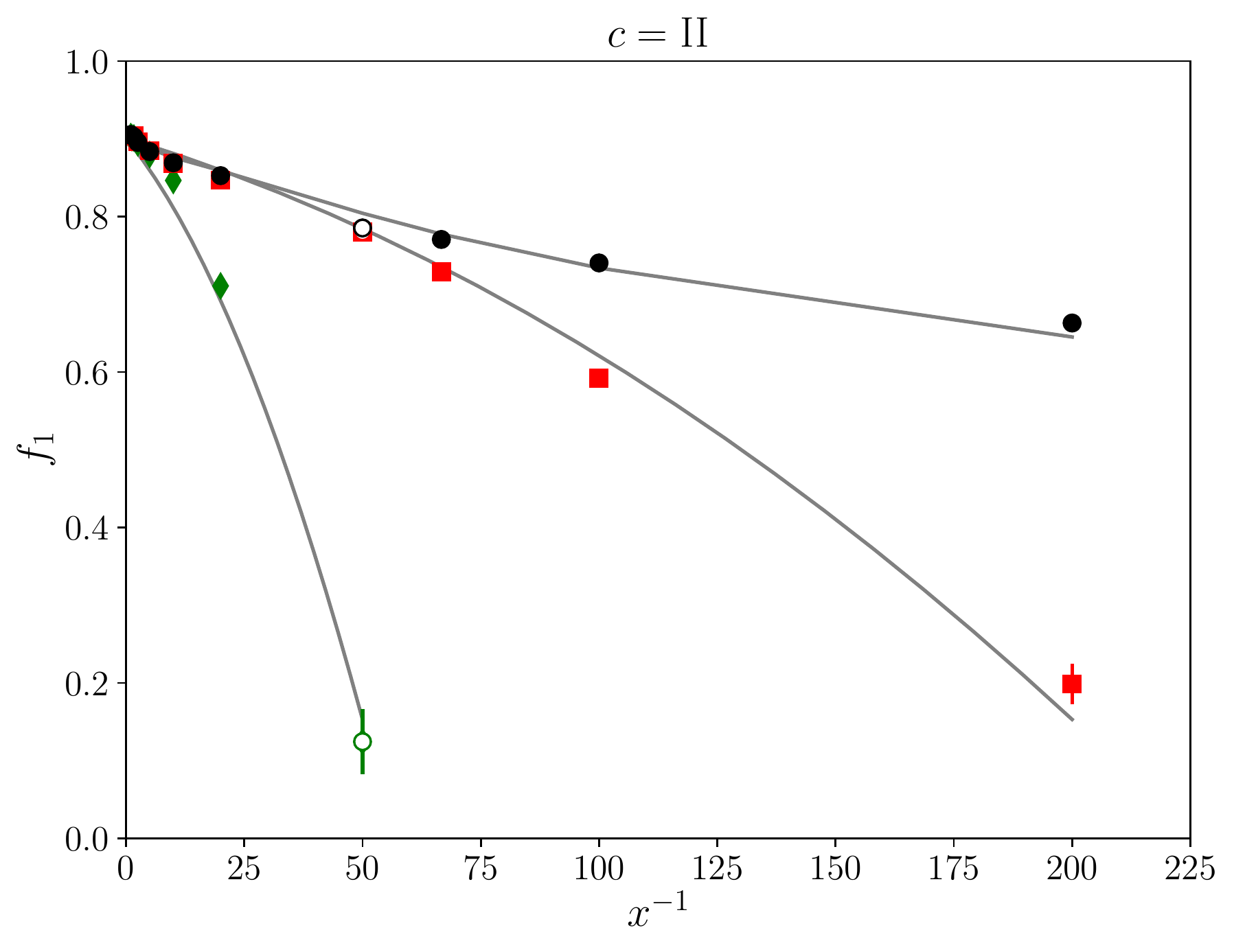}
 \includegraphics[scale=\figScale]{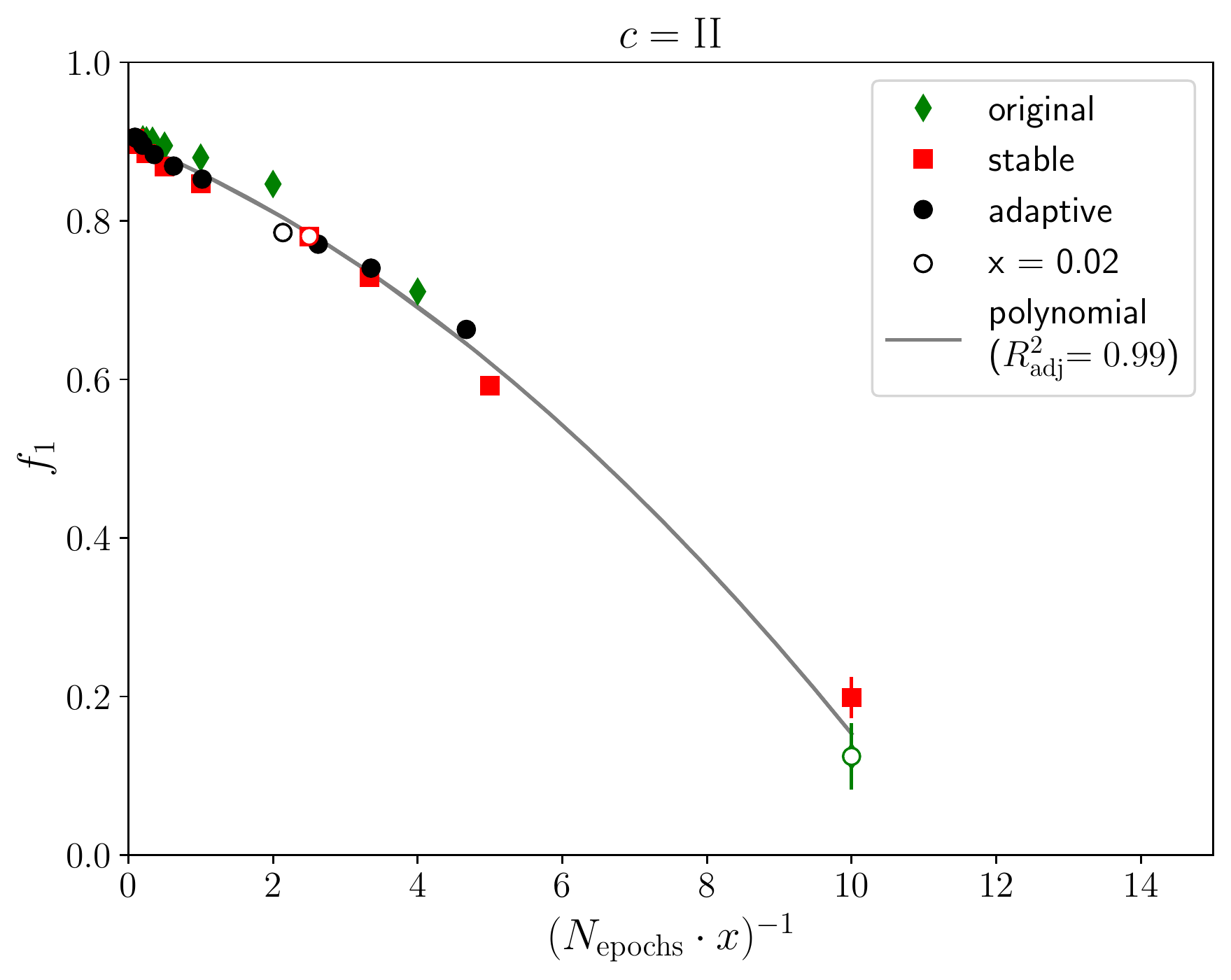}
 \caption{$f_1$ as a function of $ x ^{-1}$ (left panel) and $( N_{\rm epochs} \cdot x )^{-1}$ (right panel) for $c = \text{II}$ and all $a \in A$. The solid line represents the polynomial fit, Eq.~(\ref{eq:polynomial_fit}). For $x = 0.02$, the results for the different fine-tuning approaches are displayed as empty circles for comparison. }
 \label{fig:polynomial_fit_x_and_Nx}
\end{figure}

In this section, we study the functional dependency of $f_1 \equiv f_1 (x, N_{\rm epochs})$. For the example of $c = \text{II}$, we show the results for $f_1$ as a function of the inverse dataset scaling factor $x^{-1}$ in the left panel of Fig.~\ref{fig:polynomial_fit_x_and_Nx}.
Adding more data corresponds to going from right to left in the plot.  While this is done, the $f_1$ score constantly improves but eventually saturates, i.e.  the effect of more data drastically fades. The trajectory of each fine-tuning approach asymptotically approaches the value $f_1 (x^{-1} \to 0)$ which corresponds to infinite training data.
The impact of $N_{\rm epochs}$ can be discussed using the example of $x=0.02$, or $x^{-1}=50$.  The $f_1$ score improves a lot between $N_{\rm epochs} = 5$ (original approach) and $N_{\rm epochs} = 20$ (stable approach).  However, the result for $N_{\rm epochs} = 23.4$ (adaptive approach) is only slightly better, as the point of convergence is reached where additional training does not have an effect any longer\footnote{Strictly speaking, at some point overfitting will lead to a decrease of the considered $f_1$ score on the test dataset \citep{mosbach2021stability}.}.  
Hence, we find empirically that $f_1$ is a continuously increasing function of $x$ and $N_{\rm epochs}$ unless the latter exceeds a certain threshold $T(x)$:
\begin{subequations}
\begin{eqnarray}
  \frac{\partial f_1}{\partial x} & \geq & 0
\label{eq:derivative_f1_x}
\end{eqnarray}
\begin{eqnarray}
  \frac{\partial f_1}{\partial N_{\rm epochs}} \bigg|_{N_{\rm epochs} \leq T(x)} & \geq & 0 
  \label{eq:derivative_f1_Nepochs_below_T} \\
  \frac{\partial f_1}{\partial N_{\rm epochs}} \bigg|_{N_{\rm epochs} > T(x)} & = & 0
  \label{eq:derivative_f1_Nepochs_above_T}
\end{eqnarray}
\label{eq:derivative_f1}
\end{subequations}

\begin{figure}[ht]
 \centering
 \includegraphics[scale=\figScale]{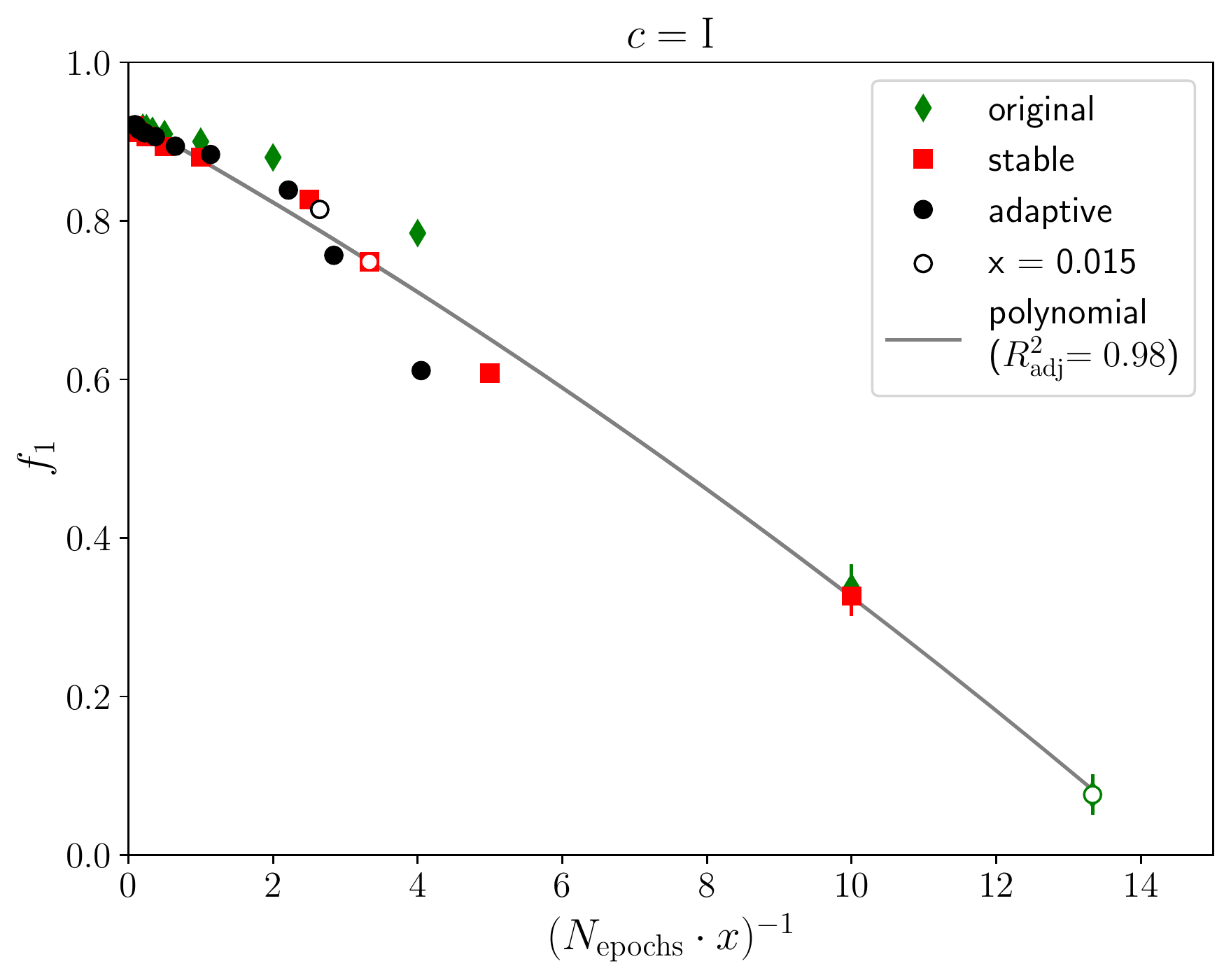}
 \includegraphics[scale=\figScale]{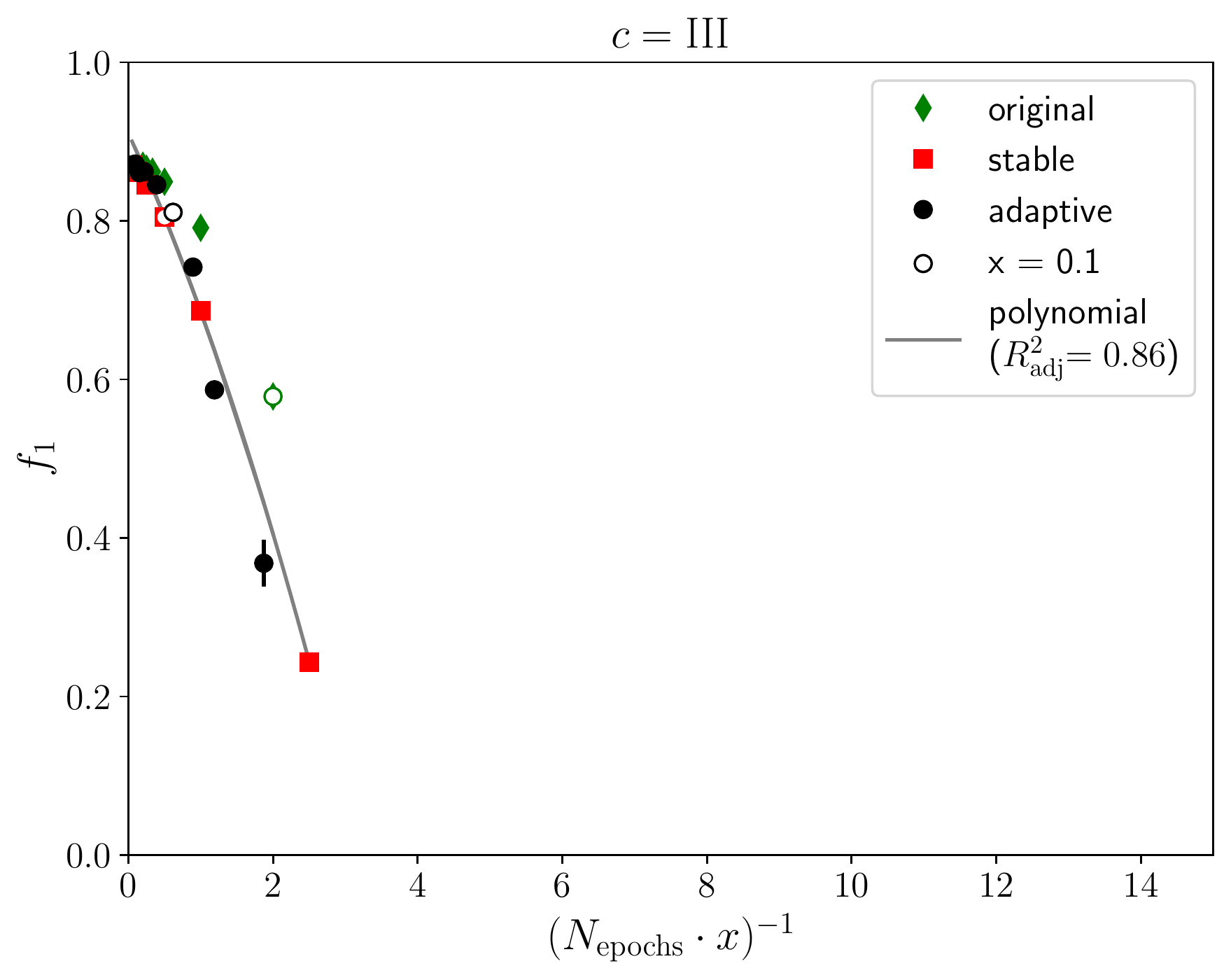} \\
 \includegraphics[scale=\figScale]{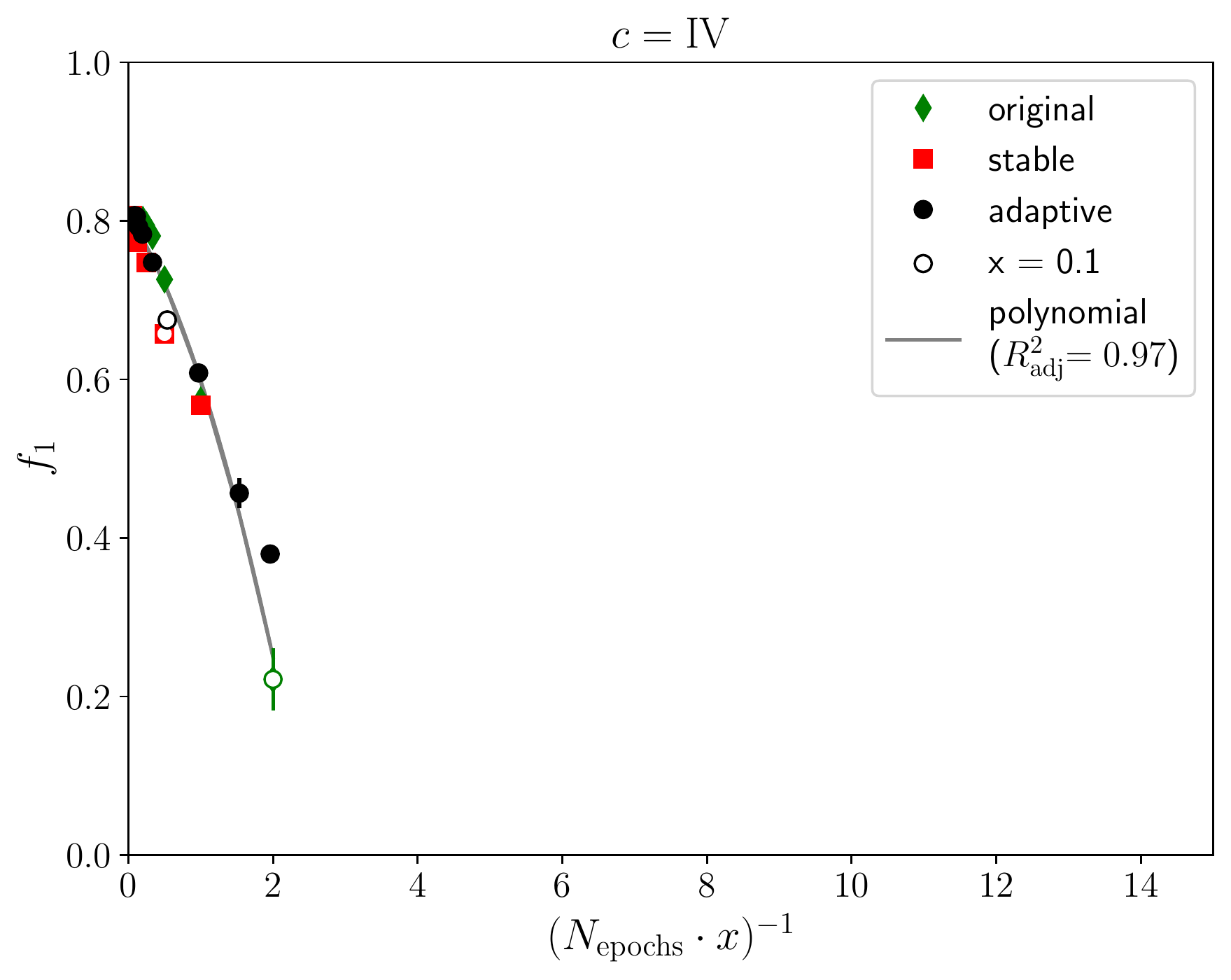}
 \includegraphics[scale=\figScale]{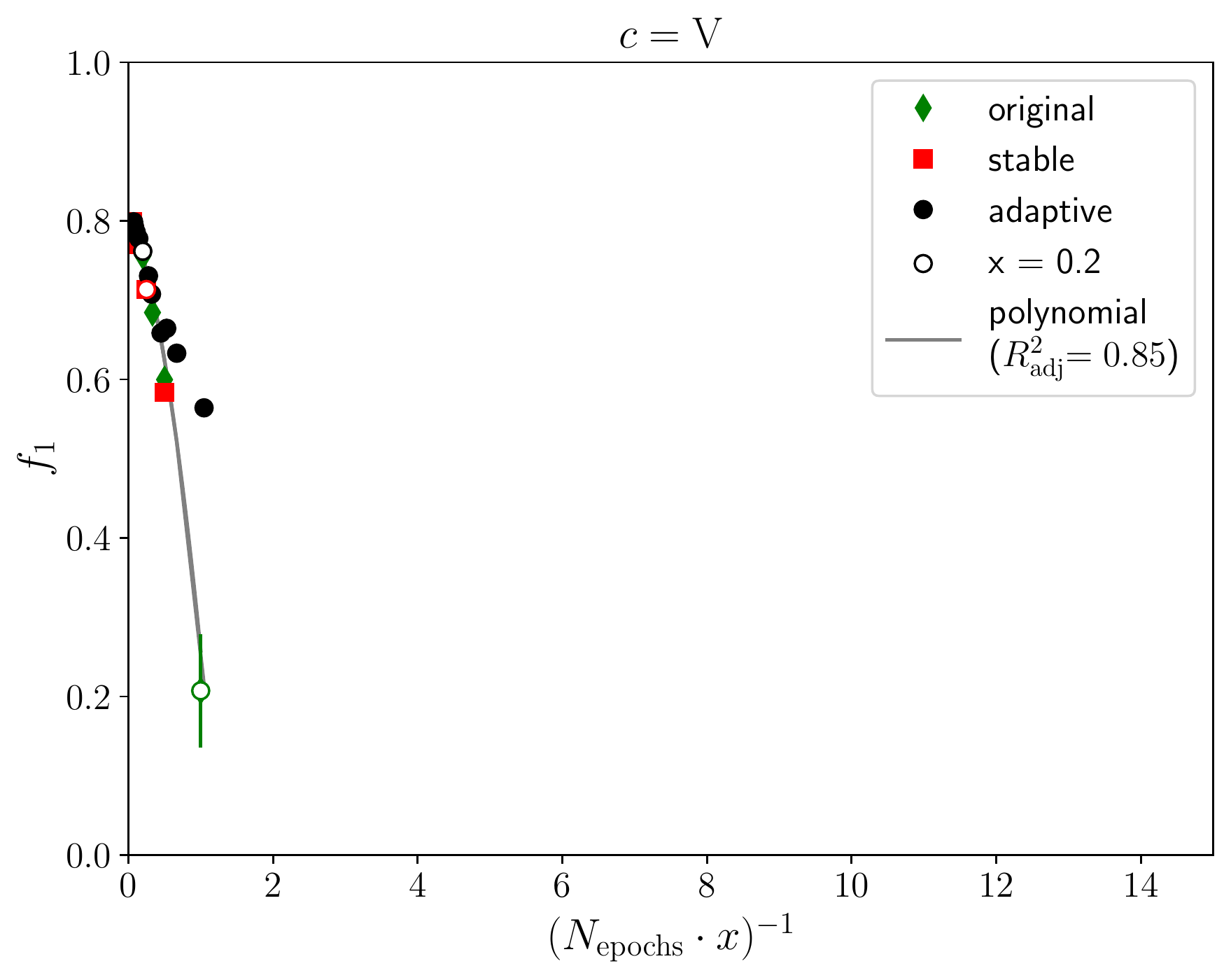} 
 \caption{$f_1$ as a function of $( N_{\rm epochs} \cdot x )^{-1}$ for all $a \in A$.  The results are shown for each dataset-model combination $c \in C$ separately, together with the respective polynomial fits, Eq.~(\ref{eq:polynomial_fit}).  In each plot, one particular dataset scaling factor $x$ is chosen for which the results for the different fine-tuning approaches are displayed as empty circles.}
 \label{fig:polynomial_fit_Nx}
\end{figure}

In the right panel of Fig.~\ref{fig:polynomial_fit_x_and_Nx}, we show $f_1$ as a function of $\left( N_{\rm epochs} \cdot x \right)^{-1}$. Interestingly, irrespective of the fine-tuning approach $a \in A$, the data  can effectively be described by a polynomial of second order\footnote{We tried first, second and third order polynomials. For all $c \in C$, the second order polynomial had the highest adjusted $R^2$ and the lowest mean squared error.} in $( N_{\rm epochs} \cdot x )^{-1}$,
\begin{eqnarray}
 f_1 = a_0 - a_1 \left( N_{\rm epochs} \cdot x \right)^{-1} - a_2 \left( N_{\rm epochs} \cdot x \right)^{-2}
\label{eq:polynomial_fit}
\end{eqnarray}
where $a_0, a_1, a_2 \in \mathbb{R}^{+} \cup \{ 0 \}$ are non-negative fit parameters\footnote{We used weighted least squares and \citep{seabold2010statsmodels}.}. 
Such a polynomial can be found with a reasonable fit quality for any $c \in C$, see Fig.~\ref{fig:polynomial_fit_Nx}.

Note that our data corresponds to $N_{\rm epochs} \leq T(x)$ only.  
However, Eq.~(\ref{eq:polynomial_fit}) can easily be generalized to all $N_{\rm epochs}$ under consideration of Eq.~(\ref{eq:derivative_f1_Nepochs_above_T}) by 
\begin{subequations}
\begin{eqnarray}
 f_1 = a_0 - a_1 \left( N_{\rm epochs}^* \cdot x \right)^{-1} - a_2 \left( N_{\rm epochs}^* \cdot x \right)^{-2}
\end{eqnarray}
with the replacement
\begin{eqnarray}
N_{\rm epochs} \to N_{\rm epochs}^* = \min(N_{\rm epochs}, T(x))
\end{eqnarray}%
\label{eq:polynomial_fit_generalization}%
\end{subequations}
The generalized functional dependency,  Eq.~(\ref{eq:polynomial_fit_generalization}),
reflects the performance boundaries one encounters when increasing $N_{\rm epochs}$ or $x$. Regarding $N_{\rm epochs}$, performance reaches the limit 
\begin{subequations}
\begin{align}
 f_1(x,  N_{\rm epochs} \to \infty) &= a_0 - a_1 \left( T(x) \cdot x \right)^{-1} - a_2 \left( T(x) \cdot x \right)^{-2}
\intertext{
defined by the limited amount of data. Regarding $x$, performance is limited by the nature of the machine learning problem, i.e.  complexity and noise of the data as well as model limitations:
}
 f_1(x \to \infty,  N_{\rm epochs} \to \infty) &= a_0
\end{align}
\end{subequations}

The advantage of the adaptive fine-tuning approach is that it automatically chooses a close-to-optimal number of training epochs for a given dataset, $N_{\rm epochs} \approx T(x)$. Hence, it makes the most of the data, without wasting computational resources.

%% file: tex/appendix_adaptive_finetuning_in_practice_xval_dependency.tex
\begin{figure}[ht]
 \centering
 \includegraphics[scale=\figScale]{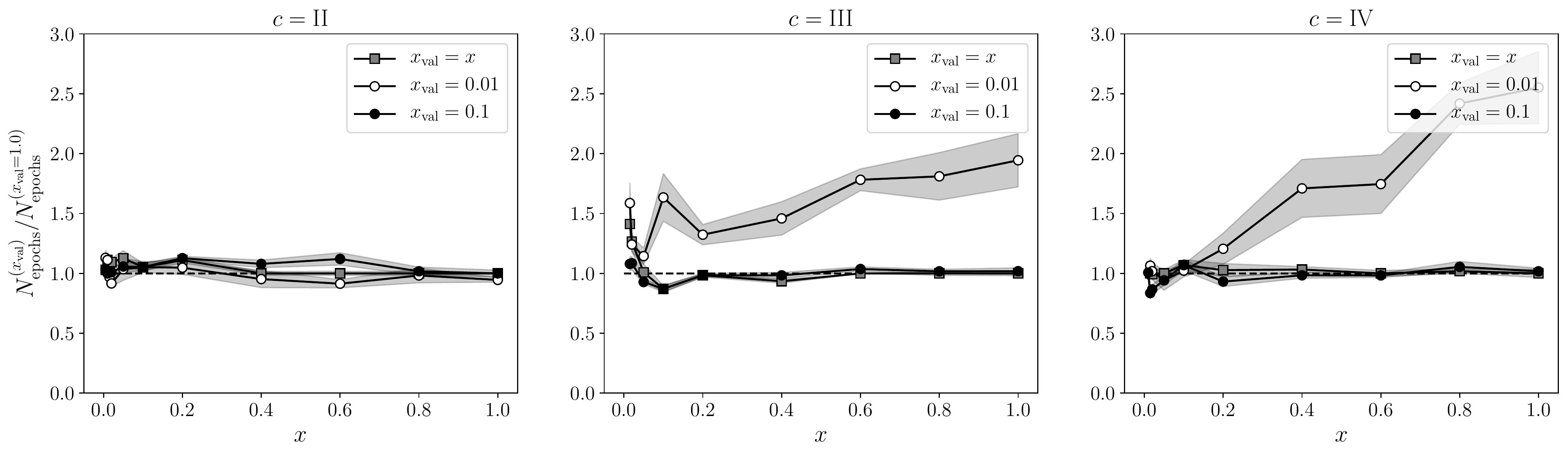}
 \includegraphics[scale=\figScale]{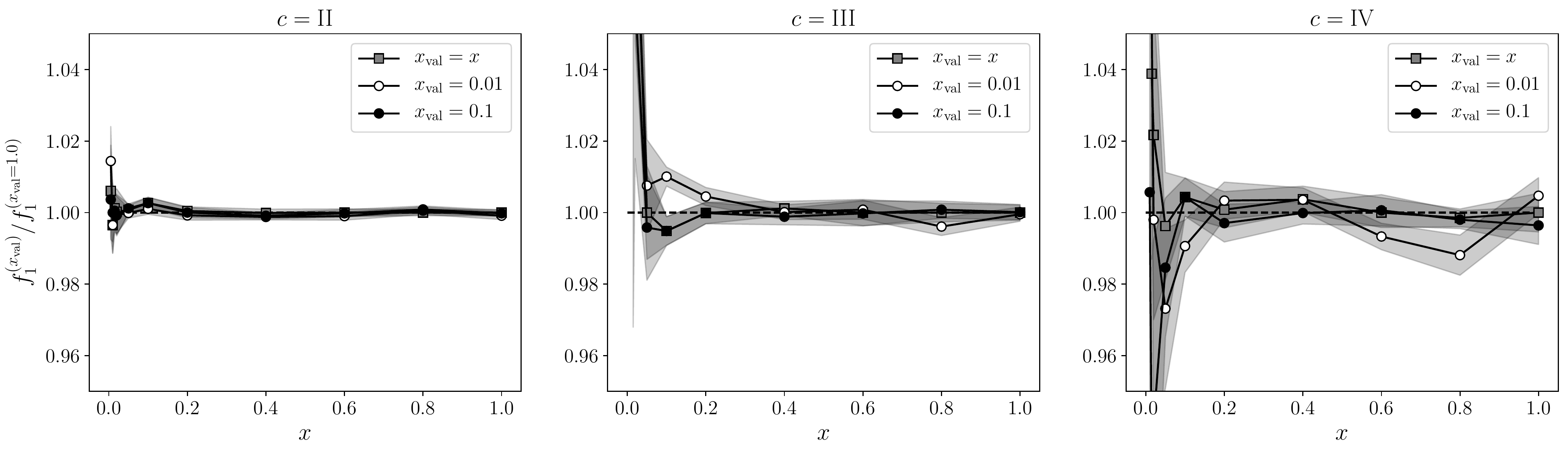}
 \caption{Comparison of adaptive fine-tuning results for $x_{\rm val} \in \{ x_{\rm train}, 0.01, 0.1 \}$ with respect to $x_{\rm val} = 1.0$.  The top (bottom) row shows the respective ratios of $N_{\rm epochs}$ ($f_1$) as a function of the training dataset scaling factor $x_{\rm train}$, separately for $c \in \{ \text{II}, \text{III}, \text{IV} \}$.} 
 \label{fig:7_xval_variation}
\end{figure}

Throughout all our previous experiments, we scaled our validation datasets proportionally to the training datasets, see Eq.~(\ref{eq:xtrain_xval}), such that the ratio $N_{\rm val}^c / N_{\rm train}^c$ always was the same for a given dataset-model combination $c$. As our adaptive fine-tuning approach crucially depends on the validation dataset through the monitoring of the validation loss (see Sec.~\ref{sec:adaptive}), the question arises what impact the size of the validation dataset, determined by $N_{\rm val}$ and the scaling factor $x_{\rm val}$, has on the fine-tuning process. 
To investigate this question, we conduct adaptive fine-tuning experiments for $c \in \{ {\rm II}, {\rm III}, {\rm IV} \}$ using \textit{constant} validation dataset scaling factors, 
\begin{equation}
 x_{\rm val} \in \{ 0.01, 0.1, 1.0 \},
 \label{eq:xval_dependency}
\end{equation} 
that are independent of the usual training dataset scaling factors $x_{\rm train} \in X$.  
The results are listed in Tab.~\ref{tab:experiments_optimization_xval} and visualized in Fig.~\ref{fig:7_xval_variation}.
\begin{table}[ht!]
 \centering
\scalebox{\tableScale}{
  \begin{tabular}{c|c||c|c|c|c||c|c|c|c}
& & \multicolumn{4}{c||}{$N_{\rm epochs}$} & \multicolumn{4}{c}{$f_1$} \\ 
\multirow{2}{*}{$c$} &  \multirow{2}{*}{$x_{\rm train}$} & \multirow{2}{*}{$x_{\rm val} = x_{\rm train}$} & \multirow{2}{*}{$x_{\rm val} = 0.01$} & \multirow{2}{*}{$x_{\rm val} = 0.1$} & \multirow{2}{*}{$x_{\rm val} = 1.0$} & \multirow{2}{*}{$x_{\rm val}= x_{\rm train}$} & \multirow{2}{*}{$x_{\rm val} = 0.01$} & \multirow{2}{*}{$x_{\rm val} = 0.1$} & \multirow{2}{*}{$x_{\rm val} = 1.0$} \\ 
& & & & & & & & & \\ \hline 
 \input{analysis/tex/7_xval_variation.tex}
 \end{tabular}
}
 \caption{Comparison of adaptive fine-tuning results for $c \in \{ \text{II}, \text{III}, \text{IV} \}$ and $x_{\rm train} \in X$ using different validation dataset scaling factors $x_{\rm val}$,  as given in Eq.~(\ref{eq:xval_dependency}). Note that the results for $x_{\rm train} = x_{\rm val} = x$ can also be found in Tab.~\ref{tab:experiments_optimization_ALL}.}
 \label{tab:experiments_optimization_xval}
\end{table}

Let us first discuss $N_{\rm epochs}$.  We observe that the results are very similar for $x_{\rm val} \in \{ x_{\rm train}, 0.1, 1.0 \}$. However, for $x_{\rm val} = 0.01$, there is a tendency for the training to take significantly longer, in particular for large $x_{\rm train}$.  This effect is more pronounced the smaller the size of the original validation dataset is (recall that $N_{\rm val}^{\rm II} > N_{\rm val}^{\rm III} > N_{\rm val}^{\rm IV}$, compare Tab.~\ref{tab:experiments_overview} and Fig.~\ref{fig:dataset_sizes}). This indicates that there is a certain threshold in terms of absolute validation dataset size, above which adaptive fine-tuning can be considered stable. 

The impact of this extensive training on the $f_1$ score is rather small though, which is in accordance with the observation that additional training hardly leads to overfitting, see App.~\ref{app:finetuning_additional} and \citep{mosbach2021stability}.
Regarding $f_1$, we see some instability for small $x_{\rm train}$ instead. Again, there seems to be a threshold in terms of absolute numbers, in this case the size of the training dataset.

To summarize, our results suggest that the adaptive fine-tuning approach is stable unless the training and validation dataset sizes are extremely small. As an estimate for the order of magnitude required, we note that there are hardly any instabilities for $c = {\rm II}$ and $x_{\rm train}, x_{\rm val} \geq 0.01$, as well as $c \in \{ \rm III, IV\}$ and $x_{\rm val} \geq 0.1$. Those scaling factors very roughly correspond to $N_{\rm train}, N_{\rm val} \approx 100$ sentences (cf.~Fig.~\ref{fig:dataset_sizes}). 
We conclude that moderately sized validation datasets are sufficient for adaptive fine-tuning to work well, so that there is no use in having unnecessarily large validation datasets. Instead, spare data can be assigned to the training dataset in the first place. 

Note that the fixed epoch approach also potentially suffers from small validation datasets, as these may lead to reduced accuracy of the results for the different hyperparameter runs and thus the danger of choosing a suboptimal $N_{\rm epochs}$.

%% file: analysis/tex/7_xval_variation.tex
\multirow{11}{*}{II}
 & 0.005 & 42.8(2.9) & 47.0(2.5) & 42.8(1.6) & 41.6(2.0) & 0.6631(72) & 0.6686(46) & 0.6615(59) & 0.6591(66)  \\
 & 0.01 & 29.8(1.2) & 29.8(1.2) & 26.8(5) & 26.8(5) & 0.7404(46) & 0.7404(46) & 0.7430(49) & 0.7430(49)  \\
 & 0.015 & 25.4(7) & 23.8(8) & 25.0(7) & 24.6(7) & 0.7706(31) & 0.7697(35) & 0.7701(31) & 0.7697(31)  \\
 & 0.02 & 23.4(7) & 19.6(6) & 21.6(7) & 21.4(5) & 0.7853(41) & 0.7846(29) & 0.7845(35) & 0.7851(37)  \\
 & 0.05 & 19.6(1.1) & 18.0(1.6) & 18.4(8) & 17.4(2) & 0.8527(6) & 0.8520(8) & 0.8531(5) & 0.8521(11)  \\
 & 0.1 & 16.0(5) & 16.0(7) & 16.0(5) & 15.2(5) & 0.8692(11) & 0.8678(1) & 0.8692(11) & 0.8669(1)  \\
 & 0.2 & 14.0(3) & 13.2(7) & 14.2(2) & 12.6(4) & 0.8837(8) & 0.8826(1) & 0.8833(11) & 0.8833(7)  \\
 & 0.4 & 12.6(2) & 12.0(9) & 13.6(4) & 12.6(4) & 0.8955(8) & 0.8945(5) & 0.8948(6) & 0.8956(6)  \\
 & 0.6 & 11.6(2) & 10.6(4) & 13.0(6) & 11.6(2) & 0.9026(7) & 0.9017(8) & 0.9024(7) & 0.9026(7)  \\
 & 0.8 & 11.4(2) & 11.2(7) & 11.6(4) & 11.4(2) & 0.9041(6) & 0.9046(8) & 0.9049(7) & 0.9041(6)  \\
 & 1.0 & 11.0(0) & 10.4(2) & 11.0(3) & 11.0(0) & 0.9055(6) & 0.9047(8) & 0.9053(6) & 0.9055(6)  \\ \hline
\multirow{11}{*}{III}
 & 0.005 & --- & --- & --- & --- & --- & --- & --- & ---  \\
 & 0.01 & --- & --- & --- & --- & --- & --- & --- & ---  \\
 & 0.015 & 35.6(3.3) & 40.0(4.2) & 27.2(1.5) & 25.2(1.4) & 0.3682(294) & 0.3820(326) & 0.3076(304) & 0.2849(307)  \\
 & 0.02 & 42.0(2.5) & 41.2(2.1) & 36.0(1.8) & 33.2(3.6) & 0.5869(102) & 0.5828(97) & 0.5670(14) & 0.5348(346)  \\
 & 0.05 & 22.4(1.5) & 25.4(1.5) & 20.6(4) & 22.2(1.3) & 0.7417(67) & 0.7473(64) & 0.7386(84) & 0.7417(95)  \\
 & 0.1 & 16.2(5) & 30.4(3.7) & 16.2(5) & 18.6(8) & 0.8108(29) &{ \bf 0.8232(15) }& 0.8108(29) & 0.8150(19)  \\
 & 0.2 & 12.8(2) & 17.2(1.1) & 12.8(2) & 13.0(3) & 0.8456(19) &{ \bf 0.8495(13) }& 0.8456(19) & 0.8457(2)  \\
 & 0.4 & 11.4(2) & 17.8(1.7) & 12.0(3) & 12.2(2) & 0.8621(13) & 0.8612(14) & 0.8601(16) & 0.8611(13)  \\
 & 0.6 & 11.0(0) & 19.6(1.0) & 11.4(2) & 11.0(0) & 0.8609(24) & 0.8616(1) & 0.8607(22) & 0.8609(24)  \\
 & 0.8 & 11.6(2) & 21.0(2.3) & 11.8(2) & 11.6(2) & 0.8691(17) & 0.8657(16) & 0.8698(15) & 0.8691(17)  \\
 & 1.0 & 10.8(2) & 21.0(2.4) & 11.0(3) & 10.8(2) & 0.8717(14) & 0.8713(12) & 0.8718(14) & 0.8717(14)  \\ \hline
\multirow{11}{*}{IV}
 & 0.005 & --- & --- & --- & 37.8(1.1) & --- & --- & --- &{ \bf 0.0369(63)  }\\
 & 0.01 & --- & --- & 30.6(1.8) & 30.4(1.0) & --- & --- & 0.1946(162) & 0.1935(122)  \\
 & 0.015 & 34.0(1.1) & 36.2(2.0) & 28.4(1.5) & 34.0(2.3) & 0.3798(81) & 0.3856(123) & 0.3418(186) & 0.3656(117)  \\
 & 0.02 & 32.6(1.5) & 33.4(2.3) & 28.4(1.4) & 32.8(2.7) & 0.4566(187) & 0.4460(247) & 0.4205(194) & 0.4469(297)  \\
 & 0.05 & 20.6(6) & 19.4(1.7) & 19.4(8) & 20.6(7) & 0.6082(78) & 0.5941(128) & 0.6011(109) & 0.6105(79)  \\
 & 0.1 & 18.6(8) & 17.8(9) & 18.6(8) & 17.4(4) & 0.6750(31) & 0.6658(46) & 0.6750(31) & 0.6721(27)  \\
 & 0.2 & 15.0(8) & 17.6(1.9) & 13.6(6) & 14.6(5) & 0.7477(32) & 0.7496(33) & 0.7449(34) & 0.7471(27)  \\
 & 0.4 & 12.8(3) & 21.2(3.0) & 12.2(3) & 12.4(2) & 0.7833(21) & 0.7832(17) & 0.7803(14) & 0.7804(25)  \\
 & 0.6 & 11.8(3) & 20.6(2.9) & 11.6(2) & 11.8(3) & 0.7923(26) & 0.7870(21) & 0.7928(28) & 0.7923(26)  \\
 & 0.8 & 11.2(2) & 26.6(1.9) & 11.6(5) & 11.0(0) & 0.8057(14) & 0.7973(43) & 0.8053(14) & 0.8069(18)  \\
 & 1.0 & 11.6(4) & 29.6(3.5) & 11.8(3) & 11.6(4) & 0.8067(34) &{ \bf 0.8105(3) }& 0.8038(33) & 0.8067(34)

%% file: tex/appendix_adaptive_finetuning_in_practice_train_on_val.tex
In practice, after hyperparameter search, the model sometimes is retrained while adding the validation dataset to the training dataset to further improve model performance.
In that case, the optimal hyperparameters found using the pure training dataset are reused, assuming that they are also optimal on the extended training dataset. 
The results in this paper (see e.g. ~Tab.~\ref{tab:experiments_optimization}) indicate that this assumption is well-justified for the adaptive fine-tuning approach, as the optimal $N_{\rm epochs}$ only decreases if the dataset size is increased, while training for some additional epochs does not do any harm. 

An important difference when using the extended dataset and adaptive fine-tuning is that $N_{\rm epochs}$ is not dynamically determined as before, but specified beforehand using the $N_{\rm epochs}$ found on the pure training dataset, in combination with the hybrid learning rate schedule of the adaptive approach (cf.~Fig.~\ref{fig:training_hyperparameters_overview}). In particular, the validation of the model after each training epoch is discarded. As App.~\ref{app:finetuning_ablation} shows, this procedure does not affect the results apart from the usual statistical fluctuations. 

Consequently, the adaptive approach should work well in the described scenario. To verify this, and to see how it compares to the fixed epoch fine-tuning approach, we repeat the fine-tuning experiments for $a \in A$ and $c \in \{ \text{II}, \text{III}, \text{IV} \}$ (see Sec.~\ref{sec:results_fine_tuning} and App.~\ref{app:finetuning_generalization}), while using the validation set as additional training data. We denote the results as $f_1^+$, and consider the ratio $f_1^+ / f_1$ with the performance $f_1$ obtained on only the original training data. This number represents the factor by which the performance improves when adding the validation data to the training data. 

\begin{figure}[ht]
 \centering
 \includegraphics[scale=\figScale]{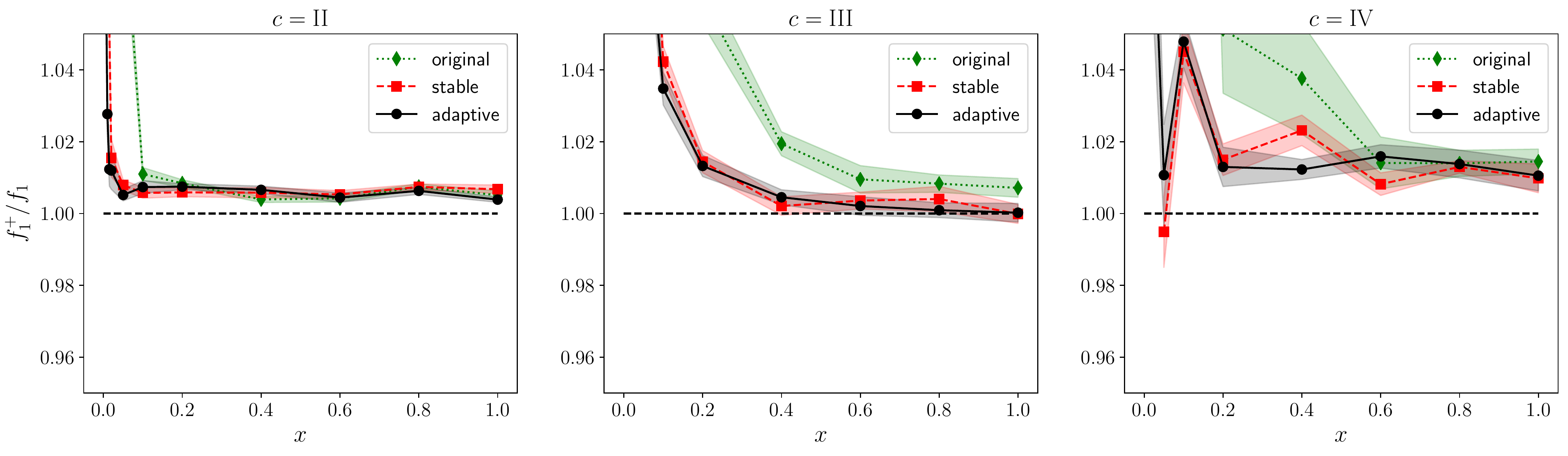}
 \caption{Ratio $f_1^+ / f_1$ of the model performance with the extended training dataset and the model performance with the pure training dataset, in dependence of the dataset scaling factor $x$. The results are shown for all fine-tuning approaches $a \in A$, and separately for $c \in \{ \text{II}, \text{III}, \text{IV} \}$.}
 \label{fig:7_train+val}
\end{figure}

The results can be found in Tab.~\ref{tab:experiments_optimization_train_plus_validation} and Fig.~\ref{fig:7_train+val}.
\begin{table}[ht!]
 \centering
\scalebox{\tableScale}{
  \begin{tabular}{c|c||c|c|c|c|c|c}
  \multicolumn{2}{c||}{approach} & \multicolumn{2}{c|}{original} & \multicolumn{2}{c|}{stable} & \multicolumn{2}{c}{adaptive} \\ \hline
  \multicolumn{2}{c||}{early stopping} & \multicolumn{2}{c|}{no} & \multicolumn{2}{c|}{no} & \multicolumn{2}{c}{no} \\
  \multicolumn{2}{c||}{training epochs} & \multicolumn{2}{c|}{5} & \multicolumn{2}{c|}{20} & \multicolumn{2}{c}{adaptive} \\
  \multicolumn{2}{c||}{learning rate schedule} & \multicolumn{2}{c|}{linear} & \multicolumn{2}{c|}{linear} & \multicolumn{2}{c}{hybrid} \\ \hline
  \multirow{2}{*}{$c$} & \multirow{2}{*}{$x$} &  \multirow{2}{*}{$f_1^+$} & \multirow{2}{*}{$f_1^+ / f_1$} & \multirow{2}{*}{$f_1^+$} &  \multirow{2}{*}{$f_1^+ / f_1$} & \multirow{2}{*}{$f_1^+$} & \multirow{2}{*}{$f_1^+ / f_1$} \\ 
    & & & & & & & \\ \hline
 \input{analysis/tex/7_train+val.tex}
 \end{tabular}
}
 \caption{Fine-tuning results for $a \in A$ and $c \in \{ \text{II}, \text{III}, \text{IV} \}$, while using the validation set for training. }
 \label{tab:experiments_optimization_train_plus_validation}
\end{table}

As expected, the $f_1$ score benefits from the additional data generally,  i.e.  for all fine-tuning approaches $a$, dataset-model combinations $c$ and dataset scaling factors $x$. The extent by which they improve varies though. It is a lot bigger for small datasets, and for those, the original setup with its low number of $N_{\rm epochs}$ benefits more from the additional data than the stable or adaptive fine-tuning approaches.  
This is in accordance with the discussion in App.~\ref{app:finetuning_dependency}, see e.g.  Eq.~(\ref{eq:polynomial_fit}) and Fig.~\ref{fig:polynomial_fit_x_and_Nx}.

In conclusion, we observe that the findings discussed in Sec.~\ref{sec:results_fine_tuning} also hold in the scenario where the validation dataset is used for training after hyperparameter search. The adaptive fine-tuning approach shows improved behavior in comparison with the fixed epoch approach, namely better performance for small datasets and improved efficiency for larger datasets.

%% file: analysis/tex/7_train+val.tex
\multirow{11}{*}{II}
 & 0.005 & --- & --- & 0.3760(392) & 1.892(203) &{ \bf 0.7175(31) }& 1.082(9)  \\
 & 0.01 & --- & --- & 0.6538(73) & 1.104(14) &{ \bf 0.7609(29) }& 1.028(6)  \\
 & 0.015 & --- & --- & 0.7669(2) & 1.052(8) &{ \bf 0.7801(27) }& 1.012(5)  \\
 & 0.02 & 0.3792(464) & 3.046(394) & 0.7926(28) & 1.016(5) & 0.7947(25) & 1.012(5)  \\
 & 0.05 & 0.7747(2) & 1.090(8) & 0.8535(8) & 1.008(1) &{ \bf 0.8571(13) }& 1.005(2)  \\
 & 0.1 & 0.8558(12) & 1.011(2) & 0.8734(11) & 1.006(2) &{ \bf 0.8756(12) }& 1.007(2)  \\
 & 0.2 & 0.8872(7) & 1.008(1) & 0.8901(7) & 1.006(1) & 0.8903(2) & 1.007(1)  \\
 & 0.4 & 0.8984(7) & 1.004(1) & 0.9015(12) & 1.006(1) & 0.9014(6) & 1.007(1)  \\
 & 0.6 & 0.9049(9) & 1.004(1) &{ \bf 0.9084(9) }& 1.005(1) & 0.9066(1) & 1.004(1)  \\
 & 0.8 & 0.9082(7) & 1.007(1) &{ \bf 0.9109(5) }& 1.007(1) & 0.9098(7) & 1.006(1)  \\
 & 1.0 & 0.9074(8) & 1.005(1) & 0.9098(7) & 1.007(1) & 0.9090(4) & 1.004(1)  \\ \hline
\multirow{11}{*}{III}
 & 0.005 & --- & --- & --- & --- & --- & ---  \\
 & 0.01 & --- & --- &{ \bf 0.0992(247) }& --- & --- & ---  \\
 & 0.015 & --- & --- & 0.3553(125) & --- &{ \bf 0.5854(88) }& 1.590(52)  \\
 & 0.02 & --- & --- & 0.5278(79) & 2.172(37) &{ \bf 0.6849(71) }& 1.167(17)  \\
 & 0.05 & 0.1966(455) & --- & 0.7869(66) & 1.146(15) &{ \bf 0.8117(49) }& 1.094(1)  \\
 & 0.1 & 0.7175(48) & 1.240(16) & 0.8386(19) & 1.042(4) & 0.8390(28) & 1.035(5)  \\
 & 0.2 & 0.8385(13) & 1.060(7) & 0.8578(18) & 1.014(3) & 0.8568(19) & 1.013(3)  \\
 & 0.4 & 0.8659(21) & 1.019(3) & 0.8639(18) & 1.002(3) & 0.8660(14) & 1.005(2)  \\
 & 0.6 &{ \bf 0.8702(28) }& 1.010(4) & 0.8640(13) & 1.004(2) & 0.8627(8) & 1.002(3)  \\
 & 0.8 &{ \bf 0.8729(1) }& 1.008(2) & 0.8697(29) & 1.004(4) & 0.8699(1) & 1.001(2)  \\
 & 1.0 &{ \bf 0.8757(2) }& 1.007(3) & 0.8699(15) & 1.000(3) & 0.8719(19) & 1.000(3)  \\ \hline
\multirow{11}{*}{IV}
 & 0.005 & --- & --- & --- & --- & --- & ---  \\
 & 0.01 & --- & --- & --- & --- & --- & ---  \\
 & 0.015 & --- & --- & 0.1041(189) & --- &{ \bf 0.4396(115) }& 1.157(32)  \\
 & 0.02 & --- & --- & 0.2801(128) & --- &{ \bf 0.4999(118) }& 1.095(33)  \\
 & 0.05 & --- & --- & 0.5642(49) & 0.995(1) &{ \bf 0.6147(71) }& 1.011(14)  \\
 & 0.1 & 0.3445(344) & 1.554(167) & 0.6872(5) & 1.045(9) &{ \bf 0.7073(44) }& 1.048(7)  \\
 & 0.2 & 0.6027(93) & 1.051(18) & 0.7588(3) & 1.015(4) & 0.7574(33) & 1.013(5)  \\
 & 0.4 & 0.7535(68) & 1.038(15) & 0.7913(2) & 1.023(4) &{ \bf 0.7929(14) }& 1.012(3)  \\
 & 0.6 & 0.7920(36) & 1.014(7) & 0.8015(21) & 1.008(3) &{ \bf 0.8049(15) }& 1.016(3)  \\
 & 0.8 & 0.8056(25) & 1.014(4) & 0.8121(8) & 1.013(2) &{ \bf 0.8168(29) }& 1.014(4)  \\
 & 1.0 & 0.8127(23) & 1.014(3) & 0.8148(18) & 1.010(4) & 0.8152(2) & 1.011(4)

%% file: main.bbl
\begin{thebibliography}{10}

\bibitem{wolf-etal-2020-transformers}
Thomas Wolf, Lysandre Debut, Victor Sanh, Julien Chaumond, Clement Delangue,
  Anthony Moi, Pierric Cistac, Tim Rault, Rémi Louf, Morgan Funtowicz, Joe
  Davison, Sam Shleifer, Patrick von Platen, Clara Ma, Yacine Jernite, Julien
  Plu, Canwen Xu, Teven~Le Scao, Sylvain Gugger, Mariama Drame, Quentin Lhoest,
  and Alexander~M. Rush.
\newblock Transformers: State-of-the-art natural language processing.
\newblock In {\em Proceedings of the 2020 Conference on Empirical Methods in
  Natural Language Processing: System Demonstrations}, pages 38--45, Online,
  October 2020. Association for Computational Linguistics.
\newblock \url{https://www.aclweb.org/anthology/2020.emnlp-demos.6}.

\bibitem{devlin2019bert}
Jacob Devlin, Ming-Wei Chang, Kenton Lee, and Kristina Toutanova.
\newblock Bert: Pre-training of deep bidirectional transformers for language
  understanding, 2019.

\bibitem{zhang2021revisiting}
Tianyi Zhang, Felix Wu, Arzoo Katiyar, Kilian~Q. Weinberger, and Yoav Artzi.
\newblock Revisiting few-sample bert fine-tuning, 2021.

\bibitem{mosbach2021stability}
Marius Mosbach, Maksym Andriushchenko, and Dietrich Klakow.
\newblock On the stability of fine-tuning bert: Misconceptions, explanations,
  and strong baselines, 2021.

\bibitem{dodge2020finetuning}
Jesse Dodge, Gabriel Ilharco, Roy Schwartz, Ali Farhadi, Hannaneh Hajishirzi,
  and Noah Smith.
\newblock Fine-tuning pretrained language models: Weight initializations, data
  orders, and early stopping, 2020.

\bibitem{dodge2019work}
Jesse Dodge, Suchin Gururangan, Dallas Card, Roy Schwartz, and Noah~A. Smith.
\newblock Show your work: Improved reporting of experimental results, 2019.

\bibitem{clark2020electra}
Kevin Clark, Minh-Thang Luong, Quoc~V. Le, and Christopher~D. Manning.
\newblock Electra: Pre-training text encoders as discriminators rather than
  generators, 2020.

\bibitem{lee2020mixout}
Cheolhyoung Lee, Kyunghyun Cho, and Wanmo Kang.
\newblock Mixout: Effective regularization to finetune large-scale pretrained
  language models, 2020.

\bibitem{tjong-kim-sang-de-meulder-2003-introduction}
Erik~F. Tjong Kim~Sang and Fien De~Meulder.
\newblock Introduction to the {C}o{NLL}-2003 shared task: Language-independent
  named entity recognition.
\newblock In {\em Proceedings of the Seventh Conference on Natural Language
  Learning at {HLT}-{NAACL} 2003}, pages 142--147, 2003.
\newblock \url{https://www.aclweb.org/anthology/W03-0419}.

\bibitem{Sanh2019DistilBERTAD}
Victor Sanh, Lysandre Debut, Julien Chaumond, and Thomas Wolf.
\newblock Distilbert, a distilled version of bert: smaller, faster, cheaper and
  lighter.
\newblock {\em ArXiv}, abs/1910.01108, 2019.

\bibitem{swedish-ner-corpus}
\url{https://github.com/klintan/swedish-ner-corpus}.

\bibitem{swedish-bert}
Martin Malmsten, Love Börjeson, and Chris Haffenden.
\newblock Playing with words at the national library of sweden -- making a
  swedish bert, 2020.

\bibitem{swe_nerc}
Lars Ahrenberg, Johan Frid, and Leif-Joran Olsson.
\newblock A new resource for swedish named-entity recognition, 2020.
\newblock \url{https://gubox.app.box.com/v/SLTC-2020-paper-17}.

\bibitem{overview_ehealthkd2020}
Alejandro Piad{-}Morffis, Yoan Guti{\'{e}}rrez, Hian Ca{\~{n}}izares-Diaz,
  Suilan Estevez{-}Velarde, Yudivi{\'{a}}n Almeida{-}Cruz, Rafael Mu{\~{n}}oz,
  and Andr{\'{e}}s Montoyo.
\newblock Overview of the ehealth knowledge discovery challenge at iberlef
  2020.
\newblock 2020.

\bibitem{transformers-electricidad}
\url{https://huggingface.co/mrm8488/electricidad-base-discriminator}.

\bibitem{ramshaw1995text}
Lance~A. Ramshaw and Mitchell~P. Marcus.
\newblock Text chunking using transformation-based learning, 1995.

\bibitem{nerblackbox}
Felix Stollenwerk.
\newblock nerblackbox: a python package to fine-tune transformer-based language
  models for named entity recognition, 2021.
\newblock \url{https://github.com/flxst/nerblackbox}.

\bibitem{transformers-conll2003-bert}
\url{https://huggingface.co/dslim/bert-base-NER}.

\bibitem{transformers-conll2003-distilbert}
\url{https://huggingface.co/gunghio/distilbert-base-multilingual-cased-finetuned-conll2003-ner}.

\bibitem{kaplan2020scaling}
Jared Kaplan, Sam McCandlish, Tom Henighan, Tom~B. Brown, Benjamin Chess, Rewon
  Child, Scott Gray, Alec Radford, Jeffrey Wu, and Dario Amodei.
\newblock Scaling laws for neural language models, 2020.

\bibitem{openai}
\url{https://openai.com/blog/ai-and-compute/}.

\bibitem{seabold2010statsmodels}
Skipper Seabold and Josef Perktold.
\newblock statsmodels: Econometric and statistical modeling with python.
\newblock In {\em 9th Python in Science Conference}, 2010.

\end{thebibliography}
